%% file: main.tex
\documentclass{article}

\usepackage{microtype}
\usepackage{graphicx}
\usepackage{subfigure}
\usepackage{booktabs} % for professional tables

\usepackage{hyperref}

\usepackage[accepted]{icml2024}

\usepackage{amsmath}
\usepackage{amssymb}
\usepackage{mathtools}
\usepackage{amsthm}
\usepackage[capitalize,noabbrev]{cleveref}

\definecolor{Green}{rgb}{0.95,1,0.95}

\theoremstyle{plain}
\newtheorem{theorem}{Theorem}[section]

\theoremstyle{definition}

\theoremstyle{remark}

\usepackage{wrapfig, blindtext} % adjusting layout
\usepackage[export]{adjustbox}  % adjusting layout
\usepackage{colortbl}
\usepackage{xcolor}
\usepackage{booktabs}       % professional-quality tables
\usepackage{multirow}       % for multirow support
\usepackage{multicol}       % for multicol support
\usepackage{array}          % for dash lines
\usepackage{threeparttable}
\usepackage{amsmath}
\usepackage{amsfonts}
\usepackage{dsfont}
\usepackage{bm}
\PassOptionsToPackage{table}{xcolor}
\let\ul\underline
\definecolor{Gray}{gray}{0.9}
\newcolumntype{g}{>{\columncolor{Gray}}l}
\newcommand{\alg}[1]{\textsc{Bat}}
\newcommand{\algp}[1]{\alg{}$_0$}
\newcommand{\algt}[1]{\alg{}$_1$}
\newcommand{\algfull}[1]{\ul{BA}lanced \ul{T}opological augmentation}
\newtoggle{highlightrev}
\togglefalse{highlightrev}

\newcommand{\rev}[1]{%
    \iftoggle{highlightrev}%
        {\textcolor{blue}{#1}}%
        {#1}%
}

\usepackage[textsize=tiny]{todonotes}

\icmltitlerunning{Class-Imbalanced Graph Learning without Class Rebalancing}

\input{math.tex}

\begin{document}

\twocolumn[
\icmltitle{Class-Imbalanced Graph Learning without Class Rebalancing}




\begin{icmlauthorlist}
\icmlauthor{Zhining Liu}{uiuc}
\icmlauthor{Ruizhong Qiu}{uiuc}
\icmlauthor{Zhichen Zeng}{uiuc}
\icmlauthor{Hyunsik Yoo}{uiuc}
\icmlauthor{David Zhou}{uiuc}
\icmlauthor{Zhe Xu}{uiuc}
\icmlauthor{Yada Zhu}{ibm}
\icmlauthor{Kommy Weldemariam}{amazon}
\icmlauthor{Jingrui He}{uiuc}
\icmlauthor{Hanghang Tong}{uiuc}
\end{icmlauthorlist}

\icmlaffiliation{uiuc}{University of Illinois Urbana-Champaign}
\icmlaffiliation{ibm}{IBM Research}
\icmlaffiliation{amazon}{Amazon Science}

\icmlcorrespondingauthor{Zhining Liu}{liu326@illinois.edu}
\icmlcorrespondingauthor{Hanghang Tong}{htong@illinois.edu}

\icmlkeywords{Machine Learning, ICML}

\vskip 0.3in
]



\printAffiliationsAndNotice{}  % leave blank if no need to mention equal contribution

\begin{abstract}
Class imbalance is prevalent in real-world node classification tasks and poses great challenges for graph learning models.
Most existing studies are rooted in a \emph{class-rebalancing} (CR) perspective and address class imbalance with class-wise reweighting or resampling.
In this work, we approach the root cause of class-imbalance bias from an topological paradigm.
Specifically, we theoretically reveal two fundamental phenomena in the graph topology that greatly exacerbate the predictive bias stemming from class imbalance.
On this basis, we devise a lightweight topological augmentation framework \alg{} to mitigate the class-imbalance bias \textit{without class rebalancing}.
Being orthogonal to CR, \alg{} can function as an \ul{\textit{efficient plug-and-play module}} that can be seamlessly combined with and significantly boost existing CR techniques.
Systematic experiments on real-world imbalanced graph learning tasks show that \alg{} can deliver up to 46.27\% performance gain and up to 72.74\% bias reduction over existing techniques.
Code, examples, and documentations are available at \url{https://github.com/ZhiningLiu1998/BAT}.
\end{abstract}

\input{sections/1-introduction}
\input{sections/2-analysis}
\input{sections/3-methodology}
\input{sections/4-experiments}
\input{sections/5-relatedwork}
\input{sections/6-conclusion}

\section*{Acknowledgements}
This work is supported by NSF (1947135), the NSF Program on Fairness in AI in collaboration with Amazon (1939725), NIFA (2020-67021-32799), DHS (17STQAC00001-07-00), the C3.ai Digital Transformation Institute, MIT-IBM Watson AI Lab, and IBM-Illinois Discovery Accelerator Institute.
The content of the information in this document does not necessarily reflect the position or the policy of the Government or Amazon, and no official endorsement should be inferred.  The U.S. Government is authorized to reproduce and distribute reprints for Government purposes notwithstanding any copyright notation here on.

\section*{Impact Statements}

This paper presents work whose goal is to advance the field of Graph Data Mining. There are many potential societal consequences of our work, none which we feel must be specifically highlighted here.

\bibliography{ref}
\bibliographystyle{icml2024}

%%%%%%%%%%%%%%%%%%%%%%%%%%%%%%%%%%%%%%%%%%%%%%%%%%%%%%%%%%%%%%%%%%%%%%%%%%%%%%%
%%%%%%%%%%%%%%%%%%%%%%%%%%%%%%%%%%%%%%%%%%%%%%%%%%%%%%%%%%%%%%%%%%%%%%%%%%%%%%%
% APPENDIX
%%%%%%%%%%%%%%%%%%%%%%%%%%%%%%%%%%%%%%%%%%%%%%%%%%%%%%%%%%%%%%%%%%%%%%%%%%%%%%%
%%%%%%%%%%%%%%%%%%%%%%%%%%%%%%%%%%%%%%%%%%%%%%%%%%%%%%%%%%%%%%%%%%%%%%%%%%%%%%%
\newpage
\appendix
\onecolumn
\input{sections/appendix}


% You can have as much text here as you want. The main body must be at most $8$ pages long.
% For the final version, one more page can be added.
% If you want, you can use an appendix like this one.  

% The $\mathtt{\backslash onecolumn}$ command above can be kept in place if you prefer a one-column appendix, or can be removed if you prefer a two-column appendix.  Apart from this possible change, the style (font size, spacing, margins, page numbering, etc.) should be kept the same as the main body.
%%%%%%%%%%%%%%%%%%%%%%%%%%%%%%%%%%%%%%%%%%%%%%%%%%%%%%%%%%%%%%%%%%%%%%%%%%%%%%%
%%%%%%%%%%%%%%%%%%%%%%%%%%%%%%%%%%%%%%%%%%%%%%%%%%%%%%%%%%%%%%%%%%%%%%%%%%%%%%%

\end{document}

%% file: math.tex
\usepackage{amsmath,amsfonts,bm}

\newcommand\RM[1]{\mathrm{#1}}

\def\e{{\mathrm{e}}}
\def\Exp{{\mathbb{E}}}
\def\Prb{{\mathds{P}}}

\def\eqref#1{equation~\ref{#1}}
\def\1{\bm{1}}

\def\vp{{\bm{p}}}

\def\vr{{\bm{r}}}
\def\vs{{\bm{s}}}

\def\vx{{\bm{x}}}
\def\vy{{\bm{y}}}

\def\mA{{\bm{A}}}

\def\mP{{\bm{P}}}
\def\mQ{{\bm{Q}}}

\def\mS{{\bm{S}}}

\def\mX{{\bm{X}}}

\DeclareMathAlphabet{\mathsfit}{\encodingdefault}{\sfdefault}{m}{sl}
\SetMathAlphabet{\mathsfit}{bold}{\encodingdefault}{\sfdefault}{bx}{n}

\def\gE{{\mathcal{E}}}

\def\gG{{\mathcal{G}}}
\def\gH{{\mathcal{H}}}

\def\gN{{\mathcal{N}}}
\def\gO{{\mathcal{O}}}

\def\gV{{\mathcal{V}}}

\def\sU{{\mathbb{U}}}

\DeclareMathOperator*{\argmax}{arg\,max}

\usepackage{amsthm}
\newcommand\AM[1]{\begin{align*}#1\end{align*}}

%% file: sections/1-introduction.tex
\section{Introduction}
\label{sec:introduction}

\begin{figure*}[t]
\hspace{15pt}
  \subfigure[
  \textbf{L}: concept of AMP.
  \textbf{R}: relative performance loss with respect to the non-self-class neighbor ratio.
  ]{
    \label{fig:intro-conf}
    \centering
    \includegraphics[width=0.4\linewidth]{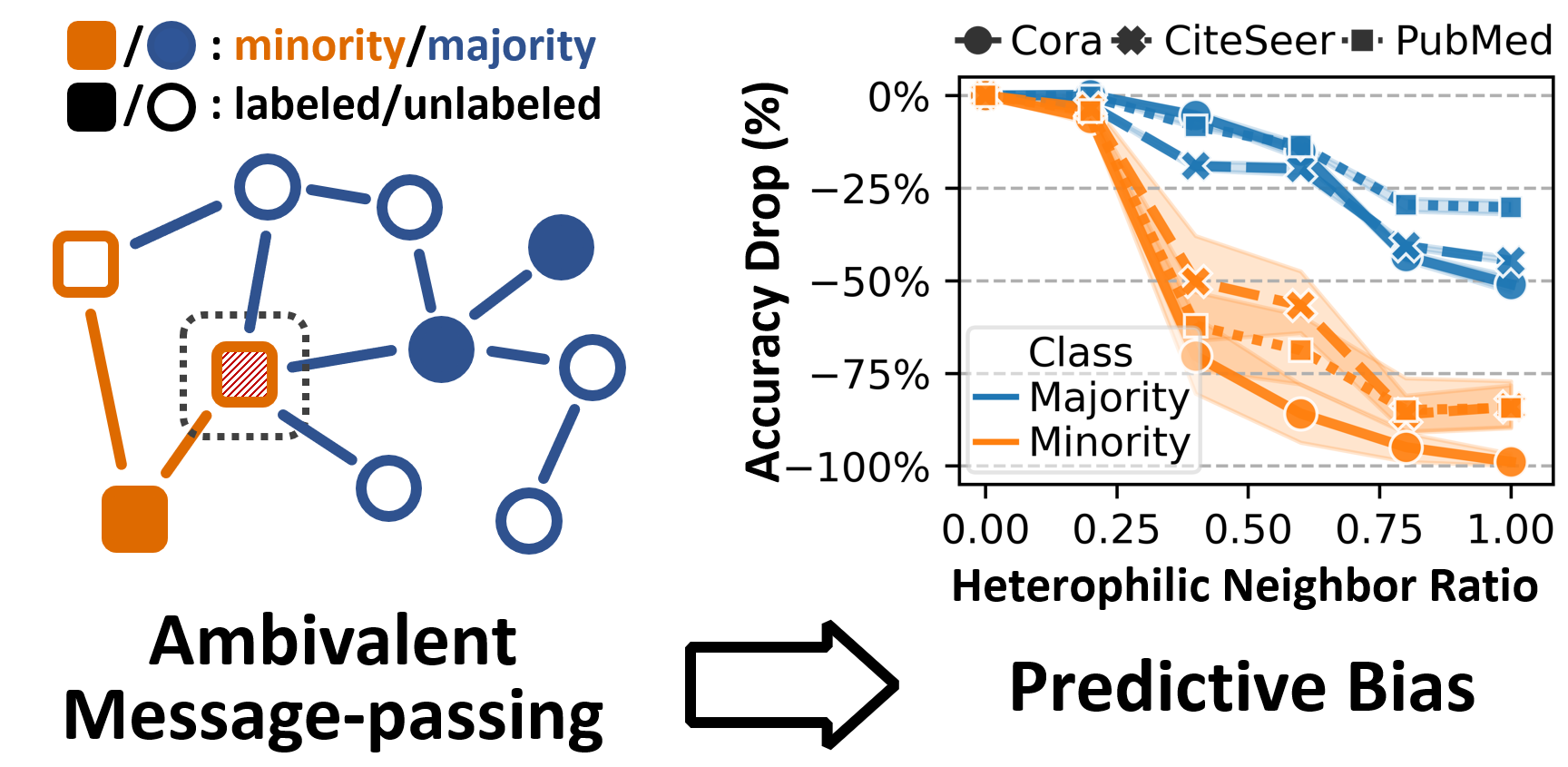}
  }
\hspace{40pt}
  \subfigure[
  \textbf{L}: concept of DMP. 
  \textbf{R}: relative performance loss w.r.t. distance to the nearest same-class labeled node.
  ]{
    \label{fig:intro-insu}
    \centering
    \includegraphics[width=0.4\linewidth]{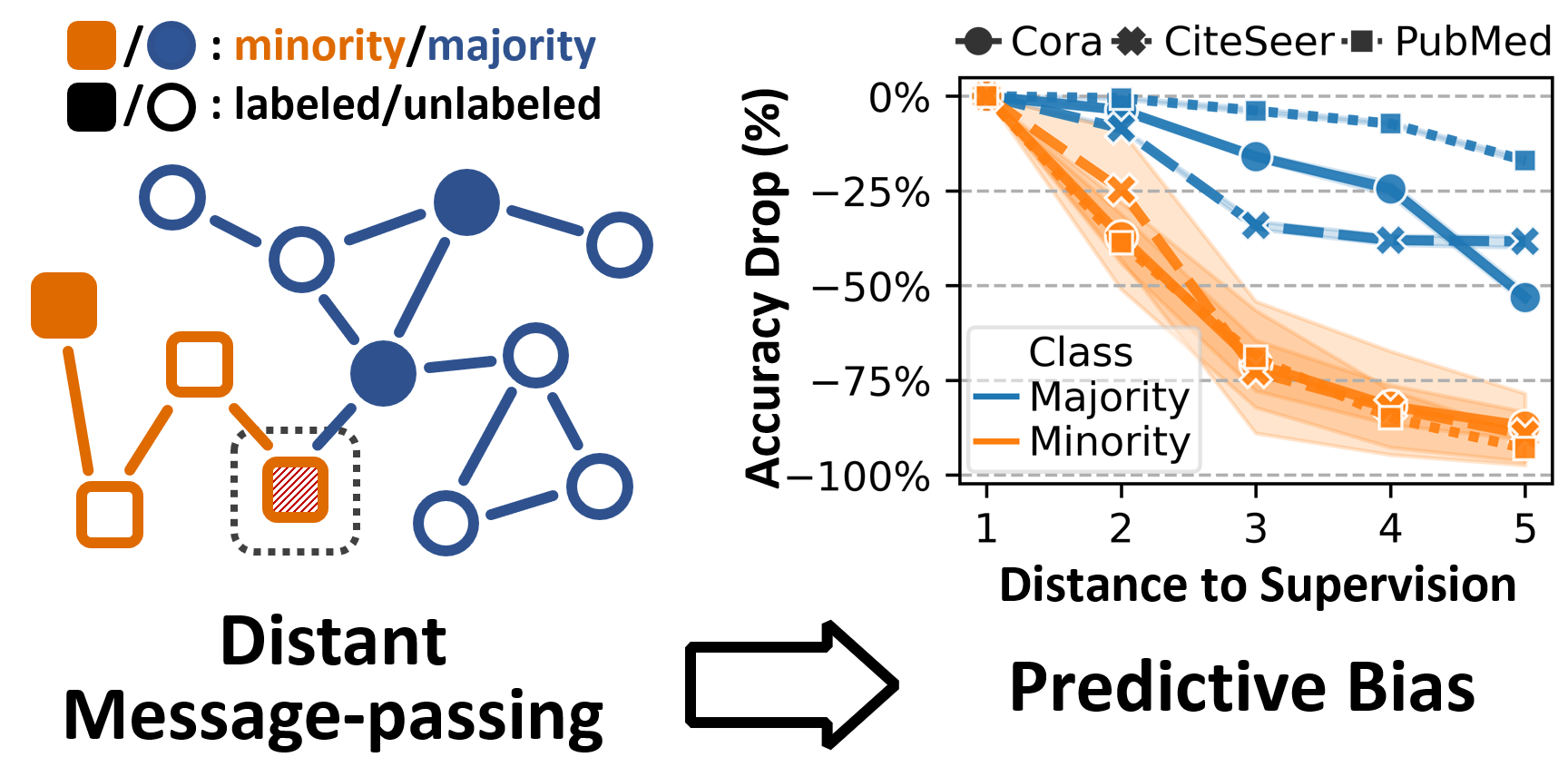}
  }
  \caption{
    Concepts of \textit{ambivalent message-passing} (AMP) and \textit{distant message-passing} (DMP) and their impact in real-world imbalanced node classification tasks~\citep{park2021graphens}.
    Both factors lead to a substantial increase in prediction errors, and further, \textbf{a larger performance disparity/bias} (i.e., the gap between the blue and orange curves) between the majority and minority classes.
  }
  \label{fig:intro}
\end{figure*}

Node classification stands as one of the most fundamental tasks in graph machine learning, holding significant relevance in various real-world applications~\citep{akoglu2015graph,tang2010community}.
Graph Neural Networks (GNNs) have demonstrated great success in tackling related tasks due to their robust representation learning capabilities~\citep{song2022graph,fu2021sdg}.
However, real-world graphs are often inherently class-imbalanced, i.e., the sizes of unique classes vary significantly, and a few majority classes have overwhelming numbers in the training set.
In \emph{Class-Imbalanced Graph Learning} (CIGL), GNNs are prone to suffer from severe performance degradation on minority class nodes~\citep{park2021graphens}.
This results in a pronounced predictive bias characterized by a large performance disparity between the majority and minority classes.

Traditional imbalance-handling techniques rely on \emph{class rebalancing (CR)} such as class reweighting and resampling \citep{chawla2002smote, cui2019class}, which works well for non-graph data. 
Recent studies propose more graph-specific CR strategies tailored for CIGL, e.g., neighborhood-aware reweighting \citep{li2022curriculum,huang2022hardweight} and oversampling \citep{zhao2021graphsmote,park2021graphens}.
Nonetheless, these works are restricted to the class-rebalancing paradigm.
Parallel to class imbalance, another emerging line of research studies topology imbalance, characterized by ``the asymmetric topological properties of the \textit{labeled nodes}''~\citep{chen2021renode}.
It is considered an orthogonal problem to class imbalance, and hence, few work theoretically investigates how topological structure affects the learning on class imbalanced graphs.
To fill this gap, we conduct an in-depth analysis of the role that topology plays in class-imbalanced graph learning.
We theoretically show that topological differences between minority and majority classes significantly amplify the class imbalance bias, imposing great challenges to CIGL.
This reveal an unexplored avenue that limits the performance of existing CIGL techniques: mitigating \textit{class imbalance bias arising from imbalanced topology structure} through topological operations.
Following this novel perspective, we devise a lightweight practical solution for CIGL that can be seamlessly combined with and further boost existing CR techniques.

In this work, we formally define and theoretically investigate two fundamental \textit{local topological phenomena} that greatly hinder CIGL:
\textbf{(i)} \textit{ambivalent message-passing} (AMP), i.e., high ratio of non-self-class neighbors in the node receptive field, and
\textbf{(ii)} \textit{distant message-passing} (DMP), i.e., poor connectivity with self-class labeled nodes.
Intuitively, AMP leads to a higher influx of noisy information and DMP leads to poor reception of effective supervision signals in message-passing.
Both result in lower signal-to-noise ratios and thus induce higher classification errors.
Our theoretical finding reveals that the minority class is inherently more susceptible to both AMP and DMP (Theorem~\ref{the:amp}~\&~\ref{the:dmp}), which leads to a more pronounced predictive bias.
Such bias induced by the graph topology escalates as the level of class imbalance increases.
We emphasize that AMP/DMP is defined for all nodes based on the local neighborhood, while influence conflict has no formal definition and influence insufficiency is defined only on labeled nodes with global PageRank score \citep{chen2021renode}.
To distinguish from them, we use the terminology AMP/DMP instead.
Further discussions can be found in Section~\ref{sec:related-works}.
Fig.~\ref{fig:intro} visually illustrates the concepts of AMP and DMP, highlighting their distinct impacts on the predictive performance of majority and minority classes.

Following our theoretical and empirical findings, we devise \alg{} (\algfull{}), a model-agnostic and efficient technique to mitigate class imbalance bias in CIGL via topological augmentation.
\alg{} dynamically locates and rectifies nodes critically influenced by AMP and DMP during learning, thereby effectively reducing the errors and biases in CIGL.
Being orthogonal to class rebalancing, our solution is able to work hand-in-hand with existing techniques based on reweighting~\citep{japkowicz2002class,chen2021renode} and resampling~\citep{zhao2021graphsmote,park2021graphens} and further boost their performance.
Systematic experiments on real-world CIGL tasks show that \alg{} delivers significant performance boost (up to 46.27\%) and bias reduction (up to 72.74\%) over various CIGL baselines with diverse GNN architectures.

Our contributions:
\textbf{(i) Novel Perspective.}
We demonstrate the feasibility of taming class-imbalance bias \emph{without} class rebalancing, which provides a new avenue that is orthogonal to the predominant class-rebalancing practice in CIGL.
\textbf{(ii) Theoretical Insights.}
We theoretically reveal the topological difference between minority and majority classes and its role in shaping predictive bias in CIGL, shedding light on future CIGL research.
Empirical results validate our findings.
\textbf{(iii) Practical Solution.} 
Motivated by theoretical and empirical finding, we devise a \textit{lightweight and versatile} framework \alg{} to handle topological challenges in CIGL. 
Being complementary to class rebalancing, it can be seamlessly combined with and significantly boost existing CIGL techniques.
\textbf{(iv) Empirical Study.}
Systematic experiments and analysis across a diverse range of real-world tasks and GNN architectures show that \alg{} consistently demonstrates superior performance in both \textit{promoting classification} and \textit{mitigating predictive bias}.

%% file: sections/2-analysis.tex
\section{Class Imbalance and Local Topology}
\label{sec:analysis}

\begin{figure*}[t]
  \hspace{5pt}
  \subfigure[
    Distribution of node AMP/DMP coefficients.
  ]{
    \label{fig:analysis-distr}
    \centering
    \includegraphics[width=0.41\linewidth]{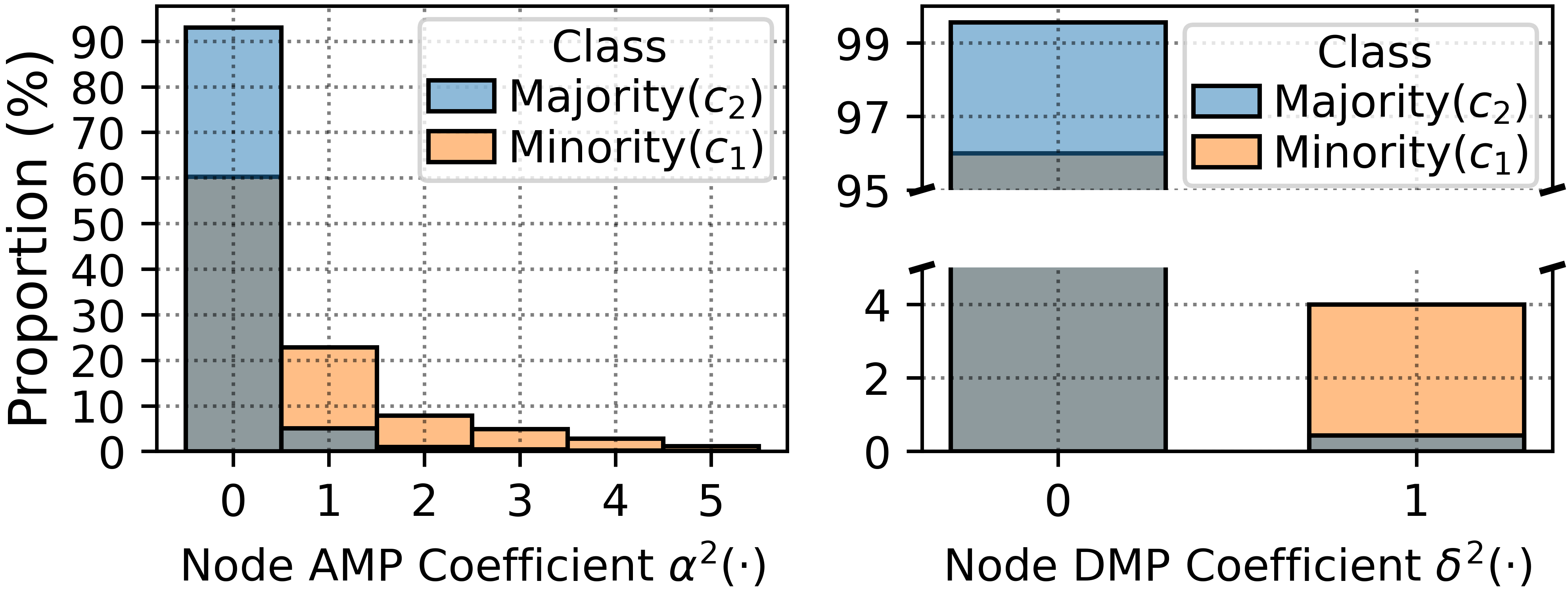}
  }
  \hspace{45pt}
  \subfigure[
    Impact of AMP/DMP on predictive performance.
  ]{
    \label{fig:analysis-acc}
    \centering
    \includegraphics[width=0.41\linewidth]{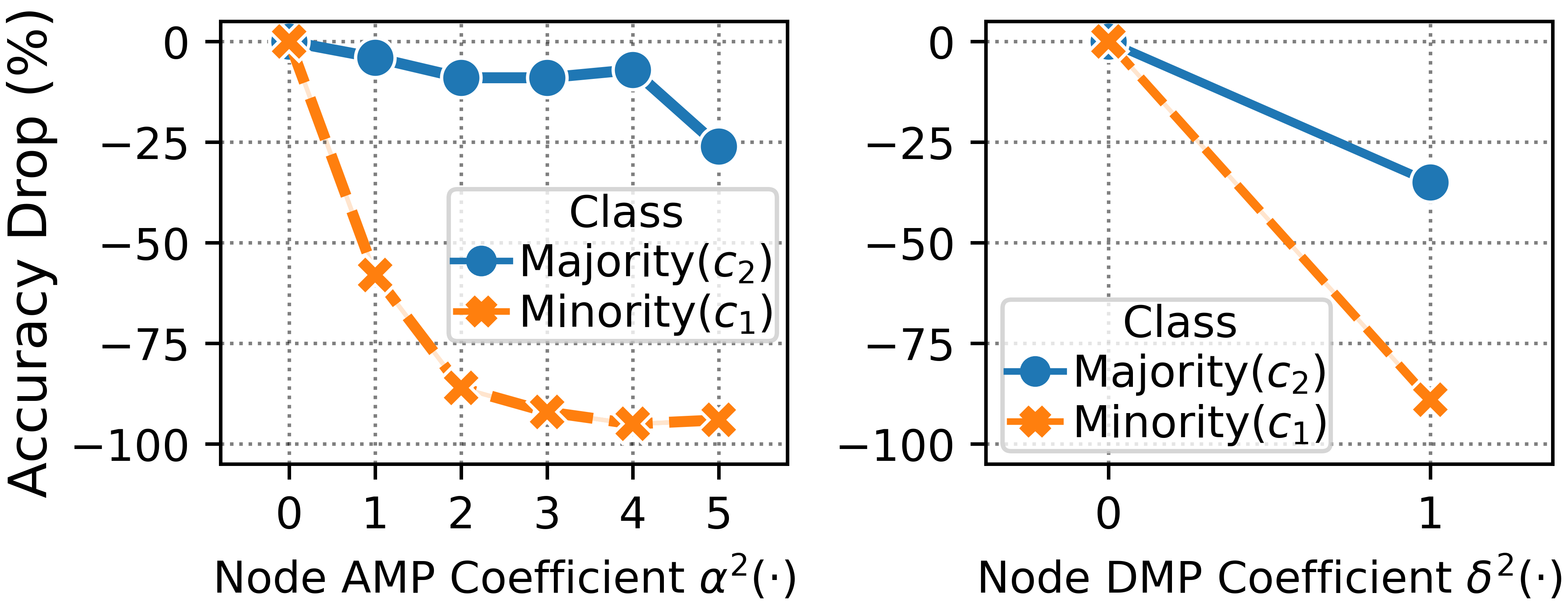}
  }
  \caption{
    Node-level distribution of AMP and DMP coefficients and their impact on learning.
  }
  \label{fig:analysis}
\end{figure*}

In this section, we delve into the impact of graph topology on the predictive bias in class-imbalanced node classification. 
We theoretically unveil that compared to the majority class, the minority class is inherently more susceptible to both Ambivalent Message Passing (AMP) and Distant Message Passing (DMP).
This significantly worsens minority-class performance and leads to a more pronounced predictive bias stemming from class imbalance.
After that, we present an empirical analysis to validate our theoretical findings, and to provide insights on how to mitigate the bias induced by AMP and DMP in practice.
Detailed proofs can be found in Appendix~\ref{sec:apd-proof}.

\textbf{Theoretical analysis on local topology.}
Consider a graph $\gG: (\gV, \gE)$ from a stochastic block model~\citep{holland1983stochastic} $\text{SBM}(n, p, q)$, where $n$ is the total number of nodes, $p$ and $q$ are the intra- and inter-class node connection probability.
To facilitate analysis, we call node $u$ \emph{homo-connected} to node $v$ if there is a path $[u, v_1, ..., v_k, v]$ where $v_1, ..., v_k, v$ are of the same class, and let $\gH(u, k)$ denote the set of $k$-hop homo-connected neighbors of $u$.
For binary node classification, we denote the number of nodes of class $i$ as $n_i$ ($n_1+n_2=n$); without loss of generality, let class $1$/$2$ be the minority/majority class (thus $n_1 \ll n_2$).
We denote class $i$'s node set as $\gV_i$ and labeled node set as $\gV^\textnormal{L}_i$ ($\subset \gV_i$). 
For asymptotic analysis, we adopt conventional assumptions:
$n_1\cdot p=\beta+\gO\big(\frac1n\big)$ (i.e., $\beta$ is the average intra-class node degree of class 1); $p/q=\gO(1)$ \citep{decelle2011asymptotic}. 

We now give formal definitions of AMP and DMP.
For a node $u$ from class $i$, we define its
\textbf{(i)} \textit{$k$-hop AMP coefficient $\alpha^k(u) \in [0, \infty)$} as the ratio of the expected number of non-self-class nodes to self-class nodes in its $k$-hop neighborhood $\gH(u, k)$, i.e., $\alpha^k(u) := \frac{|\{v | v \notin \gV_i, v \in \gH(u, k) \}|}{|\{v | v \in \gV_i, v \in \gH(u, k) \}|}$; 
\textbf{(ii)} \textit{$k$-hop DMP coefficient $\delta^k(u) \in \{0, 1\}$} as the indicator of whether all labeled nodes in its $k$-hop neighborhood are NON-self-class, i.e., $\delta^k(u) := \mathds{1}(L^k_i(u) = 0, \Sigma_{j} L^k_j(u)> 0), \text{ where } L^k_j(u) = |\{v | v \in \gV^\textnormal{L}_j, v \in \gH(u, k)\}|$.
For an intuitive example, the target node (marked by the dashed box) in Fig.~\ref{fig:intro-conf} has $\alpha^1(u) = 3/1, \delta^1(u) = 0$ and node in Fig.~\ref{fig:intro-insu} has $\alpha^1(u) = 1/1, \delta^1(u) = 1$.
Further, to characterize the level of AMP/DMP for different \textit{class}, for class $i$ we define $\alpha^k_i := \frac{\Exp_{u\in \gV_i}[|\{v | v \notin \gV_i, v \in \gH(u, k) \}|]}{\Exp_{u\in \gV_i}[|\{v | v \in \gV_i, v \in \gH(u, k) \}|]}$ and $\delta^k_i := \Prb(\delta^k(u)=1)$, where $u$ is a node of class $i$.
Intuitively, a higher $\alpha_i$ or $\delta_i$ indicates that class $i$ is more susceptible to AMP or DMP.
Building on these metrics, we analyze the disparities in $\alpha$ and $\delta$ between minority and majority classes, thereby providing insights into how the \textit{graph topology induces additional class-imbalance bias}.

\vspace{-3pt}
To simplify notation, we define imbalance ratio $\rho := n_2/n_1$.
The larger the $\rho$ is, the more imbalanced the dataset is.
Then for AMP, we have the following Theorem \ref{the:amp}.

\begin{theorem}[AMP-sourced bias]
\label{the:amp}
For a large $n$, the ratio of AMP coefficients $\alpha$ for the minority class to the majority class grows polynomially with the imbalance ratio $\rho$ and exponentially with $k$:
\vspace{-5pt}
\begin{equation}
    \frac{\alpha_1^k}{\alpha_2^k} = \bigg(\rho\cdot\frac{\sum_{t=1}^k(\rho\beta)^{t-1}}{\sum_{t=1}^k\beta^{t-1}}\bigg)^{\!2} + \gO\Big(\frac1n\Big).
\end{equation}\vspace{-20pt}
\end{theorem}
\begin{proof}
Please see Appendix~\ref{sec:apd-proof-amp}.\qedhere
\end{proof}
% \vspace*{-12pt}

Theorem \ref{the:amp} shows that the same-class neighbor proportion of minority-class nodes is significantly smaller than that of majority-class nodes, i.e., the minority class is more susceptible to AMP.
As the imbalance ratio $\rho$ increases, this issue becomes even more pronounced and introduces a higher bias into the learning process.
Moving on to DMP, we have the following theorem \ref{the:dmp}.

\begin{theorem}[DMP-sourced bias]
\label{the:dmp}
Let $r^\textnormal{L}_i := \frac{|\gV^\textnormal{L}_i|}{|\gV_i|}$ denote the label rate of class $i$. For a large $n$, the ratio of DMP coefficients $\delta$ of the minority class over the majority class grows exponentially with $\rho$:
\vspace{-5pt}
\begin{equation}
    \frac{\delta_1^k}{\delta_2^k}
    \approx \frac{1 - r^\textnormal{L}_1}{1 - r^\textnormal{L}_2} \e^{(\rho - 1)\beta} + \gO\Big(\frac1n\Big).
\end{equation}\vspace{-20pt}
\end{theorem}
\begin{proof}
Please see Appendix~\ref{sec:apd-proof-dmp}.\qedhere
\end{proof}
% \vspace*{-12pt}

Similarly, the result shows that the minority class exhibits a significantly higher susceptibility to DMP than the majority class.
Theorem \ref{the:dmp} also has several interesting implications:
\textbf{(i)} \textit{The imbalance ratio greatly affects the bias induced by DMP,} as $\delta^k_1/\delta^k_2$ grows exponentially with $\rho$.
\textbf{(ii)} \textit{Labeling more minority-class nodes can mitigate, but hardly solve the problem.}
Enlarging the minority-class label rate $r^\textnormal{L}_1$ can linearly shrink $\delta_1^k/\delta_2^k$, but it can hardly eliminate the bias induced by DMP (i.e., to have $\delta_1^k \leq \delta_2^k$) as $\e^{(\rho - 1)\beta}$ is usually very large in practice.
Take the \textit{Cora} dataset~\citep{sen2008planetoid} as an example (let class $1$/class $2$ denote the smallest/largest class): eliminating the DMP bias requires the minority-class label rate $r^\textnormal{L}_1 \geq 1 - \frac{1 - r^\textnormal{L}_2}{\e^{(\rho - 1)\beta}} > 1 - 5.05\times 10^{-8}$, which is practically infeasible.

Our theoretical findings show that both AMP and DMP affect the minority and majority classes differently, and the difference is primarily determined by the imbalance ratio $\rho$.
However, directly manipulating $\rho$ is tricky in practice as it requires sampling new nodes and edges from an unknown underlying graph generation model, or at least, simulating the process by oversampling.

\begin{figure*}[t]
  \centering
  \includegraphics[width=0.98\linewidth]{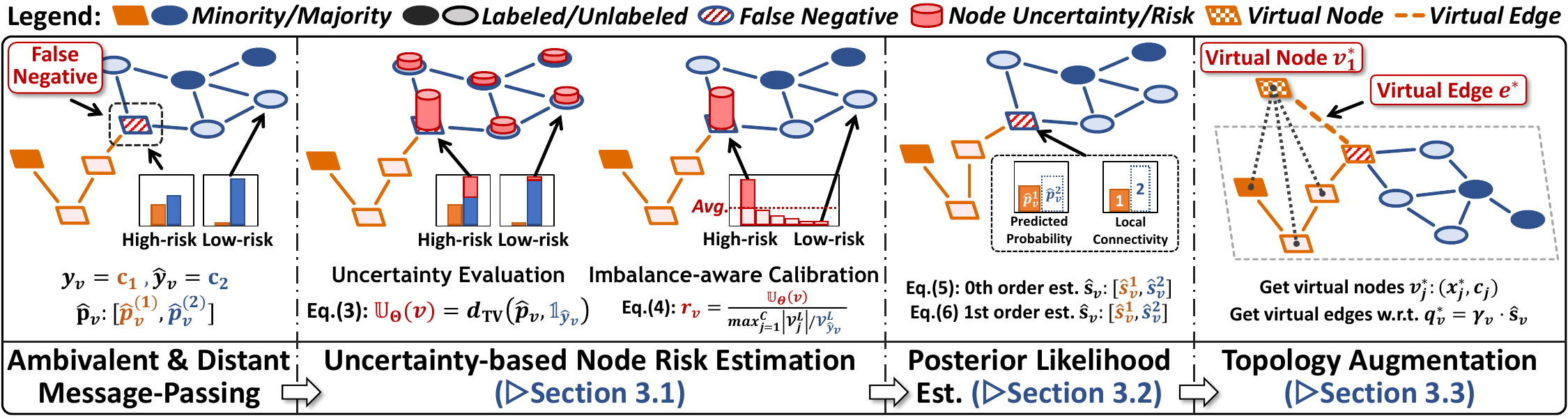}
  \caption{
  The proposed \alg{} (\algfull{}) framework, best viewed in color.
  }
  \label{fig:method}
\end{figure*}

\textbf{A closer look at AMP \& DMP in practice.}
To verify the theoretical results, and to provide more insights on how to mitigate the bias brought about by AMP and DMP in practice, we conduct a fine-grained empirical analysis on a real-world task. 
Results are detailed in Fig.~\ref{fig:analysis}\footnote{Results obtained by training a GCN on the \textit{PubMed} dataset.}.
Starting from Fig.~\ref{fig:analysis-distr}, we can first observe that the minority class $1$ has a larger proportion of nodes with high $\alpha$ or $\delta$ than the majority class $2$, i.e., minority class $1$ has higher average $\alpha$ and $\delta$ (specifically, $\alpha_1/\alpha_2=1.357/0.179$, $\delta_1/\delta_2=0.040/0.004$), which is consistent with our theoretical findings.
Further, Fig.~\ref{fig:analysis-acc} shows that both AMP and DMP significantly reduce the prediction accuracy, especially for the minority class that inherently has poorer data representation.
This can be explained from a graph signal denoising perspective~\citep{nt2019revisiting}: AMP introduces additional noise from dissimilar nodes, and DMP leads to less efficient label propagation/denoising, thus their impact is particularly significant on minority classes that are more susceptible to noise~\citep{johnson2019survey-deepimb} due to poor representation in the feature space.
We further notice an intriguing fact that, at sample-level, the impact of AMP/DMP is concentrated on a small fraction of minority class nodes with large $\alpha$ or $\delta$ (e.g., the $\alpha \geq 1$ / $\delta = 1$ part in Fig.~\ref{fig:analysis-distr}).
In other words, one can surrogate the tricky manipulation of $\rho$ and directly mitigate AMP/DMP by \textit{locating and rectifying a small number of critical nodes}, and this exactly motivates our subsequent studies.

%% file: sections/3-methodology.tex
\section{Handling Class Imbalance from a Topological Perspective}
\label{sec:methodology}

Armed with the findings from Section~\ref{sec:analysis}, we now discuss how to devise a practical strategy to mitigate the error and bias induced by graph topology in CIGL.
Earlier analyses have shown that this can be achieved by identifying and rectifying the critical nodes that are highly influenced by AMP/DMP.
This naturally poses two challenging questions: 
\textbf{(i)} How can critical nodes be located as the direct calculation of $\alpha$/$\delta$ using ground-truth labels is not possible?
\textbf{(ii)} Subsequently, how can critical nodes be rectified and minimize the negative impact caused by AMP and DMP?

In answering the above questions, we devise a \textit{lightweight} framework \alg{} (\algfull{}) for handling the topology-sourced errors and biases in CIGL.
Specifically, for locating the misclassified nodes, \alg{} leverages model-based prediction uncertainty to assess the risk of potential misclassification caused by AMP/DMP for each node (\S~\ref{sec:node-risk}).
Then to rectify a misclassified node, we estimate a posterior likelihood of each node being in each class (\S~\ref{sec:node-similarity}) and dynamically augment the misclassified node's topological context based on our risk scores and posterior likelihoods (\S~\ref{sec:virtual-node-edge}) thereby
mitigating the impact of AMP and DMP.
An overview of the proposed \alg{} framework is shown in Fig.~\ref{fig:method}.
\subsection{Node Misclassification Risk Estimation}
\label{sec:node-risk}

We now elaborate on the technical details of \alg{}.
As discussed earlier, our first step is to locate the critical nodes that are highly influenced by AMP/DMP. Given the unavailability of ground-truth labels, direct computation of AMP/DMP coefficient is infeasible in practice.
Fortunately, recent studies have shown that conflicting or lack of information from the neighborhood can disturb GNN learning and the associated graph-denoising process for affected nodes~\citep{nt2019revisiting,wu2019simplifying,ma2021denoising}.
This further yields high vacuity or dissonance uncertainty~\citep{stadler2021uncertainty,zhao2020uncertainty} in the prediction.
This motivates us to exploit the \textit{model prediction uncertainty} to estimate nodes' risk of being misclassified due to AMP/DMP.

\noindent\textbf{Uncertainty quantification.}
While there exist many techniques for uncertainty quantification (e.g., Bayesian-based~\citep{zhang2019bayesian,hasanzadeh2020bayesian}, Jackknife sampling~\citep{kang2022jurygcn}), they often either have to modify the model architecture, and/or impose additional computational overhead.
In this study, we aim to streamline the design of \alg{} for optimal efficiency and adaptability.
To this end, we employ an efficient and highly effective approach to uncertainty quantification.
Formally, let $C$ be the number of classes. For a node $v$, consider model $F(\cdot;\Theta)$'s predicted probability vector $\hat{\vp}_v=F(\mA, \mX;\Theta)_v$, i.e., $\hat{p}_v^{(j)}=\Prb(y_v=j|\mA, \mX, \Theta)$.
Let $\hat{y}_v$ be the predicted label. We measure the uncertainty score $\sU_\Theta(v)$ by the total variation (TV) distance:
\vspace{-5pt}
\begin{equation}\textstyle
\label{eq:node-unc}
\resizebox{0.9\linewidth}{!}{$
\sU_\Theta(v) := d_\text{TV}(\hat{\vp}_v,\mathds{1}_{\hat{y}_v}) = \frac{1}{2}\sum_{j=1}^{C} |\hat{\vp}_v^{(j)}-\mathds{1}_{\hat{y}_v}^{(j)}|\in[0,1].
$}
\end{equation}
Intuitively, a node has higher uncertainty if the model is less confident about its current prediction.
We remark that this metric can be naturally replaced by other uncertainty measures (e.g., information entropy or more complex ones) with additional computation cost, yet the impact on performance is marginal.
Please refer to the ablation study provided in Appendix~\ref{sec:apd-dis-ablation} for more details.

\noindent\textbf{Imbalance-calibrated misclassification risk.}
Due to the lack of training instances, minority classes generally exhibit higher uncertainty.
Therefore, using $\sU_\Theta(\cdot)$ directly as the risk score would treat most minority-class nodes as high-risk, which is contrary to our intention of rectifying the false negatives (i.e., minority nodes wrongly predicted as majority-class) that cause bias in CIGL.
To cope with this, we propose \textit{imbalance-aware calibration} for misclassification risk scores.
For each class $i$, let $\hat{\gV}_{i} := \{u\in\gV | \hat{y}_u = i\}$ and $\hat{\gV}^\textnormal{L}_{i} := \{u\in\gV^\textnormal{L} | y_u = i\}$. For node $v$ with predicted label $\hat{y}_v$, we define its risk $r_v$ as:
% \vspace{-5pt}
\begin{equation}\textstyle
r_v := \frac{\sU_\Theta(v)}{\max_{j=1}^C|\gV^\textnormal{L}_{j}|/|\gV^\textnormal{L}_{\hat y_v}|}\in[0,1].
\label{eq:node-risk}
\end{equation}
Intuitively speaking, Eq.~(\ref{eq:node-risk}) calibrates $v$'s prediction uncertainty by a \textit{label imbalance score} $\max_{j=1}^C|\gV^\textnormal{L}_{j}|/|\gV^\textnormal{L}_{\hat y_v}|$.
Minority classes with smaller labeled sets $\gV^\textnormal{L}_{i}$ will be discounted more.

\begin{figure}[H]
\centering
  \parbox{.3\linewidth}{
  \includegraphics[width=\linewidth]{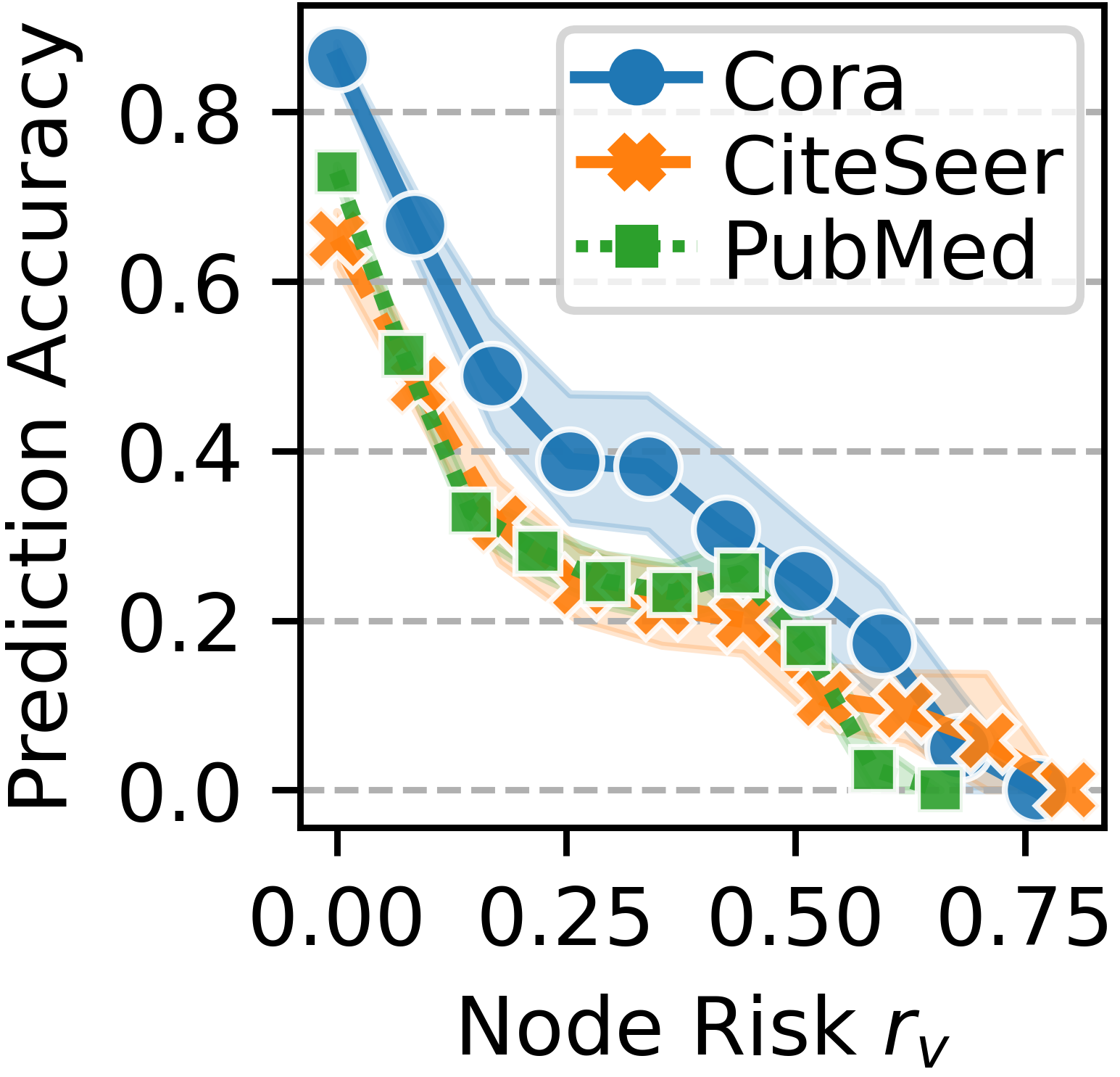}
  }
  \hspace{\fill}
  \parbox{.65\linewidth}{
    \vspace*{-15pt}
    \caption{
    The negative correlation between the estimated node risk (x-axis) and the prediction accuracy (y-axis).
    We apply 10 sliding windows to compute the mean and deviation of the accuracy.
    }
    \label{fig:risk-acc}
}
\end{figure}

\noindent\textbf{Empirical validation.}
We validate the effectiveness of the proposed node risk assessment method, as shown in Fig.~\ref{fig:risk-acc}.
The results indicate that our approach can accurately estimate node misclassification risk across various real-world CIGL tasks while enjoying computational efficiency.
\vspace{-15pt}

\subsection{Posterior Likelihood Estimation}
\label{sec:node-similarity}

With the estimated risk scores of being affected by AMP/DMP, we move to the next question: how to rectify high-risk nodes with topological augmentation?
As high-risk nodes are prone to misclassification, their true labels are more likely to be among the non-predicted classes $j \neq \hat{y}_v$. 
This motivates us to investigate schemes to harness information from these non-predicted classes. 
Since uniformly drawing from all classes probably introduces noise to learning, we propose to estimate the \emph{posterior likelihood} $\hat{s}_v^{(j)}$ that a high-risk node $v$ belongs to each class $j$ after observing the current predictions.
To estimate $\hat s_v^{(j)}$, we introduce a \emph{zeroth-order} scheme and a \emph{first-order} scheme with $\gO(|\gV|C)$ and $\gO(|\gE|C)$ time complexity, respectively. 
Please refer to \S\ref{sec:experiments} for a practical complexity analysis.
We do not employ higher-order schemes due to the $\gO(|\gV|^{k-1}|\gE|C)$ time complexity of the $k$th-order scheme.

\noindent\textbf{Zeroth-order estimation.}
A natural approach is to utilize the predicted probabilities $\hat p_v^{(j)}$. 
As we have shown, the predicted label $\hat y_v$ of a high-risk node $v$ is very likely to be wrong. Thus, we define the posterior likelihood $\hat s_v^{(j)}$ as the conditional probability given that the class is not $\hat y_v$, i.e.,
\vspace{-5pt}
\begin{equation}
\label{eq:0th-order}
\begin{aligned}
    \hat s_v^{(j)} :={}& \Prb_{y\sim\hat\vp_v}[y=j|y\ne\hat y_v] \\
    ={}& \begin{cases}
    \hat{p}_v^{(j)}/(1-\hat p_v^{(\hat y_v)}),&\text{if }j\ne\hat y_v,\\0,&\text{otherwise}.
    \end{cases}
\end{aligned}
\end{equation}

Intuitively, the zeroth-order posterior likelihoods $\hat s_v^{(j)}$ are consistent with the predicted probabilities $\hat p_v^{(j)}$ except for the wrongly predicted label $j=\hat y_v$. This can be computed efficiently on GPU in matrix form.

\noindent\textbf{First-order estimation via random walk.}
We further explore the \textit{local topology} for $\hat s_v^{(j)}$ estimation.
Since neighboring nodes on a homophily graph tend to share labels, we consider a 1-step random walk starting from node $v$. Let $\gN(v)$ be the neighboring node set of $v$, and let $v'\sim\gN(v)$ denote the ending node of the random walk. We define $\hat s_v^{(j)}$ as the conditional probability that $v'$ is predicted as class $j$ given that $v'$ is not predicted as class $\hat y_v$, i.e.,
\begin{equation}
\label{eq:1st-order}
\begin{aligned}
    \hat s_v^{(j)} :={}& \Prb_{v'\sim\gN(v)}[\hat y_{v'}=j|\hat y_{v'}\ne\hat y_v] \\
    ={}&\begin{cases}
    \frac{|\{v'\in\gN(v) | \hat{y}_{v'} = j\}|}{|\gN(v)| - |\{v'\in\gN(v) | \hat{y}_{v'} = \hat{y}_v\}|},&\text{if }j\ne\hat y_v,\\0,&\text{otherwise}.
    \end{cases}
\end{aligned}
\end{equation}

With first-order estimation, $\hat s_v^{(j)}$ is proportional to the label frequency among adjacent nodes.
Different from the zeroth-order scheme, this scheme relies on both node-level predictions and local connectivity patterns.
The computation can be done via sparse matrix operation with $\gO(|\gE|C)$ time complexity. 
As a remark, although this scheme can extend to $k$-step random walks, we do not employ them due to the $\gO(|\gV|^{k-1}|\gE|C)$ complexity of exact computation and the high variance of stochastic computation.

\begin{figure}[H]
  \parbox{.32\linewidth}{
  \includegraphics[width=1\linewidth]{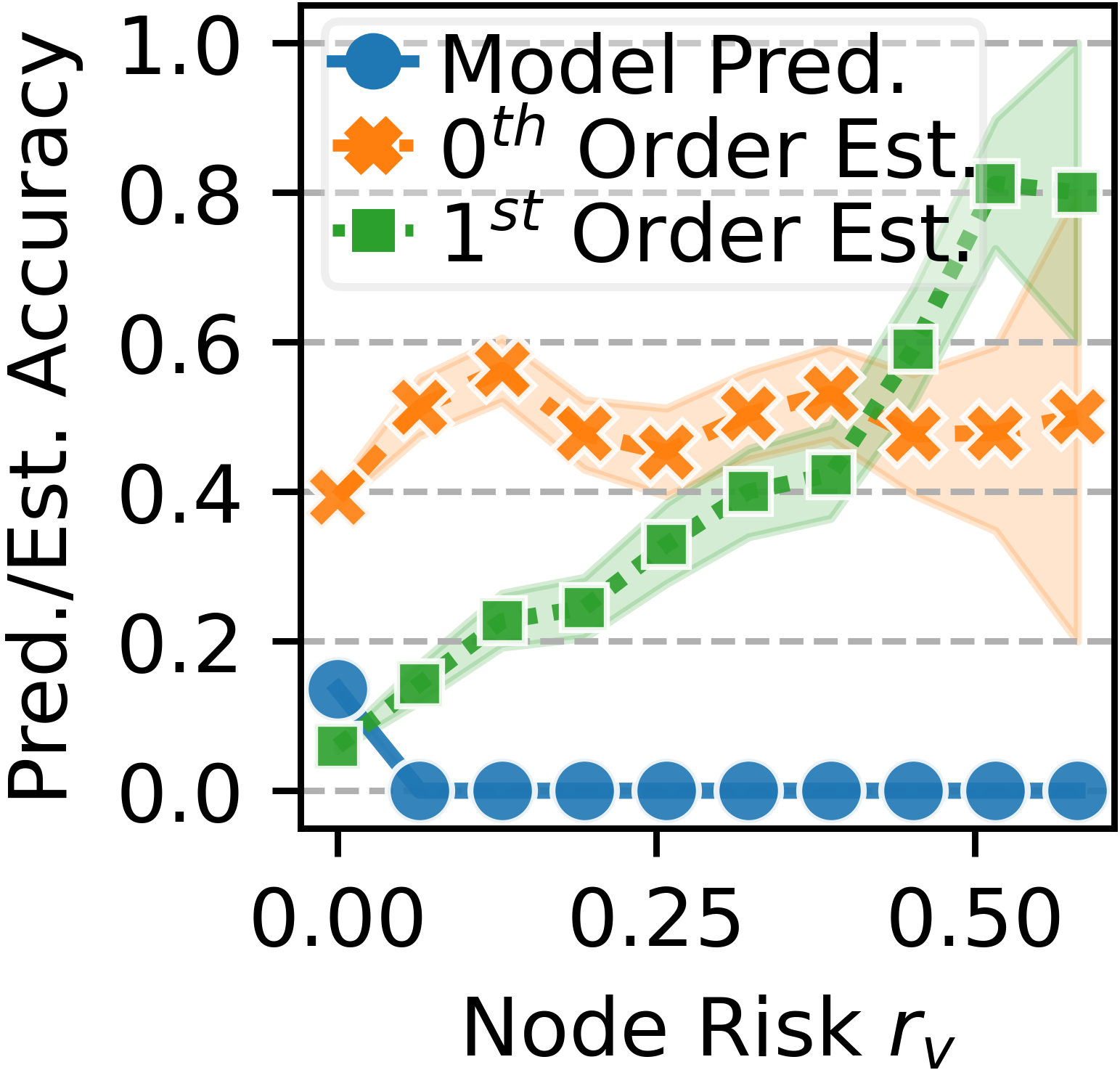}
  }
  \hspace{\fill}
  \parbox{.65\linewidth}{
    \vspace*{-10pt}
    \caption{
    The minority-class accuracy of model prediction $\hat{y}_v = F(\mA, \mX; \Theta)$, and max-likelihood-based candidate selection $\hat{y}^s_v = {\argmax(\hat{\vs}_v)}$, on PubMed dataset.
    Note that this is just an illustrative example using ${\argmax(\hat{\vs}_v)}$. In practice, we consider the whole $\hat{\vs}_v$ when sampling virtual edges, as described in Section~\ref{sec:virtual-node-edge}.
    }
    \label{fig:cand-acc}
}
\end{figure}

\noindent\textbf{Empirical validation.}
Figure~\ref{fig:cand-acc} compares the two schemes in practice.
Results show that all high-risk ($r_v>0$) minority nodes are misclassified, and both schemes can effectively find alternatives with significantly higher chances to be the ground truth class for high-risk nodes.
\vspace{-15pt}

\subsection{Virtual Topology Augmentation}
\label{sec:virtual-node-edge}
Finally, we discuss how to mitigate AMP and DMP via topology augmentation using our node risk scores and the posterior likelihoods. The general idea is to augment the local topology of high-risk nodes so as to integrate information from nodes that share similar patterns (mitigate AMP), even if they are not closely adjacent to each other in the graph topology (mitigate DMP), thus achieving less-biased CIGL.
A straightforward way is to connect high-risk nodes to nodes from high-likelihood classes in the original graph.
However, this can be problematic in practice as a massive number of possible edges could be generated, greatly disturbing the original topology structure.

To achieve efficient augmentation without disrupting the graph topology, we create virtual nodes (one per class) as ``shortcuts'' connecting to high-risk nodes according to posterior likelihoods.
These shortcuts aggregate and pass class information to high-risk nodes from nodes that exhibit similar patterns (even if they are distant in the original graph), thus mitigating both AMP and DMP.
Formally, for each class $j$, we build a virtual node $v^*_j$ with feature $\vx_{v_j^*}:=\sum_{v \in \hat{\gV}_{j}} \vx_v/|\hat{\gV}_{j}|$ and label $y_{v_j^*}:=j$, and compute the average risk $\bar r_j:=\sum_{v\in\hat{\gV}_{j}}r_v/|\hat{\gV}_{j}|$. 
Then for each node $v$, we connect a virtual edge between $v$ and virtual node $v^*_j$ with probability proportional to the posterior likelihood $\hat s_v^{(j)}$. 
However, if the connection probability is exactly $\hat s_v^{(j)}$, there will be many unnecessary virtual edges for low-risk nodes. 
Hence, we introduce a discount factor $\gamma_v$ based on risk scores and connect the virtual edge with probability $q_v^{(j)}:=\gamma_v\hat s_v^{(j)}$. 
To design the optimal $\gamma_v$, we propose to solve the following constrained quadratic program:
\begin{equation}
\min_{\boldsymbol\gamma\ge\boldsymbol0}\bigg({-\sum_{v\in\gV}(r_v-\bar r_{\hat y_v})\gamma_v+\frac12\|\boldsymbol\gamma\|_2^2}\bigg),
\label{eq:virlink}
\end{equation}
where the first term encourages virtual edges for high-risk nodes, and the second term is to minimize the number of virtual edges. 
The closed-form solution is $\gamma_v=\max(r_v-\bar r_{\hat y_v},0)$ \citep{antoniadis2001regularization}, which avoids virtual edges for low-risk nodes as we desire.
We now summarize the procedure of \alg{} in Algorithm~\ref{alg:main}.

\begin{algorithm}[t]
    \caption{\alg{}: Topological Balanced Augmentation}
    \label{alg:main}
\begin{algorithmic}[1]
    \REQUIRE Class-imbalanced graph $\gG: \{\mA, \mX\}$;
    \STATE \textbf{Initialize:} node classifier $F(\cdot;\Theta)$;
    \WHILE{not converged}
        \STATE $\hat{\mP} \gets F(\mA, \mX;\Theta)$; 
        \STATE $\hat{\vy} \gets \text{argmax}_{axis=1}(\hat{\mP})$; \hfill $\vartriangleright$ Model predictions $\hat{\vy}$
        \STATE $\vr \gets \texttt{NodeRiskEst}(\hat{\mP}$, $\hat{\vy})$; \hfill $\vartriangleright$ Eq. (\ref{eq:node-unc}) - (\ref{eq:node-risk})
        \STATE $\hat{\mS} \gets \texttt{PosteriorEst}(\mA, \hat{\mP}, \hat{\vy})$; \hfill $\vartriangleright$ Eq. (\ref{eq:0th-order}) - (\ref{eq:1st-order})
        \FOR{class $j=1$ to $C$} 
            \STATE $\vx_{v_j^*}\gets\sum_{v \in \hat{\gV}_{j}} \vx_v/|\hat{\gV}_{j}|$;
            \STATE $v^*_j:(\vx_{v_j^*}, j)$  \hfill $\vartriangleright$ Virtual node $v^*_j$ for class $j$
        \ENDFOR
        \STATE $\gV^* \gets \{v^*_j | 1 \le j \le C\}$ \hfill $\vartriangleright$ Virtual node set $\gV^*$
        \STATE $\mQ^* \gets \hat{\mS} \odot \bm{\gamma}$ \hfill $\vartriangleright$ Virtual link prob. $\mQ^*$ by Eq. (\ref{eq:virlink})
        \STATE $\gE^* \sim \mQ^*$; \hfill $\vartriangleright$ Sample virtual edges $\gE^*$ w.r.t $\mQ^*$
        \STATE Derive $\mX^*, \mA^*$ from $\gV \cup \gV^*, \gE \cup \gE^*$;
        \STATE Update $\Theta$ with augmented graph $\gG^*: \{\mA^*, \mX^*\}$;
    \ENDWHILE
    \STATE \textbf{Return:} a balanced node classifier $F(\mA, \mX; \Theta)$;
\end{algorithmic}
\end{algorithm}

\textbf{Complexity Analysis of \alg{}.}
\alg{} with 0th/1st-order estimation \textbf{scales linearly} with the number of nodes/edges (i.e., with $\gO(|\gV|C)$ or $\gO(|\gE|C)$ complexity).
This makes \alg{} highly efficient and allows dynamically graph augmentation in each training step.
Specifically, \alg{} introduces $C$ (the number of class, usally small) virtual nodes with $\mathcal{O}(n)$ edges.
Because of the long-tail distribution of node uncertainty and the discount factor used to solve Eq.~(\ref{eq:virlink}), only a small portion of nodes have positive risks with relatively few (empirically around 1-3\%) virtual edges introduced.
We provide the scalability results of \alg{} later in the experiment section. 
In short, \alg{} takes milliseconds for a single topological augmentation. 
Please refer to Table~\ref{tab:runtime} and the corresponding discussion for further details.
We also discuss how to further speedup \alg{} in practice in \ref{sec:apd-dis-speedup}.

\input{sections/tables/tab_main_icml.tex}

%% file: sections/tables/tab_main_icml.tex
\begingroup
\setlength{\tabcolsep}{5pt}
\begin{table*}[t]
\scriptsize
\caption{
    \alg{} significantly boosts existing CIGL techniques, achieving better classification performance (Balanced Acc./Marco-F1) with reduced bias (PerfStd).
    For each CIGL baseline, we report its performance before and after collaborating with \alg{}. 
    The average and the best score of all CIGL methods under three settings (base, +\algp{}, +\algt{}) are also provided, with \textbf{performance gain of \alg{} highlighted by $\Delta$}.
    Due to space limitation, we report the key results and omit the error bar, full results can be found in Appendix~\ref{sec:apd-results}.
}
\label{tab:main}
\begin{threeparttable}
\resizebox{\textwidth}{!}{%
\begin{tabular}{ccc|cccccccll|ll|ll}
\toprule
\multicolumn{3}{c|}{\textbf{Metric}} & \multicolumn{9}{c|}{\textbf{Balanced Acc.$\uparrow$}} & \multicolumn{2}{c|}{\textbf{Macro-F1$\uparrow$}} & \multicolumn{2}{c}{\textbf{PerfStd$\downarrow$}} \\ \hline
\multicolumn{3}{c|}{\textbf{CIGL Baseline}} & \tiny{ERM} & \tiny{RW} & \tiny{RN} & \tiny{RS} & \tiny{SM} & \tiny{GS} & \multicolumn{1}{c|}{\tiny{GE}} & \textbf{Avg ($\Delta$)} & \textbf{Best ($\Delta$)} & \textbf{Avg ($\Delta$)} & \textbf{Best ($\Delta$)} & \textbf{Avg ($\Delta$)} & \textbf{Best ($\Delta$)} \\ \hline
\multicolumn{1}{c|}{\multirow{9}{*}{\textbf{\rotatebox{90}{Cora}}}} & \multicolumn{1}{c|}{\multirow{3}{*}{\textbf{\rotatebox{90}{GCN}}}} & \textsc{base} & \tiny{61.6} & \tiny{67.7} & \tiny{66.6} & \tiny{59.5} & \tiny{58.3} & \tiny{68.0} & \multicolumn{1}{c|}{\tiny{70.1}} & 64.5 & 70.1 & 63.7 & 70.0 & 25.7 & 20.0 \\ \cline{3-16} 
\multicolumn{1}{c|}{} & \multicolumn{1}{c|}{} & +\algp{} & \tiny{65.5} & \tiny{71.0} & \tiny{71.4} & \tiny{72.5} & \tiny{72.2} & \tiny{68.5} & \multicolumn{1}{c|}{\tiny{72.2}} & \cellcolor{Gray} \textbf{70.5 (\tiny{+5.9})} & \cellcolor{Gray} \textbf{72.5 (\tiny{+2.4})} & \cellcolor{Gray} \textbf{69.2 (\tiny{+5.5})} & \cellcolor{Gray} \textbf{71.6 (\tiny{+1.7})} & \cellcolor{Gray} \textbf{16.7 (\tiny{-9.0})} & \cellcolor{Gray} \textbf{14.4 (\tiny{-5.6})} \\
\multicolumn{1}{c|}{} & \multicolumn{1}{c|}{} & +\algt{} & \tiny{69.8} & \tiny{72.1} & \tiny{71.8} & \tiny{74.2} & \tiny{73.9} & \tiny{71.6} & \multicolumn{1}{c|}{\tiny{72.6}} & \cellcolor{Gray} \textbf{72.3 (\tiny{+7.8})} & \cellcolor{Gray} \textbf{74.2 (\tiny{+4.1})} & \cellcolor{Gray} \textbf{71.1 (\tiny{+7.5})} & \cellcolor{Gray} \textbf{72.8 (\tiny{+2.9})} & \cellcolor{Gray} \textbf{17.6 (\tiny{-8.0})} & \cellcolor{Gray} \textbf{15.2 (\tiny{-4.8})} \\ \cline{2-16} 
\multicolumn{1}{c|}{} & \multicolumn{1}{c|}{\multirow{3}{*}{\textbf{\rotatebox{90}{GAT}}}} & \textsc{base} & \tiny{61.5} & \tiny{66.9} & \tiny{66.8} & \tiny{57.8} & \tiny{58.8} & \tiny{64.7} & \multicolumn{1}{c|}{\tiny{69.8}} & 63.8 & 69.8 & 63.1 & 70.0 & 26.0 & 20.1 \\ \cline{3-16} 
\multicolumn{1}{c|}{} & \multicolumn{1}{c|}{} & +\algp{} & \tiny{66.3} & \tiny{71.8} & \tiny{72.1} & \tiny{71.9} & \tiny{70.5} & \tiny{69.3} & \multicolumn{1}{c|}{\tiny{70.6}} & \cellcolor{Gray} \textbf{70.4 (\tiny{+6.6})} & \cellcolor{Gray} \textbf{72.1 (\tiny{+2.4})} & \cellcolor{Gray} \textbf{69.0 (\tiny{+6.0})} & \cellcolor{Gray} \textbf{70.9 (\tiny{+0.9})} & \cellcolor{Gray} \textbf{17.2 (\tiny{-8.9})} & \cellcolor{Gray} \textbf{15.1 (\tiny{-5.0})} \\
\multicolumn{1}{c|}{} & \multicolumn{1}{c|}{} & +\algt{} & \tiny{70.1} & \tiny{71.6} & \tiny{70.3} & \tiny{73.3} & \tiny{72.2} & \tiny{71.1} & \multicolumn{1}{c|}{\tiny{71.0}} & \cellcolor{Gray} \textbf{71.4 (\tiny{+7.6})} & \cellcolor{Gray} \textbf{73.3 (\tiny{+3.5})} & \cellcolor{Gray} \textbf{70.2 (\tiny{+7.1})} & \cellcolor{Gray} \textbf{72.3 (\tiny{+2.4})} & \cellcolor{Gray} \textbf{18.0 (\tiny{-8.0})} & \cellcolor{Gray} \textbf{17.3 (\tiny{-2.8})} \\ \cline{2-16} 
\multicolumn{1}{c|}{} & \multicolumn{1}{c|}{\multirow{3}{*}{\textbf{\rotatebox{90}{SAGE}}}} & \textsc{base} & \tiny{59.2} & \tiny{63.8} & \tiny{65.3} & \tiny{57.8} & \tiny{58.8} & \tiny{61.6} & \multicolumn{1}{c|}{\tiny{68.8}} & 62.2 & 68.8 & 60.9 & 68.2 & 27.1 & 19.8 \\ \cline{3-16} 
\multicolumn{1}{c|}{} & \multicolumn{1}{c|}{} & +\algp{} & \tiny{66.2} & \tiny{70.1} & \tiny{71.3} & \tiny{71.2} & \tiny{70.3} & \tiny{69.9} & \multicolumn{1}{c|}{\tiny{69.8}} & \cellcolor{Gray} \textbf{69.8 (\tiny{+7.7})} & \cellcolor{Gray} \textbf{71.3 (\tiny{+2.5})} & \cellcolor{Gray} \textbf{68.9 (\tiny{+8.0})} & \cellcolor{Gray} \textbf{70.4 (\tiny{+2.2})} & \cellcolor{Gray} \textbf{16.4 (\tiny{-10.7})} & \cellcolor{Gray} \textbf{13.3 (\tiny{-6.5})} \\
\multicolumn{1}{c|}{} & \multicolumn{1}{c|}{} & +\algt{} & \tiny{66.5} & \tiny{71.1} & \tiny{71.5} & \tiny{73.0} & \tiny{73.0} & \tiny{72.3} & \multicolumn{1}{c|}{\tiny{71.9}} & \cellcolor{Gray} \textbf{71.4 (\tiny{+9.2})} & \cellcolor{Gray} \textbf{73.0 (\tiny{+4.2})} & \cellcolor{Gray} \textbf{70.1 (\tiny{+9.2})} & \cellcolor{Gray} \textbf{71.7 (\tiny{+3.5})} & \cellcolor{Gray} \textbf{16.6 (\tiny{-10.5})} & \cellcolor{Gray} \textbf{14.9 (\tiny{-4.9})} \\ \hline
\multicolumn{1}{c|}{\multirow{9}{*}{\textbf{\rotatebox{90}{CiteSeer}}}} & \multicolumn{1}{c|}{\multirow{3}{*}{\textbf{\rotatebox{90}{GCN}}}} & \textsc{base} & \tiny{37.6} & \tiny{42.5} & \tiny{42.6} & \tiny{39.2} & \tiny{39.3} & \tiny{45.1} & \multicolumn{1}{c|}{\tiny{56.0}} & 43.2 & 56.0 & 36.1 & 54.5 & 27.0 & 16.9 \\ \cline{3-16}
\multicolumn{1}{c|}{} & \multicolumn{1}{c|}{} & +\algp{} & \tiny{52.7} & \tiny{57.9} & \tiny{57.5} & \tiny{57.9} & \tiny{60.1} & \tiny{57.7} & \multicolumn{1}{c|}{\tiny{60.6}} & \cellcolor{Gray} \textbf{57.8 (\tiny{+14.6})} & \cellcolor{Gray} \textbf{60.6 (\tiny{+4.6})} & \cellcolor{Gray} \textbf{56.9 (\tiny{+20.8})} & \cellcolor{Gray} \textbf{59.9 (\tiny{+5.4})} & \cellcolor{Gray} \textbf{17.6 (\tiny{-9.4})} & \cellcolor{Gray} \textbf{13.8 (\tiny{-3.1})} \\
\multicolumn{1}{c|}{} & \multicolumn{1}{c|}{} & +\algt{} & \tiny{55.4} & \tiny{58.4} & \tiny{59.3} & \tiny{58.8} & \tiny{62.0} & \tiny{57.6} & \multicolumn{1}{c|}{\tiny{62.7}} & \cellcolor{Gray} \textbf{59.2 (\tiny{+16.0})} & \cellcolor{Gray} \textbf{62.7 (\tiny{+6.7})} & \cellcolor{Gray} \textbf{58.4 (\tiny{+22.3})} & \cellcolor{Gray} \textbf{62.5 (\tiny{+8.0})} & \cellcolor{Gray} \textbf{19.3 (\tiny{-7.7})} & \cellcolor{Gray} \textbf{13.9 (\tiny{-3.0})} \\ \cline{2-16} 
\multicolumn{1}{c|}{} & \multicolumn{1}{c|}{\multirow{3}{*}{\textbf{\rotatebox{90}{GAT}}}} & \textsc{base} & \tiny{39.2} & \tiny{41.3} & \tiny{43.2} & \tiny{36.0} & \tiny{37.0} & \tiny{41.8} & \multicolumn{1}{c|}{\tiny{51.5}} & 41.4 & 51.5 & 34.1 & 48.3 & 29.0 & 25.2 \\ \cline{3-16} 
\multicolumn{1}{c|}{} & \multicolumn{1}{c|}{} & +\algp{} & \tiny{55.7} & \tiny{59.3} & \tiny{58.3} & \tiny{60.1} & \tiny{60.6} & \tiny{56.1} & \multicolumn{1}{c|}{\tiny{60.9}} & \cellcolor{Gray} \textbf{58.7 (\tiny{+17.3})} & \cellcolor{Gray} \textbf{60.9 (\tiny{+9.4})} & \cellcolor{Gray} \textbf{58.1 (\tiny{+23.9})} & \cellcolor{Gray} \textbf{60.0 (\tiny{+11.7})} & \cellcolor{Gray} \textbf{16.3 (\tiny{-12.6})} & \cellcolor{Gray} \textbf{10.7 (\tiny{-14.5})} \\
\multicolumn{1}{c|}{} & \multicolumn{1}{c|}{} & +\algt{} & \tiny{60.3} & \tiny{61.2} & \tiny{59.1} & \tiny{60.3} & \tiny{62.4} & \tiny{57.7} & \multicolumn{1}{c|}{\tiny{63.5}} & \cellcolor{Gray} \textbf{60.6 (\tiny{+19.2})} & \cellcolor{Gray} \textbf{63.5 (\tiny{+12.0})} & \cellcolor{Gray} \textbf{59.9 (\tiny{+25.8})} & \cellcolor{Gray} \textbf{62.5 (\tiny{+14.2})} & \cellcolor{Gray} \textbf{17.7 (\tiny{-11.3})} & \cellcolor{Gray} \textbf{13.2 (\tiny{-12.0})} \\ \cline{2-16} 
\multicolumn{1}{c|}{} & \multicolumn{1}{c|}{\multirow{3}{*}{\textbf{\rotatebox{90}{SAGE}}}} & \textsc{base} & \tiny{43.0} & \tiny{45.9} & \tiny{48.6} & \tiny{39.4} & \tiny{38.4} & \tiny{42.2} & \multicolumn{1}{c|}{\tiny{52.6}} & 44.3 & 52.6 & 37.9 & 51.0 & 27.1 & 19.8 \\ \cline{3-16} 
\multicolumn{1}{c|}{} & \multicolumn{1}{c|}{} & +\algp{} & \tiny{55.0} & \tiny{58.0} & \tiny{56.3} & \tiny{61.4} & \tiny{64.1} & \tiny{60.9} & \multicolumn{1}{c|}{\tiny{64.4}} & \cellcolor{Gray} \textbf{60.0 (\tiny{+15.7})} & \cellcolor{Gray} \textbf{64.4 (\tiny{+11.8})} & \cellcolor{Gray} \textbf{59.4 (\tiny{+21.6})} & \cellcolor{Gray} \textbf{63.9 (\tiny{+12.8})} & \cellcolor{Gray} \textbf{17.5 (\tiny{-9.7})} & \cellcolor{Gray} \textbf{13.2 (\tiny{-6.6})} \\
\multicolumn{1}{c|}{} & \multicolumn{1}{c|}{} & +\algt{} & \tiny{53.2} & \tiny{55.9} & \tiny{56.5} & \tiny{61.9} & \tiny{66.3} & \tiny{62.3} & \multicolumn{1}{c|}{\tiny{63.8}} & \cellcolor{Gray} \textbf{60.0 (\tiny{+15.7})} & \cellcolor{Gray} \textbf{66.3 (\tiny{+13.8})} & \cellcolor{Gray} \textbf{59.3 (\tiny{+21.4})} & \cellcolor{Gray} \textbf{65.9 (\tiny{+14.9})} & \cellcolor{Gray} \textbf{18.6 (\tiny{-8.6})} & \cellcolor{Gray} \textbf{12.8 (\tiny{-7.0})} \\ \hline
\multicolumn{1}{c|}{\multirow{9}{*}{\textbf{\rotatebox{90}{PubMed}}}} & \multicolumn{1}{c|}{\multirow{3}{*}{\textbf{\rotatebox{90}{GCN}}}} & \textsc{base} & \tiny{64.2} & \tiny{71.2} & \tiny{71.5} & \tiny{65.0} & \tiny{64.4} & \tiny{74.0} & \multicolumn{1}{c|}{\tiny{73.7}} & 69.1 & 74.0 & 63.5 & 71.3 & 23.2 & 11.9 \\ \cline{3-16} 
\multicolumn{1}{c|}{} & \multicolumn{1}{c|}{} & +\algp{} & \tiny{68.6} & \tiny{74.2} & \tiny{73.2} & \tiny{72.5} & \tiny{73.2} & \tiny{73.1} & \multicolumn{1}{c|}{\tiny{76.1}} & \cellcolor{Gray} \textbf{73.0 (\tiny{+3.8})} & \cellcolor{Gray} \textbf{76.1 (\tiny{+2.1})} & \cellcolor{Gray} \textbf{72.0 (\tiny{+8.5})} & \cellcolor{Gray} \textbf{75.8 (\tiny{+4.5})} & \cellcolor{Gray} \textbf{9.1 (\tiny{-14.1})} & \cellcolor{Gray} \textbf{3.4 (\tiny{-8.6})} \\
\multicolumn{1}{c|}{} & \multicolumn{1}{c|}{} & +\algt{} & \tiny{67.6} & \tiny{73.4} & \tiny{72.5} & \tiny{72.9} & \tiny{73.1} & \tiny{76.6} & \multicolumn{1}{c|}{\tiny{76.9}} & \cellcolor{Gray} \textbf{73.3 (\tiny{+4.1})} & \cellcolor{Gray} \textbf{76.9 (\tiny{+2.9})} & \cellcolor{Gray} \textbf{72.6 (\tiny{+9.1})} & \cellcolor{Gray} \textbf{76.9 (\tiny{+5.6})} & \cellcolor{Gray} \textbf{9.9 (\tiny{-13.4})} & \cellcolor{Gray} \textbf{5.1 (\tiny{-6.8})} \\ \cline{2-16} 
\multicolumn{1}{c|}{} & \multicolumn{1}{c|}{\multirow{3}{*}{\textbf{\rotatebox{90}{GAT}}}} & \textsc{base} & \tiny{65.5} & \tiny{68.4} & \tiny{71.2} & \tiny{65.1} & \tiny{64.8} & \tiny{68.7} & \multicolumn{1}{c|}{\tiny{73.1}} & 68.1 & 73.1 & 62.4 & 71.8 & 24.8 & 10.3 \\ \cline{3-16} 
\multicolumn{1}{c|}{} & \multicolumn{1}{c|}{} & +\algp{} & \tiny{73.2} & \tiny{75.3} & \tiny{75.6} & \tiny{73.3} & \tiny{73.9} & \tiny{74.7} & \multicolumn{1}{c|}{\tiny{74.3}} & \cellcolor{Gray} \textbf{74.3 (\tiny{+6.2})} & \cellcolor{Gray} \textbf{75.6 (\tiny{+2.4})} & \cellcolor{Gray} \textbf{73.4 (\tiny{+11.1})} & \cellcolor{Gray} \textbf{75.1 (\tiny{+3.3})} & \cellcolor{Gray} \textbf{6.5 (\tiny{-18.2})} & \cellcolor{Gray} \textbf{3.0 (\tiny{-7.3})} \\
\multicolumn{1}{c|}{} & \multicolumn{1}{c|}{} & +\algt{} & \tiny{74.8} & \tiny{74.5} & \tiny{75.2} & \tiny{73.9} & \tiny{74.1} & \tiny{74.4} & \multicolumn{1}{c|}{\tiny{75.7}} & \cellcolor{Gray} \textbf{74.6 (\tiny{+6.5})} & \cellcolor{Gray} \textbf{75.7 (\tiny{+2.5})} & \cellcolor{Gray} \textbf{73.9 (\tiny{+11.5})} & \cellcolor{Gray} \textbf{75.0 (\tiny{+3.2})} & \cellcolor{Gray} \textbf{6.9 (\tiny{-17.8})} & \cellcolor{Gray} \textbf{4.6 (\tiny{-5.7})} \\ \cline{2-16} 
\multicolumn{1}{c|}{} & \multicolumn{1}{c|}{\multirow{3}{*}{\textbf{\rotatebox{90}{SAGE}}}} & \textsc{base} & \tiny{67.6} & \tiny{68.0} & \tiny{69.1} & \tiny{69.2} & \tiny{65.0} & \tiny{71.5} & \multicolumn{1}{c|}{\tiny{71.4}} & 68.8 & 71.5 & 64.5 & 70.1 & 22.1 & 11.8 \\ \cline{3-16} 
\multicolumn{1}{c|}{} & \multicolumn{1}{c|}{} & +\algp{} & \tiny{75.3} & \tiny{74.6} & \tiny{74.2} & \tiny{74.9} & \tiny{74.6} & \tiny{74.7} & \multicolumn{1}{c|}{\tiny{75.9}} & \cellcolor{Gray} \textbf{74.9 (\tiny{+6.1})} & \cellcolor{Gray} \textbf{75.9 (\tiny{+4.3})} & \cellcolor{Gray} \textbf{74.2 (\tiny{+9.7})} & \cellcolor{Gray} \textbf{75.3 (\tiny{+5.3})} & \cellcolor{Gray} \textbf{7.6 (\tiny{-14.4})} & \cellcolor{Gray} \textbf{3.4 (\tiny{-8.4})} \\
\multicolumn{1}{c|}{} & \multicolumn{1}{c|}{} & +\algt{} & \tiny{77.4} & \tiny{75.4} & \tiny{75.3} & \tiny{75.8} & \tiny{77.3} & \tiny{76.1} & \multicolumn{1}{c|}{\tiny{76.5}} & \cellcolor{Gray} \textbf{76.3 (\tiny{+7.4})} & \cellcolor{Gray} \textbf{77.4 (\tiny{+5.8})} & \cellcolor{Gray} \textbf{75.8 (\tiny{+11.3})} & \cellcolor{Gray} \textbf{76.9 (\tiny{+6.9})} & \cellcolor{Gray} \textbf{6.8 (\tiny{-15.3})} & \cellcolor{Gray} \textbf{4.1 (\tiny{-7.7})} \\ \bottomrule
\end{tabular}
}
\begin{tablenotes}
    \tiny\item[]\hspace*{-10pt}
    *\algp{}/\algt{}: \alg{} with $0^\textnormal{th}$/$1^\textnormal{st}$-order posterior likelihood estimation. ERM: Empirical Risk Minimization (standard training), RW: Reweight~\citep{japkowicz2002class}, RN: ReNode~\citep{chen2021renode}, RS: Resampling~\citep{japkowicz2002class}, SM: SMOTE~\citep{chawla2002smote}, GS: GraphSMOTE~\citep{zhao2021graphsmote}, GE: GraphENS~\citep{park2021graphens}.\\
    \hspace*{-8pt}
\end{tablenotes}
\end{threeparttable}
\vspace{-10pt}
\end{table*}
\endgroup

%% file: sections/4-experiments.tex
\section{Experiments}
\label{sec:experiments}

We carry out systematic experiments and analysis to validate \alg{} in the following aspects:
\textbf{(i) Effectiveness} in both \textit{promoting imbalanced node classification} and \textit{mitigating the prediction bias} between different classes.
\textbf{(ii) Versatility} in cooperating with and further boosting various CIGL techniques and GNN backbones.
\textbf{(iii) Robustness} to extreme class imbalance.
\textbf{(iv) Efficiency} in real-world applications.

\textbf{Experiment Protocol.}
We validate \alg{} on five benchmark datasets for semi-supervised node classification, 
including the \textit{Cora, CiteSeer, PubMed} from Plantoid graphs~\citep{sen2008planetoid}, and larger-scale \textit{CS, Physics} from co-author networks~\citep{shchur2018pitfalls} with high-dimensional features.
Following the same setting as prior studies~\citep{park2021graphens,song2022tam,zhao2021graphsmote}, we select half of the classes as minority.
The imbalance ratio $\rho = n_{max}/n_{min} \geq 1$ is the ratio between the size of the largest class to the smallest class, i.e., more imbalance $\Leftrightarrow$ higher IR.
Detailed data statistics and class distributions can be found in Appendix~\ref{sec:apd-setup-data}.
We test \alg{} with six CIGL techniques~\citep{park2021graphens,chen2021renode,zhao2021graphsmote,chawla2002smote,japkowicz2002class} and three GNN backbones~\citep{gat,sage,gcn} under all possible combinations to fully validate \alg{}'s effectiveness and versatility in practice.
Note that although there are other techniques available for CIGL~\citep{hong2021disentangling,kang2019decoupling,shi2020drgcn}, previous studies~\citep{park2021graphens,song2022tam} have shown they are generally outperformed by the baselines we use.
Detailed settings can be found in Appendix~\ref{sec:apd-setup-baseline}.
To ensure a comprehensive evaluation, we employ \textbf{three} metrics to assess both the classification performance (Balanced Accuracy, Macro-F1) and the model predictive bias (PerfStd, i.e., the standard deviation of accuracy scores across all classes).
Lower PerfStd indicates smaller performance gap between all majority and minority classes, and thus smaller predictive bias.
For clarity, we use $\uparrow$/$\downarrow$ to denote larger/smaller is better for each metric.

\noindent\textbf{\alg{} significantly boosts various CIGL techniques.}
We report the main results in Table~\ref{tab:main} (IR=10).
In all settings (3 datasets$\times$3 backbones$\times$7 baselines$\times$3 metrics), \alg{} achieves significant and consistent performance improvements over other CIGL techniques, which also yields new state-of-the-art performance.
Specifically:
\textbf{(1)} By mitigating AMP and DMP, \alg{} further boosts the best CIGL baseline by a large margin, e.g., it boosts the best balanced accuracy score by 4.1/13.8/5.8 on Cora/CiteSeer/PubMed datasets.
\textbf{(2)} In addition to better classification performance, \alg{} also greatly reduces the predictive bias in CIGL, with up to 10.7/14.5/18.2 average performance deviation reduction on Cora/CiteSeer/PubMed.
\textbf{(3)} Compared with \algp{}, \algt{} achieves better classification performance with first-order posterior likelihood estimation. But we also note that \algp{} performs better in terms of reducing predictive bias and is more computationally efficient, as we will discuss later in the scalability experiments (Table~\ref{tab:runtime}).

\input{sections/tables/tab_varyimb}

\noindent\textbf{\alg{} is robust even under extreme class imbalance.}
We further extend Table~\ref{tab:main} and test \alg{}'s robustness to varying types and levels of imbalance, as reported in Table~\ref{tab:main-varyimb}.
In this experiment, we extend the step imbalance ratio from 10 (used in Table~\ref{tab:main}) to 20 to test \alg{} under even more challenging class imbalance scenarios.
In addition, we consider the natural (long-tail) class imbalance~\citep{park2021graphens} that is commonly observed in real-world graphs with IR of 50 and 100.
Datasets from~\citep{shchur2018pitfalls} (\textit{CS, Physics}) are also included to test \alg{} on large-scale tasks. 
Results show that:
\textbf{(1)} \alg{} is robust to extreme class imbalance, and it consistently boosts the CIGL performance by a significant margin under varying types and levels of imbalance.
\textbf{(2)} The performance drop from increasing IR is significantly lowered by \alg{}, i.e., applying \alg{} improves model's robustness to extreme class imbalance.
\textbf{(3)} \textit{\alg{}'s advantage is even more prominent under higher class imbalance}, e.g., on Cora with step IR, the performance gain of applying \alg{} on Base raised from 8.2 to 18.5 when IR increased from 10 to 20, and similar patterns can be observed in other settings.

\begin{figure}[H]
  \centering
  \subfigure[\alg{} mitigates AMP in multiple CIGL tasks.]{
    \includegraphics[width=0.9\linewidth]{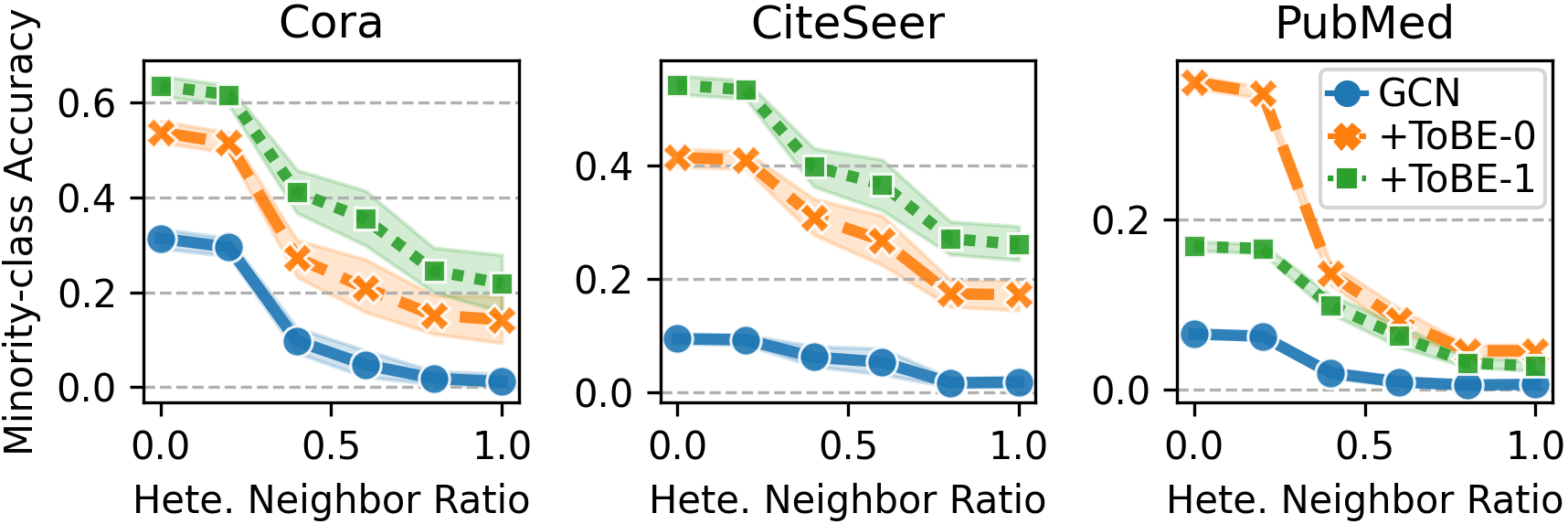}
  }
  \subfigure[\alg{} mitigates DMP in multiple CIGL tasks.]{
    \includegraphics[width=0.9\linewidth]{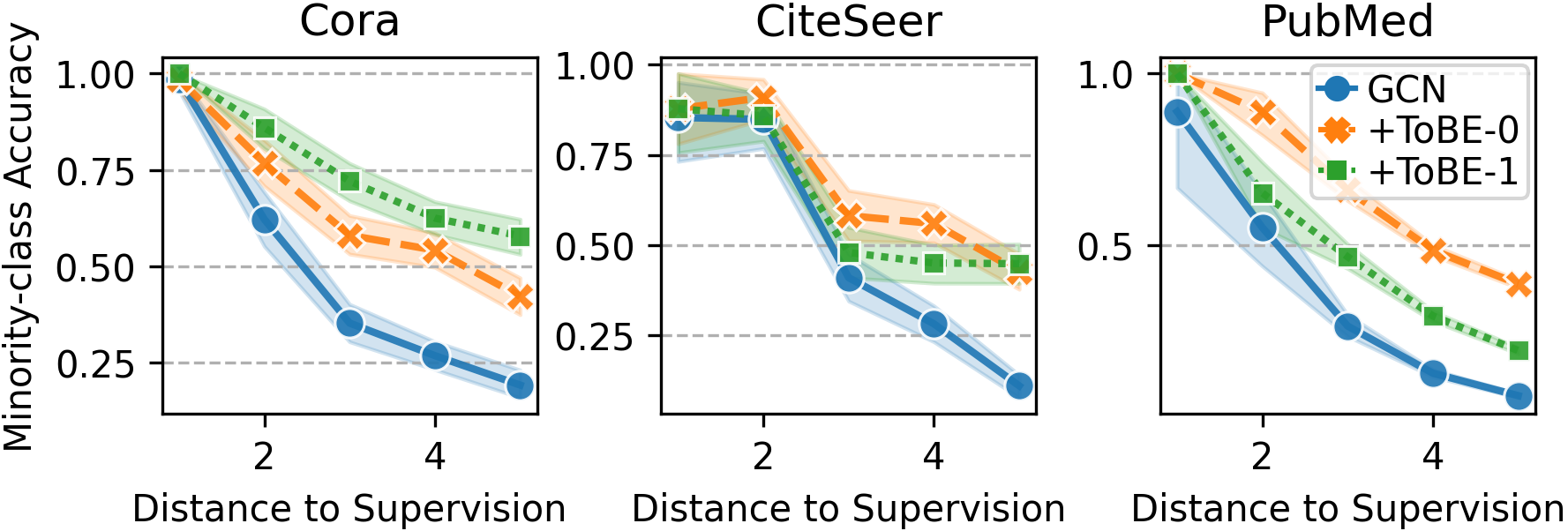}
  }
  \vspace*{-8pt}
  \caption{
  \alg{} effectively alleviates both AMP and DMP.
  }
  \label{fig:mitigate}
\vspace*{-15pt}
\end{figure}

\noindent\textbf{\alg{} effectively alleviates both AMP and DMP.}
We further design experiments to verify to what extent \alg{} can effectively handle the topological challenges identified in this paper, i.e., ambivalent and distant message-passing.
Specifically, we investigate whether \alg{} can improve the prediction accuracy of minority class nodes that are highly influenced by AMP/DMP, i.e., with high heterophilic neighbor ratio/long distance to supervision.
Results are shown in Fig.~\ref{fig:mitigate} (5 independent runs with GCN classifier, IR=10).
As can be observed, \alg{} effectively alleviates the negative impact of AMP and DMP and helps node classifiers to achieve better performance in minority classes.

\noindent\textbf{\alg{} is computationally efficient.}
As previously discussed in the complexity analysis, \alg{} with 0th/1st-order estimation \textbf{scales linearly} with the number of nodes/edges (i.e., with $\gO(|\gV|C)$ or $\gO(|\gE|C)$ complexity).
Since all the operations can be executed in parallel in matrix form, \algp{}/\algt{} has $\gO(\frac{|\gV|C}{D})$/$\gO(\frac{|\gE|C}{D})$ time complexity, where $D$ is the number of available computational units and is usually large for modern GPUs.
Table~\ref{tab:runtime} reports the ratio of virtual nodes/edges to the original graph introduced and the running time of \alg{}.
It can be observed that \alg{} only introduces a small number of virtual nodes/edges, and is highly efficient (taking milliseconds for augmentation) in practice.

\begin{table}[H]
\small
\centering
\caption{Efficiency results of \algp{}/\algt{}.}
\label{tab:runtime}
\begin{threeparttable}
\resizebox{\linewidth}{!}
{
\begin{tabular}{l|ccc}
\toprule
\multicolumn{1}{c|}{\textbf{Dataset}} & \textbf{$\Delta$ Nodes (\%)} & \textbf{$\Delta$ Edges (\%)} & \textbf{$\Delta$ Time (ms)} \\ \hline
Cora                                  & 0.258\%                    & 2.842\%/1.509\%            & 4.50/4.65ms                  \\
CiteSeer                              & 0.180\%                    & 3.715\%/1.081\%            & 4.72/4.97ms                  \\
PubMed                                & 0.015\%                    & 3.175\%/1.464\%            & 6.23/6.64ms                  \\
CS                           & 0.082\%                    & 1.395\%/1.053\%            & 16.97/18.61ms                  \\
Physics                      & 0.014\%                    & 0.797\%/0.527\%            & 30.68/31.91ms                  \\ \bottomrule
\end{tabular}
}
\begin{tablenotes}
    \tiny
    \item[] * Results obtained on an NVIDIA$^\text{\textregistered}$ Tesla V100 32GB GPU.
\end{tablenotes}
\end{threeparttable}
\end{table}
\textbf{Further discussions.} We refer the readers to Appendix for reproducibility details (\S\ref{sec:apd-setup}), ablation study and extended discussions (\S\ref{sec:apd-discussion}), and additional empirical results (\S\ref{sec:apd-results}).

%% file: sections/tables/tab_varyimb.tex
\begin{table*}[t]
\vspace*{-5pt}
\caption{
    \alg{} deliver consistent and significant performance gain to CIGL methods under varying types and levels of class imbalance.
    The numbers in brackets are the performance gain brought about by \alg{} over Base/BestCIGL method.
    We report the Balanced Accuracy here, full results with other metrics can be found in Appendix~\ref{sec:apd-res-full}.
}
\label{tab:main-varyimb}
\begin{threeparttable}
\centering
\resizebox{\linewidth}{!}
{
\begin{tabular}{l|ll|ll|ll|ll|ll}
\toprule
\multicolumn{1}{c|}{\textbf{Dataset}} & \multicolumn{2}{c|}{\textbf{Cora}} & \multicolumn{2}{c|}{\textbf{CiteSeer}} & \multicolumn{2}{c|}{\textbf{PubMed}} & \multicolumn{2}{c|}{\textbf{CS}} & \multicolumn{2}{c}{\textbf{Physics}} \\ \hline
\multicolumn{1}{c|}{\textbf{Step IR}} & \multicolumn{1}{c}{10} & \multicolumn{1}{c|}{20} & \multicolumn{1}{c}{10} & \multicolumn{1}{c|}{20} & \multicolumn{1}{c}{10} & \multicolumn{1}{c|}{20} & \multicolumn{1}{c}{10} & \multicolumn{1}{c|}{20} & \multicolumn{1}{c}{10} & \multicolumn{1}{c}{20} \\ \hline
\hspace{8pt}Base & 61.6 & 52.7 & 37.6 & 34.2 & 64.2 & 60.8 & 75.4 & 65.3 & 80.1 & 67.7 \\
\rowcolor{Gray} + \alg{}  & \textbf{69.8 \footnotesize{(+8.2)}} & \textbf{71.3 \footnotesize{(+18.5)}} & \textbf{55.4 \footnotesize{(+17.7)}} & \textbf{51.3 \footnotesize{(+17.1)}} & \textbf{68.6 \footnotesize{(+4.4)}} & \textbf{63.3 \footnotesize{(+2.5)}} & \textbf{82.6 \footnotesize{(+7.2)}} & \textbf{79.9 \footnotesize{(+14.5)}} & \textbf{87.6 \footnotesize{(+7.5)}} & \textbf{88.0 \footnotesize{(+20.2)}} \\
\hspace{8pt}BestCIGL  & 70.1 & 66.5 & 56.0 & 47.2 & 74.0 & 71.1 & 84.1 & 81.3 & 89.4 & 85.7 \\
\rowcolor{Gray} + \alg{}  & \textbf{74.2 \footnotesize{(+4.1)}} & \textbf{71.6 \footnotesize{(+5.1)}} & \textbf{62.7 \footnotesize{(+6.7)}} & \textbf{62.5 \footnotesize{(+15.3)}} & \textbf{76.9 \footnotesize{(+2.9)}} & \textbf{75.7 \footnotesize{(+4.6)}} & \textbf{86.3 \footnotesize{(+2.2)}} & \textbf{85.6 \footnotesize{(+4.3)}} & \textbf{91.2 \footnotesize{(+1.9)}} & \textbf{90.9 \footnotesize{(+5.2)}} \\
\midrule \midrule
\multicolumn{1}{c|}{\textbf{Natural IR}} & \multicolumn{1}{c}{50} & \multicolumn{1}{c|}{100} & \multicolumn{1}{c}{50} & \multicolumn{1}{c|}{100} & \multicolumn{1}{c}{50} & \multicolumn{1}{c|}{100} & \multicolumn{1}{c}{50} & \multicolumn{1}{c|}{100} & \multicolumn{1}{c}{50} & \multicolumn{1}{c}{100} \\ \hline
\hspace{8pt}Base  & 60.1 & 47.0 & 28.1 & 21.9 & 55.1 & 46.4 & 72.7 & 59.2 & 80.7 & 64.7 \\
\rowcolor{Gray} + \alg{}  & \textbf{68.7 \footnotesize{(+8.6)}} & \textbf{69.6 \footnotesize{(+22.6)}} & \textbf{54.9 \footnotesize{(+26.9)}} & \textbf{48.9 \footnotesize{(+27.0)}} & \textbf{67.2 \footnotesize{(+12.1)}} & \textbf{60.7 \footnotesize{(+14.3)}} & \textbf{78.6 \footnotesize{(+5.9)}} & \textbf{74.7 \footnotesize{(+15.5)}} & \textbf{88.8 \footnotesize{(+8.1)}} & \textbf{87.8 \footnotesize{(+23.2)}} \\
\hspace{8pt}BestCIGL  & 70.0 & 66.2 & 54.5 & 45.0 & 71.3 & 68.9 & 83.9 & 80.9 & 89.5 & 86.2 \\
\rowcolor{Gray} + \alg{}  & \textbf{72.8 \footnotesize{(+2.9)}} & \textbf{70.2 \footnotesize{(+4.0)}} & \textbf{62.5 \footnotesize{(+8.0)}} & \textbf{62.1 \footnotesize{(+17.1)}} & \textbf{76.9 \footnotesize{(+5.6)}} & \textbf{74.9 \footnotesize{(+6.0)}} & \textbf{85.4 \footnotesize{(+1.6)}} & \textbf{84.6 \footnotesize{(+3.7)}} & \textbf{90.7 \footnotesize{(+1.2)}} & \textbf{90.0 \footnotesize{(+3.8)}} \\
\bottomrule
\end{tabular}
}
\begin{tablenotes}
    \tiny
    \item[]
    \hspace*{8pt}
    *\textbf{Base}: vanilla GCN model; \textbf{Base+\alg{}}: applying \alg{} without any other CIGL method; \textbf{BestCIGL}: best CIGL baseline w/o \alg{}; \textbf{BestCIGL+\alg{}}: best CIGL baseline w/ \alg{};
\end{tablenotes}
\end{threeparttable}
\end{table*}

%% file: sections/5-relatedwork.tex
\section{Related Works}
\label{sec:related-works}

\noindent\textbf{Imbalanced graph learning.}
Class imbalance is ubiquitous in many machine-learning tasks and has been extensively studied~\citep{he2008imblearn,krawczyk2016imblearn}.
However, most of the existing works focus on i.i.d. scenarios, which may not be tailored to the unique characteristics of graph data.
To handle imbalanced graph learning, several techniques have been proposed in recent studies (e.g., by adversarial training~\citep{shi2020drgcn,qu2021imgagn}, designing new GNN architectures~\citep{wang2020network,liu2021pick} or loss functions~\citep{song2022tam}), we review the most closely related model-agnostic CR methods here.
One of the early works GraphSMOTE~\citep{zhao2021graphsmote} adopts SMOTE~\citep{chawla2002smote} oversampling in the node embedding space to synthesize minority nodes and complements the topology with a learnable edge predictor.
A more recent work GraphENS~\citep{park2021graphens} synthesizes the ego network through saliency-based ego network mixing to handle the neighbor-overfitting problem.
Most studies are rooted in a \textit{class-rebalancing} perspective and address the imbalance by node/class-wise reweighting or resampling.

\vspace{-3pt}
\noindent\textbf{Topology-imbalance in graphs.}
Topology imbalance is firstly discussed in \citet{chen2021renode}.
They found that ``the unequal structure role of labeled nodes'' can cause influence conflict, and propose to re-weight the labeled nodes based on a conflict detection measure.
Other works further discussed how to better address the issue via position-aware structure learning~\citep{sun2022pastel}, and handle topology-imbalance in fake news detection~\citep{gao2022fakenews} and bankruptcy prediction~\citep{liu2023qtiah}.
These studies discussed concepts related to ``influence conflict/insufficiency''~\citep{chen2021renode}, which motivated us to investigate AMP/DMP in this work.
\rev{
It is worth noting that AMP/DMP coefficients are defined on all nodes based on the $k$-hop local neighborhood, while 
the influence conflict measure in~\citet{chen2021renode} is defined only on labeled nodes by global personalized PageRank score, and influence insufficient has no formal definition.
Therefore, we did not inherit their naming and employ AMP/DMP to distinguish and avoid confusion with the existing concepts.
We theoretically investigate the role of AMP/DMP in shaping class-imbalance bias, and propose an augmentation technique \alg{} tailored for handling AMP and DMP in CIGL. 
}
Results show that the topology imbalance algorithm can also significantly boosted by \alg{}.

\vspace{-3pt}
\noindent\textbf{Heterophilic graph/long-distance propagation.}
Numerous studies exist in the literature concerning learning from heterophilic graphs~\citep{yan2024trainable,xu2023node} and employing multi-hop propagation~\citep{zhao2021adaptivediffusion,fu2024vcr}. 
In particular, heterophilic GNNs often combine intermediate representations to derive more refined structure-aware features~\citep{zhu2020h2gcn}. 
The GPRGNN~\citep{chien2020gprgnn} takes a step further by introducing learnable weights to adaptively combine representations from each layer. 
Meanwhile, in multi-hop propagation, APPNP~\citep{gasteiger2018appnp} stands as a representative technique that leverages personalized PageRank for extracting information from a broader neighborhood.
Nevertheless, these works focus on addressing \textit{global} graph heterophily and long-distance propagation by modifying the GNN architecture or aggregation operators. 
They are not tailored to address class imbalance and cannot readily handle the AMP and DMP.
We refer the readers to Appendix~\ref{sec:apd-results} where we show that \alg{} can also significantly boost the performance of such GNNs~\citep{chien2020gprgnn,gasteiger2018appnp} in various CIGL tasks.

%% file: sections/6-conclusion.tex
\section{Conclusion}
\label{sec:conclusion}

In this paper, we study class-imbalanced graph learning from a novel topological perspective.
We theoretically reveal that two fundamental topological phenomena, i.e., ambivalent and distant message-passing, can greatly exacerbate the predictive bias stemming from class imbalance.
Our findings reveal an unexplored avenue that limits the performance of existing class-rebalancing-based CIGL techniques.
In light of this, we propose \alg{} to handle the topological challenges in CIGL by dynamic topological augmentation.
\alg{} is a swift and model-agnostic framework that can seamlessly complement other CIGL techniques, augmenting their performance and mitigating predictive bias.
Systematic experiments validate \alg{}'s superior effectiveness, versatility, robustness, and efficiency across various CIGL tasks. 

%% file: sections/appendix.tex
\newpage
\begin{center}\Large\textbf{Appendix}\end{center}

% We provide the following content in the Appendix:
\begin{itemize}
\setlength\itemsep{-3pt}
    \item \textbf{Section \ref{sec:apd-proof}: Proofs of Theoretical Results}
    \begin{itemize}
    \setlength\itemsep{-2pt}
        \item \ref{sec:apd-proof-0} - Limiting the distribution of $H_{ij}^k$.
        \item \ref{sec:apd-proof-amp} - Proof of Theorem~\ref{the:amp} (AMP).
        \item \ref{sec:apd-proof-dmp} - Proof of Theorem~\ref{the:dmp} (DMP).
    \end{itemize}
    \item \textbf{Section \ref{sec:apd-setup}: Reproducibility Details}
    \begin{itemize}
    \setlength\itemsep{-2pt}
        \item \ref{sec:apd-setup-data} - Statistics of the used datasets.
        \item \ref{sec:apd-setup-baseline} - Implementation details of baselines.
        \item \ref{sec:apd-setup-eval} - Evaluation protocols.
    \end{itemize}
    \item \textbf{Section \ref{sec:apd-discussion}: Further Discussions}
    \begin{itemize}
    \setlength\itemsep{-2pt}
        \item \ref{sec:apd-dis-ablation} - Ablation study of \alg{}.
        \item \ref{sec:apd-dis-speedup} - Further speedup of \alg{} in practice.
        \item \ref{sec:apd-dis-choose} - Remarks on choosing between \algp{}/\algt{}.
        \item \ref{sec:apd-dis-limitation} - Limitation and future works.
    \end{itemize}
    \item \textbf{Section \ref{sec:apd-results}: Additional Experiments, Results, and Analysis}
    \begin{itemize}
    \setlength\itemsep{-2pt}
        \item \ref{sec:apd-res-large} - Experiments on additional large-scale graphs
        \item \ref{sec:apd-res-additional} - Comparison with additional independent CIGL baselines.
        \item \ref{sec:apd-res-full} - Full results with all metrics, error bar, and additional GNN backbones.
    \end{itemize}
\end{itemize}

\section{Proofs of Theoretical Results}
\label{sec:apd-proof}

Define random variables $H^k_{ij}:=|\gV_j\cap\gH(u,k)|$ denote the number of class-$j$ $k$-hop homo-connected neighbors of a node $u\in\gV_i$. 
Note that the results of both Theorems~\ref{the:amp} \& \ref{the:dmp} depend only on the distributions of $H^k_{ij}$. Thus, we will first derive the limiting distributions of $H^k_{ij}$ as a technical lemma, and then give the proofs of Theorems~\ref{the:amp} \& \ref{the:dmp}.

\subsection{Limiting Distributions of $H^k_{ij}$}
\label{sec:apd-proof-0}
To count the number of homo-connected neighbors, consider the breadth-first search (BFS) tree rooted at node $u\in\gV_1$. By enumerating the numbers of $1,\ldots,k$-hop homo-connected neighbors in the BFS tree respectively, we can calculate the exact joint distribution of $(H^k_{11},H^k_{12})$:
\begin{align*}
\mathbb P\{H^k_{11}=s,H^k_{12}=s'\}=\!\!\!\!\!\!\!\!\sum_{\begin{subarray}{c}a_1+\cdots+a_k=s\\b_1+\cdots+b_k=s'\end{subarray}}\!\!\!\!{}&\binom{n_1-1}{a_1,\dots,a_k,n_1-1-s}\binom{n_2}{b_1,\dots,b_k,n_2-s'}\\
&\!\!\!\!\!\!\!\!p^{a_1}\bigg(\prod_{t=2}^k(1-p)^{a_t(1+a_1+\cdots+a_{t-2})}(1-(1-p)^{a_{t-1}})^{a_t}\bigg)(1-p)^{(n_1-1-s)(1+s-a_k)}\\
&\!\!\!\!\!\!\!\!q^{b_1}\bigg(\prod_{t=2}^k(1-q)^{b_t}(1-p)^{b_t(b_1+\cdots+b_{t-2})}(1-(1-p)^{b_{t-1}})^{b_t}\bigg)(1-q)^{n_2-s'}(1-p)^{(n_2-s')(s'-b_k)}.
\end{align*}
Thus, $H^k_{11}$ and $H^k_{12}$ are independent, and their marginal distributions are:
\begin{align*}
\mathbb P\{H^k_{11}=s\}=\sum_{a_1+\cdots+a_k=s}&\binom{n_1-1}{a_1,\dots,a_k,n_1-1-s}\\&p^{a_1}\bigg(\prod_{t=2}^k(1-p)^{a_t(1+a_1+\cdots+a_{t-2})}(1-(1-p)^{a_{t-1}})^{a_t}\bigg)(1-p)^{(n_1-1-s)(1+s-a_k)},\\
\mathbb P\{H^k_{12}=s\}=\sum_{b_1+\cdots+b_k=s}&\binom{n_2}{b_1,\dots,b_k,n_2-s}\\&q^{b_1}\bigg(\prod_{t=2}^k(1-q)^{b_t}(1-p)^{b_t(b_1+\cdots+b_{t-2})}(1-(1-p)^{b_{t-1}})^{b_t}\bigg)(1-q)^{n_2-s}(1-p)^{(n_2-s)(s-b_k)}.
\end{align*}
Now consider $n\to\infty$. Let
\AM{
\beta_{11}&:=\lim_{n\to\infty}n_1\cdot p=\beta,\\
\beta_{22}&:=\lim_{n\to\infty}n_2\cdot p=\rho\beta,\\
\beta_{21}&:=\lim_{n\to\infty}n_1\cdot q=\beta\frac qp,\\
\beta_{12}&:=\lim_{n\to\infty}n_2\cdot q=\rho\beta\frac qp.
}
Then, the limiting distributions of $H^k_{11}$ and $H^k_{12}$ are:
\begin{align*}
\mathbb P\{H^k_{11}=s\}\sim{}&\sum_{a_1+\cdots+a_k=s}\frac{n_1^s}{a_1!\cdots a_k!}p^{a_1}\bigg(\prod_{t=2}^k(a_{t-1}p)^{a_t}\bigg)(1-p)^{n_1(1+s-a_k)}\\
\to{}&\mathrm e^{-\beta_{11}}\sum_{a_1+\cdots+a_k=s}\frac{(\beta_{11}\mathrm e^{-\beta_{11}})^{a_1}}{a_1!}\bigg(\prod_{t=2}^{k-1}\frac{(a_{t-1}\beta_{11}\mathrm e^{-\beta_{11}})^{a_t}}{a_t!}\bigg)\frac{(a_{k-1}\beta_{11})^{a_k}}{a_k!},\\
\mathbb P\{H^k_{12}=s\}\sim{}&\sum_{b_1+\cdots+b_k=s}\frac{n_2^s}{b_1!\cdots b_k!}q^{b_1}\bigg(\prod_{t=2}^k(b_{t-1}p)^{b_t}\bigg)(1-q)^{n_2}(1-p)^{n_2(s-b_k)}\\
\to{}&\mathrm e^{-\beta_{12}}\sum_{b_1+\cdots+b_k=s}\frac{(\beta_{12}\mathrm e^{-\beta_{22}})^{b_1}}{b_1!}\bigg(\prod_{t=2}^{k-1}\frac{(b_{t-1}\beta_{22}\mathrm e^{-\beta_{22}})^{b_t}}{b_t!}\bigg)\frac{(b_{k-1}\beta_{22})^{b_k}}{b_k!}.
\end{align*}

\begin{figure}[h]
\centering
\includegraphics[width=.8\linewidth]{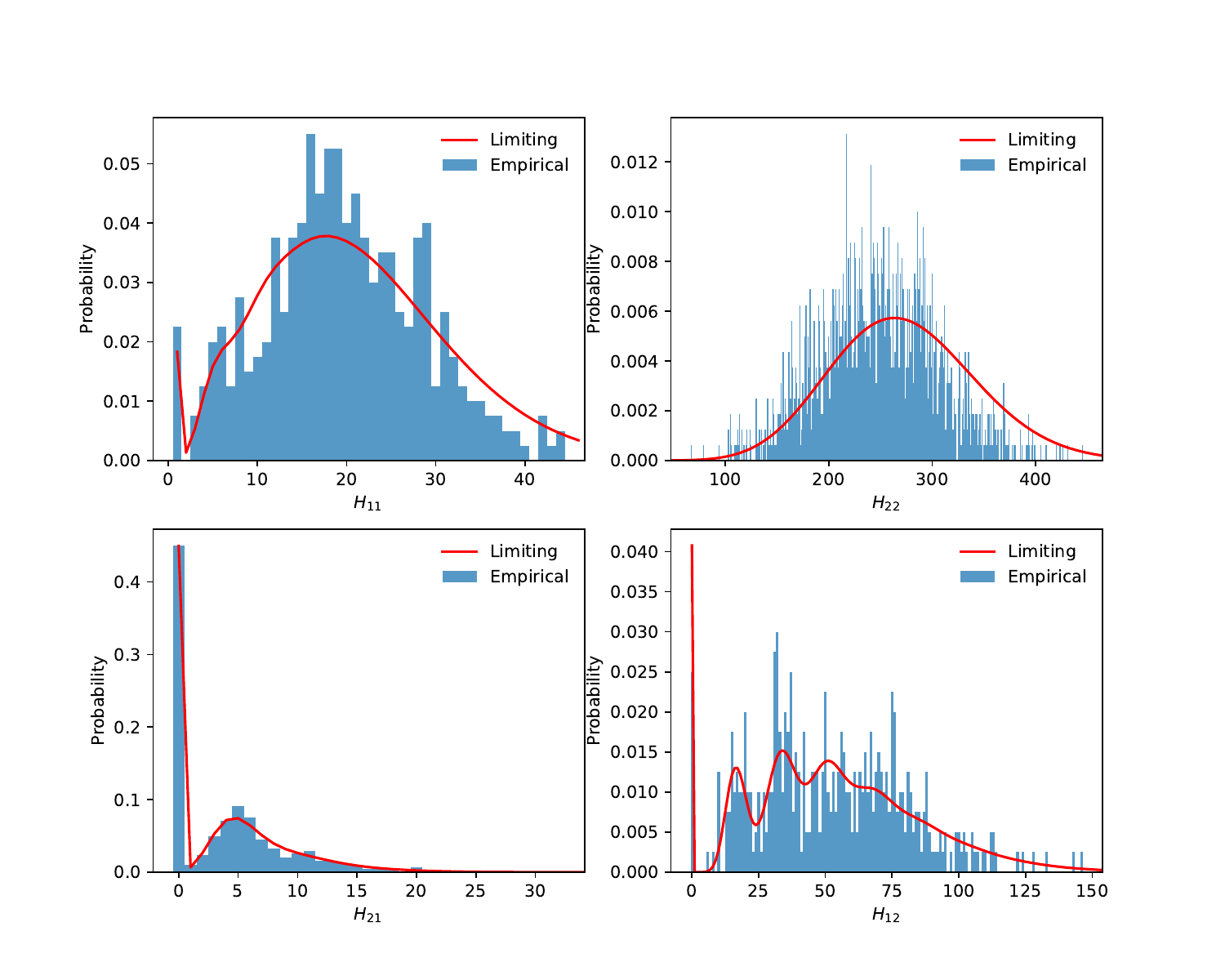}
\caption{Distributions of $H^k_{ij}$. Simulated with $n=2000$, $\rho=4$, $p=0.01$, $q=0.002$, $k=2$.}
\label{fig:proof-distr}
\end{figure}

Figure~\ref{fig:proof-distr} shows that the limiting distributions are good approximation for finite $n$.

\subsection{Proof of Theorem~\ref{the:amp}}
\label{sec:apd-proof-amp}
\begin{proof}[\unskip\nopunct]
Note that for any $t'=1,\dots,k$,
$$
\mathrm e^{-\beta_{11}}\sum_{a_1=0}^\infty\cdots\sum_{a_k=0}^\infty a_{t'}\cdot\frac{(\beta_{11}\mathrm e^{-\beta_{11}})^{a_1}}{a_1!}\bigg(\prod_{t=2}^{k-1}\frac{(a_{t-1}\beta_{11}\mathrm e^{-\beta_{11}})^{a_t}}{a_t!}\bigg)\frac{(a_{k-1}\beta_{11})^{a_k}}{a_k!}=\beta_{11}^{t'}.
$$
Thus,
$$
\begin{aligned}
&\lim_{n\to\infty}\mathbb E[H_{11}^k]=\sum_{s=0}^\infty s\cdot\lim_{n\to\infty}\Prb\{H^k_{11}=s\}\\
={}&\sum_{s=0}^\infty s\cdot\mathrm e^{-\beta_{11}}\sum_{a_1+\cdots+a_k=s}\frac{(\beta_{11}\mathrm e^{-\beta_{11}})^{a_1}}{a_1!}\bigg(\prod_{t=2}^{k-1}\frac{(a_{t-1}\beta_{11}\mathrm e^{-\beta_{11}})^{a_t}}{a_t!}\bigg)\frac{(a_{k-1}\beta_{11})^{a_k}}{a_k!}\\
={}&\sum_{s=0}^\infty\mathrm e^{-\beta_{11}}\sum_{a_1+\cdots+a_k=s}s\cdot\frac{(\beta_{11}\mathrm e^{-\beta_{11}})^{a_1}}{a_1!}\bigg(\prod_{t=2}^{k-1}\frac{(a_{t-1}\beta_{11}\mathrm e^{-\beta_{11}})^{a_t}}{a_t!}\bigg)\frac{(a_{k-1}\beta_{11})^{a_k}}{a_k!}\\
={}&\mathrm e^{-\beta_{11}}\sum_{a_1=0}^\infty\cdots\sum_{a_k=0}^\infty(a_1+\cdots+a_k)\cdot\frac{(\beta_{11}\mathrm e^{-\beta_{11}})^{a_1}}{a_1!}\bigg(\prod_{t=2}^{k-1}\frac{(a_{t-1}\beta_{11}\mathrm e^{-\beta_{11}})^{a_t}}{a_t!}\bigg)\frac{(a_{k-1}\beta_{11})^{a_k}}{a_k!}\\
={}&\sum_{t'=1}^k\mathrm e^{-\beta_{11}}\sum_{a_1=0}^\infty\cdots\sum_{a_k=0}^\infty a_{t'}\cdot\frac{(\beta_{11}\mathrm e^{-\beta_{11}})^{a_1}}{a_1!}\bigg(\prod_{t=2}^{k-1}\frac{(a_{t-1}\beta_{11}\mathrm e^{-\beta_{11}})^{a_t}}{a_t!}\bigg)\frac{(a_{k-1}\beta_{11})^{a_k}}{a_k!}\\
={}&\sum_{t'=1}^k\beta_{11}^{t'}.
\end{aligned}
$$
Similarly,
\begin{align*}
\lim_{n\to\infty}\mathbb E[H_{22}^k]&=\sum_{t=1}^k\beta_{22}^{t},\\
\lim_{n\to\infty}\mathbb E[H_{12}^k]&=\sum_{t=1}^k\beta_{12}\beta_{22}^{t-1},\\
\lim_{n\to\infty}\mathbb E[H_{21}^k]&=\sum_{t=1}^k\beta_{21}\beta_{11}^{t-1}.
\end{align*}
It follows that
\begin{align*}
\lim_{n\to\infty}\frac{\alpha_1^k}{\alpha_2^k}&=\lim_{n\to\infty}\frac{\mathbb E[H^k_{12}]/\mathbb E[H^k_{11}]}{\mathbb E[H^k_{21}]/\mathbb E[H^k_{22}]}\\
&=\frac{\sum_{t=1}^k\beta_{12}\beta_{22}^{t-1}\big/\sum_{t=1}^k\beta_{11}^t}{\sum_{t=1}^k\beta_{21}\beta_{11}^{t-1}\big/\sum_{t=1}^k\beta_{22}^t}\\
&=\frac{\beta_{12}\beta_{22}}{\beta_{21}\beta_{11}}\cdot\frac{\big(\sum_{t=1}^k\beta_{22}^{t-1}\big)^2}{\big(\sum_{t=1}^k\beta_{11}^{t-1}\big)^2}\\
&=\bigg(\rho\cdot\frac{\sum_{t=1}^k(\rho\beta)^{t-1}}{\sum_{t=1}^k\beta^{t-1}}\bigg)^{\!2}.\qedhere
\end{align*}
\end{proof}

\subsection{Proof of Theorem~\ref{the:dmp}}\label{sec:apd-proof-dmp}

\begin{proof}[\unskip\nopunct]
For $k=2$, note the identity:
\AM{\sum_{s=0}^\infty\sum_{a=0}^s\frac{\lambda^a(\mu a)^{s-a}}{a!(s-a)!}=\sum_{a=0}^\infty\frac{\lambda^a}{a!}\sum_{b=0}^\infty\frac{(\mu a)^b}{b!}=\sum_{a=0}^\infty\frac{\lambda^a\RM e^{\mu a}}{a!}=\RM e^{\lambda\RM e^\mu}.}
It follows that (with $\lambda=(1-r_1^\text L)\beta_{11}\RM e^{-\beta_{11}}$ and $\mu=(1-r_1^\text L)\beta_{11}$)
\AM{\lim_{n\to\infty}\Exp[(1-r_1^\text L)^{H^2_{11}}]={}&\sum_{s=0}^\infty(1-r_1^\text L)^{s}\cdot\lim_{n\to\infty}\Prb\{H^2_{11}=s\}\\
={}&\sum_{s=0}^\infty(1-r_1^\text L)^{s}\cdot\RM e^{-\beta_{11}}\sum_{a=0}^{s}\frac{(\beta_{11}\RM e^{-\beta_{11}})^a(\beta_{11}a)^{s-a}}{a!(s-a)!}\\
={}&\RM e^{-\beta_{11}}\sum_{s=0}^\infty\sum_{a=0}^{s}\frac{((1-r_1^\text L)\beta_{11}\RM e^{-\beta_{11}})^a((1-r_1^\text L)\beta_{11}a)^{s-a}}{a!(s-a)!}\\
={}&\RM e^{-\beta_{11}}\RM e^{(1-r_1^\text L)\beta_{11}\RM e^{-\beta_{11}}\RM e^{(1-r_1^\text L)\beta_{11}}}\\
={}&\RM e^{-(1-(1-r_1^\text L)\RM e^{-r_1^\text L\beta_{11}})\beta_{11}}\\
\approx{}&\RM e^{-\beta_{11}}.
}
For $k=3$, similarly,
\AM{
\lim_{n\to\infty}\Exp[(1-r_1^\text L)^{H^3_{11}}]&=\mathrm e^{-\big(1-(1-r_1^{\text L})\beta_{11}\mathrm e^{-\big(1-(1-r_1^{\text L})\beta_{11}\mathrm e^{-r_1^{\text L}\beta_{11}}\big)\beta_{11}}\big)\beta_{11}}\approx\RM e^{-\beta_{11}}.
}
In general, the result for $k$ has $k$ nested exponentiations, but we still have:
\AM{\lim_{n\to\infty}\Exp[(1-r_1^\text L)^{H^k_{11}}]\approx\RM e^{-\beta_{11}}.}
Similarly,
\AM{\lim_{n\to\infty}\Exp[(1-r_2^\text L)^{H^k_{12}}]&\approx\RM e^{-\beta_{12}},\\
\lim_{n\to\infty}\Exp[(1-r_2^\text L)^{H^k_{22}}]&\approx\RM e^{-\beta_{22}},\\
\lim_{n\to\infty}\Exp[(1-r_1^\text L)^{H^k_{21}}]&\approx\RM e^{-\beta_{21}}.}
By the law of total probability and the independence of $H^k_{i1}$ and $H^k_{i2}$,
\AM{\frac{\delta^k_1}{\delta^k_2}={}&\frac{\Exp[(1-r_1^\text L)^{H^k_{11}+1}(1-(1-r_2^\text L)^{H^k_{12}})]}{\Exp[(1-r_2^\text L)^{H^k_{22}+1}(1-(1-r_1^\text L)^{H^k_{21}})]}\\
={}&\frac{(1-r_1^\text L)\Exp[(1-r_1^\text L)^{H^k_{11}}](1-\Exp[(1-r_2^\text L)^{H^k_{12}}])}{(1-r_2^\text L)\Exp[(1-r_2^\text L)^{H^k_{22}}](1-\Exp[(1-r_1^\text L)^{H^k_{21}}])}.}
It follows that
\AM{
\lim_{n\to\infty}\frac{\delta^k_1}{\delta^k_2}\approx{}&\frac{(1-r_1^\text L)\RM e^{-\beta_{11}}(1-\RM e^{-\beta_{12}})}{(1-r_2^\text L)\RM e^{-\beta_{22}}(1-\RM e^{-\beta_{21}})}\\
\approx{}&\frac{1-r_1^\text L}{1-r_2^\text L}\RM e^{\beta_{22}-\beta_{11}}=\frac{1-r_1^\text L}{1-r_2^\text L}\RM e^{(\rho-1)\beta}
.\qedhere}
\end{proof}

\section{Reproducibility}\label{sec:apd-setup}

In this section, we describe the detailed experimental settings including (\textbf{\S\ref{sec:apd-setup-data}}) data statistics, (\textbf{\S\ref{sec:apd-setup-baseline}}) baseline settings, and (\textbf{\S\ref{sec:apd-setup-eval}}) evaluation protocols.
The source code for implementing and evaluating \alg{} and all the CIGL baseline methods will be released after the paper is published.

\subsection{Data Statistics}\label{sec:apd-setup-data}

As previously described, we adopt 5 benchmark graph datasets: the Cora, CiteSeer, and PubMed citation networks~\citep{sen2008planetoid}, and the CS and Physics coauthor networks~\citep{shchur2018pitfalls} to test \alg{} on large graphs with more nodes and high-dimensional features.
All datasets are publicly available\footnote{\href{https://pytorch-geometric.readthedocs.io/en/latest/modules/datasets.html}{https://pytorch-geometric.readthedocs.io/en/latest/modules/datasets.html}.}.
Table~\ref{tab:dataset} summarizes the dataset statistics.

\begin{table}[h]
\centering
\footnotesize
% % \vspace*{-10pt}
\caption{Statistics of datasets.}
\label{tab:dataset}
\begin{tabular}{l|cccc}
\toprule
\multicolumn{1}{c|}{\textbf{Dataset}} & \textbf{\#nodes} & \textbf{\#edges} & \textbf{\#features} & \textbf{\#classes} \\ \midrule
Cora                                  & 2,708            & 10,556           & 1,433               & 7                  \\
CiteSeer                              & 3,327            & 9,104            & 3,703               & 6                  \\
PubMed                                & 19,717           & 88,648           & 500                 & 3                  \\
CS                                    & 18,333           & 163,788          & 6,805               & 15                 \\
Physics                               & 34,493           & 495,924          & 8,415               & 5                  \\ \bottomrule
\end{tabular}
\end{table}

\begin{figure}[h]
\centering
  \includegraphics[width=\linewidth]{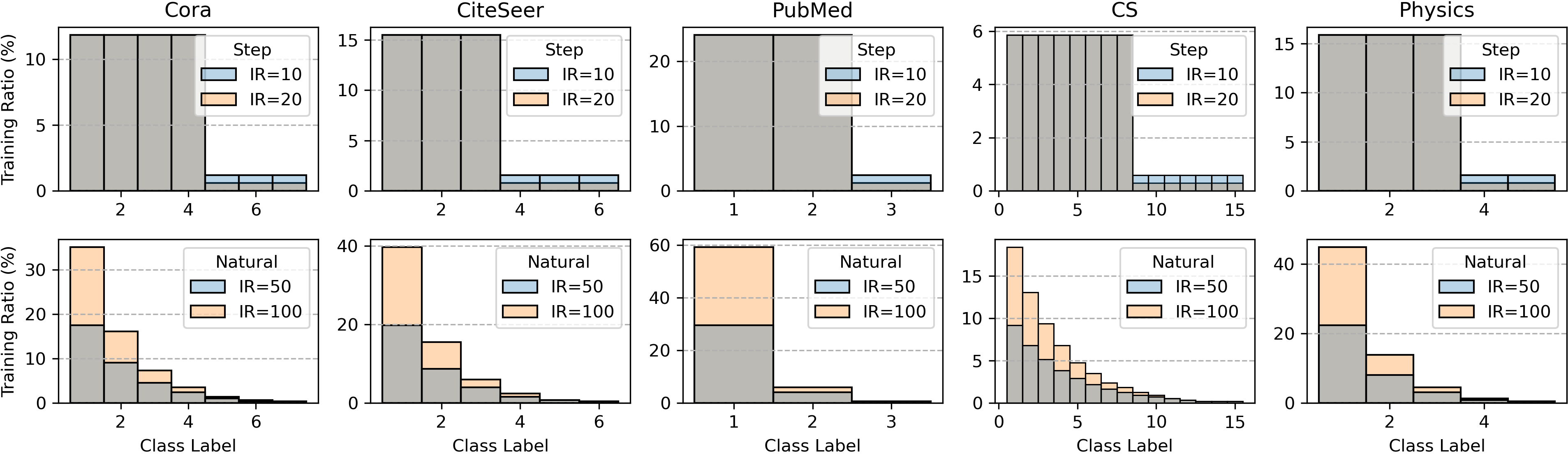}
  \caption{
    Class distribution of training datasets under step and natural imbalance.
  }
  \label{fig:train_dist}
\end{figure}

We follow previous works~\citep{zhao2021graphsmote,park2021graphens,song2022tam} to construct and adjust the class imbalanced node classification tasks.
For step imbalance, we select half of the classes ($\lfloor m/2 \rfloor$) as minority classes and the rest as majority classes.
We follow the public split~\citep{sen2008planetoid} for semi-supervised node classification where each class has 20 training nodes, then randomly remove minority class training nodes until the given imbalance ratio (IR) is met.
The imbalance ratio is defined as $\text{IR}=\frac{\#\text{majority training nodes}}{\#\text{minority training nodes}} \in [1, \infty)$, i.e., more imbalanced data has higher IR.
For natural imbalance, we simulate the long-tail class imbalance present in real-world data by utilizing a power-law distribution.
Specifically, for a given IR, the largest head class have $n_\text{head} = \text{IR}$ training nodes, and the smallest tail class have 1 training node.
The number of training nodes of the $k$-th class is determined by $n_k = \lfloor n_\text{head}^{\lambda_k} \rfloor, \lambda_k = \frac{m-k}{m-1}$.
We set the IR (largest class to smallest class) to 50/100 to test \alg{}'s robustness under natural and extreme class imbalance.
We show the training data distribution under step and natural imbalance in Fig.~\ref{fig:train_dist}.

\subsection{Baseline Settings}\label{sec:apd-setup-baseline}

To fully validate \alg{}'s performance and compatibility with existing CIGL techniques and GNN backbones, we include six baseline methods with five popular GNN backbones in our experiments, and combine \alg{} with them under all possible combinations.
The included CIGL baselines can be generally divided into two categories: reweighting-based (i.e., Reweight~\citep{japkowicz2002class}, ReNode~\citep{chen2021renode}) and augmentation-based (i.e., Oversampling~\citep{japkowicz2002class}, SMOTE~\citep{chawla2002smote}, GraphSMOTE~\citep{zhao2021graphsmote}, and GraphENS~\citep{park2021graphens}).
\begin{itemize}
    \setlength\itemsep{-2pt}
    \item Reweight~\citep{japkowicz2002class} assigns minority classes with higher misclassification costs (i.e., weights in the loss function) by the inverse of the class frequency in the training set.
    \item ReNode~\citep{chen2021renode} measures the influence conflict of training nodes, and perform instance-wise node reweighting to alleviate the topology imbalance.
    \item Oversample~\citep{japkowicz2002class} augments minority classes with additional synthetic nodes by replication-base oversampling.
    \item SMOTE~\citep{chawla2002smote} synthesizes minority nodes by 1) randomly selecting a seed node, 2) finding its $k$-nearest neighbors in the feature space, and 3) performing linear interpolation between the seed and one of its $k$-nearest neighbors.
    \item GraphSMOTE~\citep{zhao2021graphsmote} extends SMOTE~\citep{chawla2002smote} to graph-structured data by 1) performing SMOTE in the low-dimensional embedding space of GNN and 2) utilizing a learnable edge predictor to generate better topology connections for synthetic nodes.
    \item GraphENS~\citep{park2021graphens} directly synthesize the whole ego network (node with its 1-hop neighbors) for minority classes by similarity-based ego network combining and saliency-based node mixing to prevent neighbor memorization.
\end{itemize}

\textbf{Baseline implementation details.}
We use the public implementations\footnote{\href{https://github.com/victorchen96/renode}{https://github.com/victorchen96/renode}}\footnote{\href{https://github.com/TianxiangZhao/GraphSmote}{https://github.com/TianxiangZhao/GraphSmote}}\footnote{\href{https://github.com/JoonHyung-Park/GraphENS}{https://github.com/JoonHyung-Park/GraphENS}} of the baseline methods for a fair comparison.
For ReNode~\citep{chen2021renode}, we use its transductive version and search hyperparameters among the lower bound of cosine annealing $w_{min}\in\{0.25, 0.5, 0.75\}$ and upper bound of the cosine annealing $w_{max}\in\{1.25, 1.5, 1.75\}$ following the original paper.
We set the teleport probability of PageRank $\alpha=0.15$ as given by the default setting in the released implementation.
As Oversample~\citep{cui2019class} and SMOTE~\citep{chawla2002smote} were not proposed to handle graph data, we adopt their enhanced versions provided by GraphSMOTE~\citep{zhao2021graphsmote}, which also duplicate the edges from the seed nodes to the synthesized nodes in order to connect them to the graph.
For GraphSMOTE~\citep{zhao2021graphsmote}, we use the version that predicts edges with binary values as it performs better than the variant with continuous edge predictions in many datasets.
For GraphENS~\citep{park2021graphens}, we follow the settings in the paper: $\text{Beta}(2,2)$ distribution is used to sample $\lambda$, the feature masking hyperparameter $k$ and temperature $\tau$ are tuned among $\{1,5,10\}$ and $\{1,2\}$, and the number of warmup epochs is set to 5.

\textbf{Combining \alg{} and baseline CIGL techniques.}
Since \alg{} only manipulates the graph data and remains independent of the model architecture, it seamlessly integrates with the aforementioned CIGL techniques. 
During each training epoch, \alg{} enhances the original graph $\gG$ using the current model $f$ and yields the augmented graph $\gG^*$.
Subsequently, other CIGL methods operate on the augmented graph $\gG^*$.
Specifically, loss function engineering methods (Reweight and ReNode) perform loss computation and backpropagation based on $\gG^*$, and data augmentation methods (Resampling, SMOTE, GSMOTE, GENS) carry out additional class-balancing operations on $\gG^*$, generating new minority nodes based on its structure.

\textbf{GNN backbone implementation details.}
We use \texttt{pytorch}~\citep{paszke2019pytorch} and \texttt{torch\_geometric}~\citep{fey2019torch-geo} to implement all five GNN backbones used in this paper, i.e., GCN~\citep{gcn}, GAT~\citep{gat}, GraphSAGE~\citep{sage}, APPNP~\citep{gasteiger2018appnp}, and GPRGNN~\citep{chien2020gprgnn}.
Most of our settings are aligned with prevailing works~\citep{park2021graphens,chen2021renode,song2022tam} to obtain fair and comparable results.
Specifically, we implement all GNNs' convolution layer with ReLU activation and dropout~\citep{srivastava2014dropout} with a dropping rate of 0.5 before the last layer.
For GAT, we set the number of attention heads to 4.
For APPNP and GPRGNN, we follow the best setting in the original paper and use 2 \texttt{APPNP}/\texttt{GPR\_prop} convolution layers with 64 hidden units.
Note that GraphENS's official implementation requires modifying the graph convolution for resampling and thus cannot be directly combined with APPNP and GPRGNN.
The teleport probability = 0.1 and the number of power iteration steps K = 10.
We search for the best architecture for other backbones from \#layers $l \in \{1,2,3\}$ and hidden dimension $d \in \{64, 128, 256\}$ based on the average of validation accuracy and F1 score.
We train each GNN for 2,000 epochs using Adam optimizer~\citep{kingma2014adam} with an initial learning rate of 0.01.
To achieve better convergence, we follow~\citep{park2021graphens} to use 5e-4 weight decay and adopt the \texttt{ReduceLROnPlateau} scheduler in Pytorch, which reduces the learning rate by half if the validation loss does not improve for 100 epochs.

\subsection{Evaluation Protocol}\label{sec:apd-setup-eval}

To evaluate the predictive performance on class-imbalanced data, we use two balanced metrics, i.e., balanced accuracy (\textit{BAcc.}) and macro-averaged F1 score (\textit{Macro-F1}). 
They compute accuracy/F1-score for each class independently and use the unweighted average mean as the final score, i.e., \textit{BAcc.} $= \frac{1}{m}\sum_{i=1}^m Acc(c_i)$, \textit{Macro-F1} $= \frac{1}{m}\sum_{i=1}^m F1(c_i)$.
Additionally, we use performance standard deviation (PerfStd) to evaluate the level of model predictive bias.
Formally, let $Acc(c_i)$ be the classification accuracy of class $c_i$, the PerfStd is defined as the standard deviation of the accuracy scores of all classes, i.e., $\sqrt{\frac{1}{m}\sum_{i=1}^{m}(Acc(c_i) - \textit{BAcc.})^2}$.
All the experiments are conducted on a Linux server with Intel$^\text{\textregistered}$ Xeon$^\text{\textregistered}$ Gold 6240R CPU and NVIDIA$^\text{\textregistered}$ Tesla V100 32GB GPU.

\section{Extended Discussions}
\label{sec:apd-discussion}

In this section, we present an ablation study \textbf{(\S\ref{sec:apd-dis-ablation})} validate the effectiveness and efficiency of the key modules, then we discuss how to further speed up \alg{} in practice \textbf{(\S\ref{sec:apd-dis-speedup})}; how to choose between \algp{} and \algt{} in practice \textbf{(\S\ref{sec:apd-dis-choose})}; and finally, the limitation and future works \textbf{(\S\ref{sec:apd-dis-limitation})}. 

\subsection{Ablation Study}
\label{sec:apd-dis-ablation}
We present an ablation study to validate the effectiveness and efficiency of the key modules in \alg{}. 
Specifically, for node risk estimation, we compare our total-variation-distance-based uncertainty with (i) the na\"ive random assignment that drawn uncertainty score from a uniform distribution $U(0, 1)$ and (ii) the information entropy $\mathbf{H}(Y) = -\sum_{y\in\mathcal{Y}}p(y) \log_2(p(y))$.
We substitute the original uncertainty metric with these aforementioned methods in \algp{}, and assess their impact on performance as well as the computational time required for uncertainty estimation.
It is worth noting that in practical implementation, the computation is parallelized on GPU (assuming sufficient GPU memory). Therefore, the computational time of a given uncertainty measure remains consistent across the three datasets we employed (Cora, CiteSeer, PubMed with IR=10).
The detailed results are presented in Table~\ref{tab:ablation-unc}, revealing that:
(i) Randomly assigned uncertainty scores significantly impede the performance of \alg{}, resulting in a large drop in both balanced accuracy and Marco-f1.
(ii) In comparison to our approach, employing information entropy as the node uncertainty score necessitates $\sim$2.3x computation time, yet the influence on performance remains marginal.

\begin{table}[h]
\caption{Ablation study on node risk estimation of \alg{}.}
\label{tab:ablation-unc}
\resizebox{\textwidth}{!}{%
\begin{tabular}{l|ll|ll|ll|c}
\toprule
\multicolumn{1}{c|}{\multirow{2}{*}{\textbf{Uncertainty}}} & \multicolumn{2}{c|}{\textbf{Cora}}                       & \multicolumn{2}{c|}{\textbf{CiteSeer}}                   & \multicolumn{2}{c|}{\textbf{PubMed}}                     & \textbf{Computation} \\ \cline{2-7}
\multicolumn{1}{c|}{}                                  & \multicolumn{1}{c}{BAcc} & \multicolumn{1}{c|}{Macro-F1} & \multicolumn{1}{c}{BAcc} & \multicolumn{1}{c|}{Macro-F1} & \multicolumn{1}{c}{BAcc} & \multicolumn{1}{c|}{Macro-F1} & \textbf{Time(ms)}    \\ \midrule
Random                                                 & 61.64$\pm$1.89           & 59.44$\pm$1.71                & 46.59$\pm$2.29           & 44.37$\pm$3.23                & 61.60$\pm$1.69           & 58.13$\pm$1.81                & 0.0249               \\
Information Entropy                                    & 65.18$\pm$1.68           & 63.11$\pm$1.91                & 51.87$\pm$2.96           & 50.36$\pm$3.43                & 67.72$\pm$1.27           & 67.19$\pm$1.57                & 0.1257               \\
\rowcolor{Gray} TVDistance (ours)                                      & 65.54$\pm$1.25           & 63.28$\pm$1.07                & 52.65$\pm$1.08           & 51.55$\pm$1.28                & 68.62$\pm$0.77           & 67.16$\pm$1.53                & 0.0543               \\ \bottomrule
\end{tabular}%
}
\end{table}

Further, we conduct an ablation study for our posterior likelihood estimation strategy by comparing our $0^\text{th}$-order (\algp{}) and $1^\text{st}$-order (\algt{}) likelihood estimation methods with the random method that assigns (unnormalized) node-class likelihood by drawing from a uniform distribution $U(0, 1)$.
Results are shown in Table~\ref{tab:ablation-sim}.
We can observe that the random method significantly worsens the predictive performance on all CIGL tasks.
Altogether, the ablation study results confirm the effectiveness and efficiency of the design of \alg{}, showcasing its ability to deliver strong performance with minimal computational overhead.

\begin{table}[h]
\caption{Ablation study on posterior likelihood estimation of \alg{}.}
\label{tab:ablation-sim}
\resizebox{\textwidth}{!}{%
\begin{tabular}{l|ll|ll|ll|c}
\toprule
\multicolumn{1}{c|}{\multirow{2}{*}{\textbf{Estimation}}} & \multicolumn{2}{c|}{\textbf{Cora}}                       & \multicolumn{2}{c|}{\textbf{CiteSeer}}                   & \multicolumn{2}{c|}{\textbf{PubMed}}                     & \textbf{Computation} \\ \cline{2-7}
\multicolumn{1}{c|}{}                                  & \multicolumn{1}{c}{BAcc} & \multicolumn{1}{c|}{Macro-F1} & \multicolumn{1}{c}{BAcc} & \multicolumn{1}{c|}{Macro-F1} & \multicolumn{1}{c}{BAcc} & \multicolumn{1}{c|}{Macro-F1} & \textbf{Time(ms)}    \\ \midrule
Random                                                 & 63.85$\pm$2.17           & 61.94$\pm$2.68                & 46.51$\pm$3.27           & 41.70$\pm$4.33                & 64.32$\pm$1.23           & 53.58$\pm$2.48                & 0.0883               \\
\rowcolor{Gray} 0th-order (\algp{})                                   & 65.54$\pm$1.25           & 63.28$\pm$1.07                & 52.65$\pm$1.08           & 51.55$\pm$1.28                & 68.62$\pm$0.77           & 67.16$\pm$1.53                & 0.1251               \\
\rowcolor{Gray} 1st-order (\algt{})                                     & 69.80$\pm$1.30           & 68.68$\pm$1.49                & 55.37$\pm$1.39           & 54.94$\pm$1.44                & 67.57$\pm$3.22           & 64.40$\pm$3.68                & 0.3030               \\ \bottomrule
\end{tabular}%
}
\end{table}

\subsection{On the Further Speedup of \alg{}}
\label{sec:apd-dis-speedup}
\begin{figure}[h]
  \centering
  \subfigure[Influence of \alg{} granularity on BAcc.]{
    \includegraphics[width=0.48\linewidth]{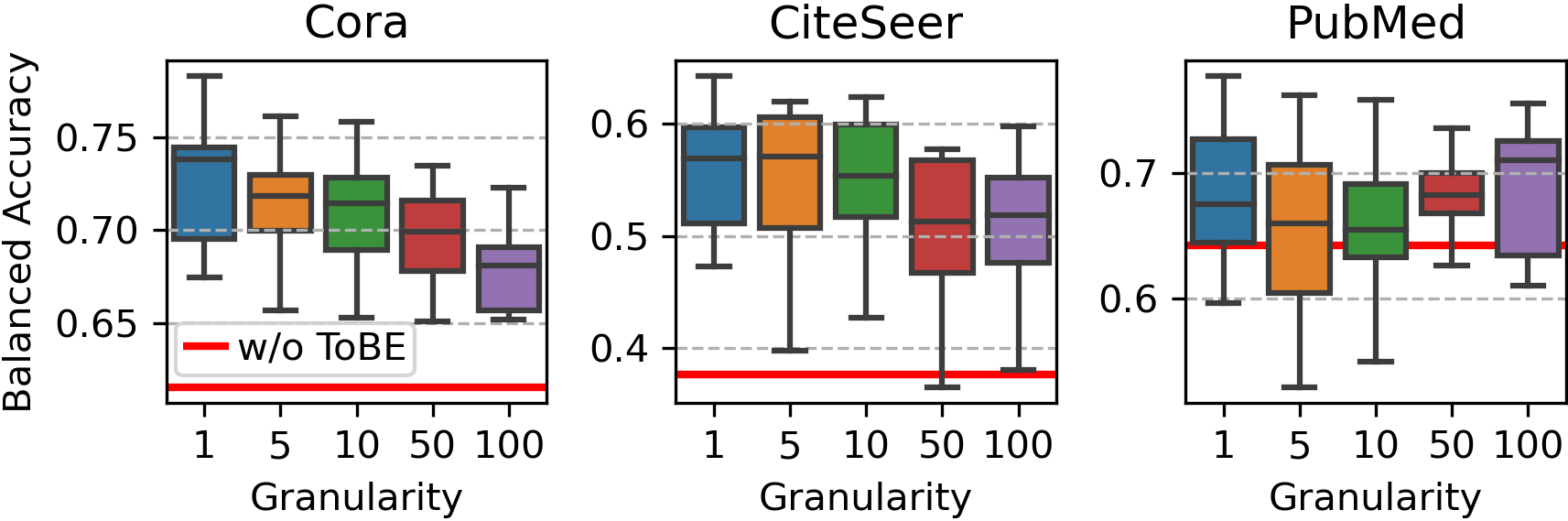}
  }
  \hfill
  \subfigure[Influence of \alg{} granularity on Macro-F1.]{
    \includegraphics[width=0.48\linewidth]{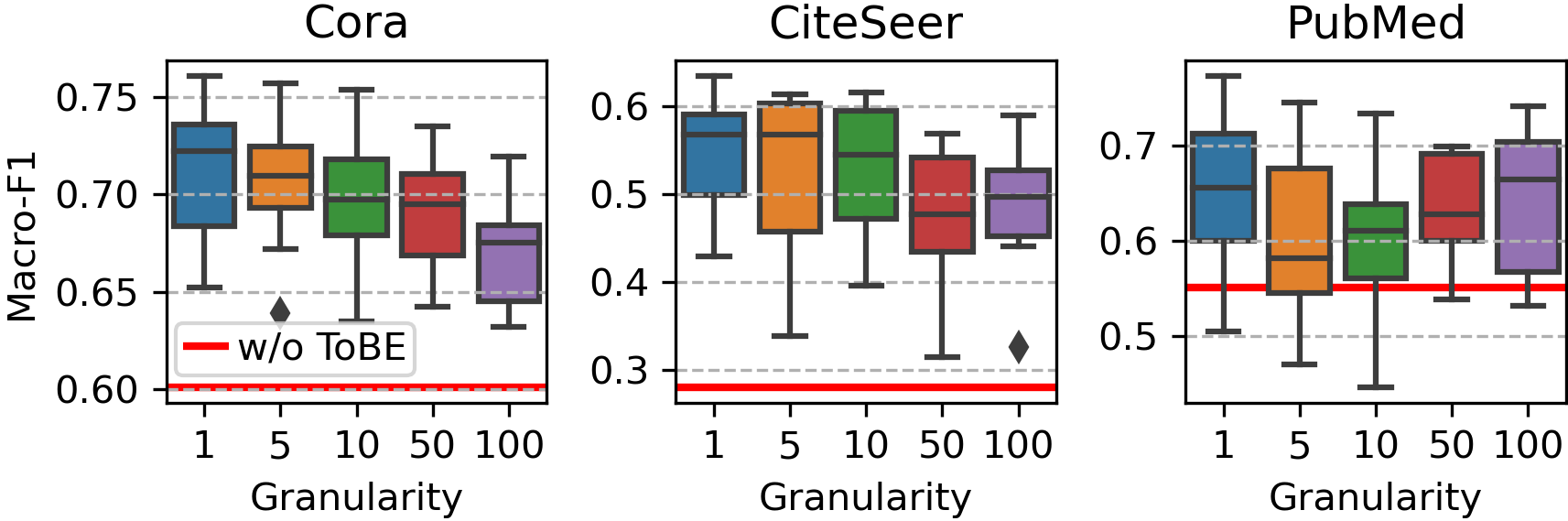}
  }
  \caption{
  Influence of the \alg{} granularity (i.e., the number of iterations per each use of \alg{}).
  Note that this brings a linear speedup ratio in practice, e.g., granularity = 100 $\Leftrightarrow$ 100x speedup.
  }
  \label{fig:granularity}
\end{figure}

As stated in the paper, thanks to its simple and efficient design, \alg{} can be integrated into the GNN training process to perform dynamic topology augmentation based on the training state.
By default, we run \alg{} in every iteration of GNN training, i.e., the granularity of applying \alg{} is 1, as described in Alg.~\ref{alg:main}.
However, we note that in practice, this granularity can be increased to further reduce the cost of applying \alg{}.
This operation can result in a significant linear speedup ratio: setting the granularity to $N$ reduces the computational overhead of \alg{} to $1/N$ of the original (i.e., $N$x speedup ratio), with minor performance degradation.
This could be helpful for scaling \alg{} to large-scale graphs in practice.
In this section, we design experiments to validate the influence of different \alg{} granularity (i.e., the number of iterations per each use of \alg{}) in real-world CIGL tasks.
We set the granularity to 1/5/10/50/100 and test the performance of \alg{}$_T$ with a vanilla GCN classifier on the Cora/CiteSeer/PubMed dataset with an imbalance ratio of 10.
Fig.~\ref{fig:granularity} shows the empirical results from 10 independent runs. 
The red horizontal line in each subfigure represents the baseline (vanilla GCN) performance.
It can be observed that setting a larger \alg{} granularity is an effective way to further speed up \alg{} in practice.
The performance drop of adopting this trick is relatively minor, especially considering the significant linear speedup ratio it brings.
The predictive performance boost brought by \alg{} is still significant even with a large granularity at 100 (i.e., with 100x \alg{} speedup).

\subsection{Choosing between \algp{} and \algt{}}
\label{sec:apd-dis-choose}

In this section, we summarize the strengths and limitations of \algp{} and \algt{}, and give suggestions for choosing between them in practice.
In short, we recommend using \algt{} to achieve better classification performance. But in case that computational resources are limited, \algp{} can serve as a more efficient alternative.
The reasons are as follows:

\textbf{Performance.} 
We observe a noticeable performance gap between \algp{} and \algt{}, wherein \algt{} consistently demonstrates superior classification performance due to its incorporation of local topological structure. Across the 15 scenarios outlined in Table~\ref{tab:main} (the best scores for 3 datasets x 5 GNN backbones):
(1) \algt{} outperforms \algp{} significantly in 12 out of 15 scenarios for BAcc/F1 scores, with an average F1 advantage of 1.692 in the 11 leading cases.
(2) Conversely, \algp{} exhibits a less pronounced advantage in the instances where it outperforms \algt{}, with an average advantage of 0.518 in the 4 leading scenarios\footnote{Despite \algt{} holding a relative performance edge, both \algp{} and \algt{} substantially enhance the performance of the \textit{best-performing} CIGL baseline methods. Over the 15 scenarios, \algp{} yields an average improvement of 4.789/5.848 in the best BAcc/F1, while \algt{} brings an average improvement of 5.805/6.950.}.

\textbf{Efficiency.}
On the other hand, it's worth noting that \algt{} generally incurs higher time and space complexity compared to \algp{}. Specifically, \algp{} demonstrates linear complexity concerning the number of nodes, whereas \algt{} exhibits linear growth in complexity with the number of edges. Given that real-world graph data often features a significantly larger number of edges than nodes, \algp{} is usually the more efficient option (especially for densely connected graphs).

\subsection{Limitations and Future Works}
\label{sec:apd-dis-limitation}

One potential limitation of the proposed \alg{} framework is its reliance on exploiting model prediction for risk and likelihood estimation.
This strategy may not provide accurate estimation when the model itself exhibits extremely poor predictive performance.
However, this rarely occurs in practice and can be prevented by more careful fine-tuning of parameters and model architectures.
In addition to this, as described in Section~\ref{sec:methodology}, we adopt several fast measures to estimate node uncertainty, prediction risk, and posterior likelihood for the sake of efficiency.
Other techniques for such purposes (e.g., deterministic~\citep{liu2020uncertainty,zhao2020uncertainty}/Bayesian~\citep{zhang2019bayesian,hasanzadeh2020bayesian}/Jackknife~\citep{kang2022jurygcn} uncertainty estimation) can be easily integrated into the proposed \alg{} framework, although the computational efficiency might be a major bottleneck. How to exploit alternative uncertainty/risk measures while retaining computational efficiency is an interesting future direction.

Beyond the class imbalance in the node label distribution, graph data can also exhibit multi-facet skewness in other aspects.
For instance, class imbalance may also exist in edge-level (e.g., in edge classification/prediction~\citep{cai2021line,pandey2019comprehensive}) and graph-level (e.g., in graph classification/alignment~\citep{zeng2023parrot,zeng2024hierarchical,yan2021bright}).
Handling class imbalance can be even more challenging on evolving/dynamic graphs~\citep{fu2021sdg,yan2021dynamic}.
Beyond the quantity imbalance among classes, skewness may also exists in the topological structure, such as degree imbalance~\citep{kang2022rawlsgcn}, and motif-level imbalance~\citep{zhao2022topoimb,zeng2023generative}.
How to jointly consider the multi-facet node/edge/graph-level imbalance to benefit more graph learning tasks is an exciting future direction.
Finally, recent work on fairness-aware graph learning~\citep{kang2020inform,fu2023fair} has observed that the topology of the graph can also introduce bias/discrimination towards certain groups: extending BAT's concept to mitigate group unfairness through topology augmentation poses an intriguing future direction.

\section{Additional Experimental Results and Analysis}
\label{sec:apd-results}

\subsection{Results on Additional Large-scale Graphs}
\label{sec:apd-res-large}

To further test the scalability of \alg{} and the baseline CIGL methods, we extended the experiment to two large-scale graph datasets, \textit{CoraFull} (19,793 nodes, 126,842 edges)~\citep{bojchevski2017corafull} and \textit{arXiv} (169,343 nodes, 1,166,243 edges)~\citep{hu2020ogb}.
To ensure comprehensive and fair comparisons, we followed the same protocol as Table~\ref{tab:main} to evaluate six baseline methods and the performance gains brought by \alg{}. We used GCN as the backbone. 
It is worth noting that CIGL on these large graphs is a highly challenging task due to (i) the inherent task complexity (with 70/172 classes), (ii) the label scarcity for numerous minority classes, and (iii) the requirement for the algorithm's scalability. This may be the reason why most existing works (ReNode~\citep{chen2021renode}, GraphSMOTE~\citep{zhao2021graphsmote}, GraphENS~\citep{park2021graphens}, TAM~\cite{song2022tam}, LTE4G~\cite{yun2022lte4g}, etc.) did not consider these large datasets. 
Nevertheless, we conduct the experiments report the results in Table~\ref{tab:apd-red-large}.
We can observe that:
\begin{itemize}
    \setlength\itemsep{-2pt}
    \item \alg{} can scale to large-scale CIGL tasks, consistently and significantly enhancing the performance of various CIGL baselines (overall with ~10\% relative performance gain).
    \item The linear complexity of \alg{} makes it scalable to large graphs, while GraphENS, ReNode, and GraphSMOTE face scalability issues when applied to the arXiv dataset (with 169,343 nodes) due to their $O(n^2)$ space complexity.
\end{itemize}

\begin{table*}[t]
\centering
\caption{
    Experiments on large-scale graphs, the numbers in parentheses are the performance gain brought by \alg{}.
}
\label{tab:apd-red-large}
\begin{threeparttable}
% \centering
\resizebox{\columnwidth}{!}{%
\begin{tabular}{c|c|c|lllllll}
\hline
\multirow{2}{*}{\textbf{Metric}} & \multirow{2}{*}{\textbf{Dataset}} & \multirow{2}{*}{\textbf{Setting}} & \multicolumn{7}{c}{\textbf{CIGL Baseline}} \\ \cline{4-10} 
 &  &  & \multicolumn{1}{c}{\textbf{ERM}} & \multicolumn{1}{c}{\textbf{Reweight}} & \multicolumn{1}{c}{\textbf{ReNode}} & \multicolumn{1}{c}{\textbf{Resample}} & \multicolumn{1}{c}{\textbf{SMOTE}} & \multicolumn{1}{c}{\textbf{GraphSMOTE}} & \multicolumn{1}{c}{\textbf{GraphENS}} \\ \hline
\multirow{6}{*}{\rotatebox{90}{BAcc.$\uparrow$}} & \multirow{3}{*}{CoraFull} & Base & 36.12 & 40.93 & 40.10 & 35.71 & 35.59 & 39.25 & 43.76 \\
 &  & \cellcolor{Gray}+\algp{} & \cellcolor{Gray}39.40 (+3.28) & \cellcolor{Gray}43.81 (+2.88) & \cellcolor{Gray}42.78 (+2.68) & \cellcolor{Gray}40.77 (+5.05) & \cellcolor{Gray}40.31 (+4.72) & \cellcolor{Gray}43.71 (+4.46) & \cellcolor{Gray}46.20 (+2.44) \\
 &  & \cellcolor{Gray}+\algt{} & \cellcolor{Gray}40.88 (+4.76) & \cellcolor{Gray}43.77 (+2.84) & \cellcolor{Gray}42.82 (+2.72) & \cellcolor{Gray}40.67 (+4.95) & \cellcolor{Gray}41.04 (+5.46) & \cellcolor{Gray}43.70 (+4.45) & \cellcolor{Gray}47.19 (+3.44) \\ \cline{2-10} 
 & \multirow{3}{*}{OGBN-arXiv} & Base & 32.20 & 35.69 & OOM & 32.24 & 32.18 & OOM & OOM \\
 &  & \cellcolor{Gray}+\algp{} & \cellcolor{Gray}34.36 (+2.16) & \cellcolor{Gray}37.59 (+1.90) & \cellcolor{Gray}OOM & \cellcolor{Gray}37.20 (+4.96) & \cellcolor{Gray}36.88 (+4.69) & \cellcolor{Gray}OOM & \cellcolor{Gray}OOM \\
 &  & \cellcolor{Gray}+\algt{} & \cellcolor{Gray}36.57 (+4.37) & \cellcolor{Gray}39.28 (+3.60) & \cellcolor{Gray}OOM & \cellcolor{Gray}37.36 (+5.12) & \cellcolor{Gray}37.50 (+5.32) & \cellcolor{Gray}OOM & \cellcolor{Gray}OOM \\ \hline
\multirow{6}{*}{\rotatebox{90}{Macro-F1$\uparrow$}} & \multirow{3}{*}{CoraFull} & Base & 33.54 & 38.86 & 37.98 & 32.84 & 32.70 & 37.70 & 41.28 \\
 &  & \cellcolor{Gray}+\algp{} & \cellcolor{Gray}37.25 (+3.71) & \cellcolor{Gray}41.21 (+2.35) & \cellcolor{Gray}40.58 (+2.61) & \cellcolor{Gray}39.04 (+6.19) & \cellcolor{Gray}38.35 (+5.65) & \cellcolor{Gray}41.25 (+3.55) & \cellcolor{Gray}43.90 (+2.61) \\
 &  & \cellcolor{Gray}+\algt{} & \cellcolor{Gray}38.58 (+5.04) & \cellcolor{Gray}41.16 (+2.30) & \cellcolor{Gray}40.37 (+2.39) & \cellcolor{Gray}38.48 (+5.64) & \cellcolor{Gray}39.18 (+6.48) & \cellcolor{Gray}41.75 (+4.05) & \cellcolor{Gray}44.61 (+3.33) \\ \cline{2-10} 
 & \multirow{3}{*}{OGBN-arXiv} & Base & 29.90 & 32.14 & OOM & 30.16 & 29.96 & OOM & OOM \\
 &  & \cellcolor{Gray}+\algp{} & \cellcolor{Gray}32.42 (+2.52) & \cellcolor{Gray}34.51 (+2.38) & \cellcolor{Gray}OOM & \cellcolor{Gray}34.50 (+4.34) & \cellcolor{Gray}34.44 (+4.48) & \cellcolor{Gray}OOM & \cellcolor{Gray}OOM \\
 &  & \cellcolor{Gray}+\algt{} & \cellcolor{Gray}33.99 (+4.08) & \cellcolor{Gray}34.78 (+2.64) & \cellcolor{Gray}OOM & \cellcolor{Gray}34.55 (+4.38) & \cellcolor{Gray}34.69 (+4.74) & \cellcolor{Gray}OOM & \cellcolor{Gray}OOM \\ \hline
\end{tabular}
}
\begin{tablenotes}
    \tiny\item[]\hspace*{-12pt}
    *OOM: Out-Of-Memory on a NVIDIA$^\text{\textregistered}$ Tesla V100 32GB GPU.
\end{tablenotes}
\end{threeparttable}
\end{table*}

\subsection{Comparison with Additional CIGL Baselines}
\label{sec:apd-res-additional}

In the main results, we chose representative \textit{model-independent} CIGL baselines to test \alg{}'s ability to cooperate and boost various CIGL techniques.
In this section, we compare \alg{} with other representative CIGL techniques that are either model-dependent or with design constraints making them incompatible with \alg{}.
Specifically, we further include three CIGL baselines LTE4G~\citep{yun2022lte4g}, GraphMixup~\citep{wu2022graphmixup}, TAM~\citep{song2022tam}, and compare them with \alg{} on seven datasets (including the newly introduced large-scale graphs \textit{CoraFull} and \textit{arXiv} in Section~\ref{sec:apd-res-large}).
We use GCN as the backbone, and follow the main experiment protocol used in Table~\ref{tab:main} to ensure fair and comparable results.

The results are reported in Table~\ref{tab:apd-res-additional}.
In short, we observe that: 
(i) \alg{} consistently demonstrates better performance and scalability compared to the additional baselines.
(ii) Some methods (such as GraphMixup and LTE4G) exhibit higher space complexity and thus cannot scale to large datasets.
(iii) Among the baseline methods, LTE4G achieves the overall best performance due to its three-stage training (encoder pre-training, expert training, student training) and divide-and-conquer strategy. We note that its complex training and inference strategies introduce additional computational overhead and make it challenging to integrate with other methods. In comparison, \alg{} has better compatibility and can further achieve better performance with existing class-rebalancing techniques.

\begin{table}[h]
\centering
\caption{
Comparison with independent CIGL baselines.
We use \textbf{bold}/\textit{italics} to mark the best/second-best results.
}
\label{tab:apd-res-additional}
\begin{threeparttable}
\begin{tabular}{c|l|ccccccc}
\hline
\multirow{2}{*}{\textbf{Metric}} & \multicolumn{1}{c|}{\multirow{2}{*}{\textbf{Method}}} & \multicolumn{7}{c}{\textbf{Dataset}} \\ \cline{3-9} 
 & \multicolumn{1}{c|}{} & \textbf{Cora} & \textbf{CiteSeer} & \textbf{PubMed} & \textbf{CS} & \textbf{Physics} & \textbf{CoraFull} & \textbf{arXiv} \\ \hline
\multirow{5}{*}{\rotatebox{90}{BAcc.$\uparrow$}} & GCN & 61.6 & 37.6 & 64.2 & 75.4 & 80.1 & 32.2 & 36.1 \\
 & GraphMixup & 64.1 & 50.8 & OOM & OOM & OOM & OOM & OOM \\
 & LTE4G & \textit{68.0} & \textit{51.7} & 63.7 & \textit{77.2} & 81.7 & \textbf{36.6} & OOM \\
 & TAM & 65.3 & 50.1 & \textit{65.7} & 77.0 & \textit{82.8} & 35.0 & \textit{36.3} \\
 & \cellcolor{Gray} \alg{} (Ours) & \cellcolor{Gray} \textbf{69.8} & \cellcolor{Gray} \textbf{55.4} & \cellcolor{Gray} \textbf{68.6} & \cellcolor{Gray} \textbf{82.6} & \cellcolor{Gray} \textbf{87.6} & \cellcolor{Gray} \textbf{36.6} & \cellcolor{Gray} \textbf{40.1} \\ \hline
\multirow{5}{*}{\rotatebox{90}{Macro-F1$\uparrow$}} & GCN & 60.1 & 28.1 & 55.1 & 72.7 & 80.7 & 33.5 & 29.9 \\
 & GraphMixup & 62.7 & 47.3 & OOM & OOM & OOM & OOM & OOM \\
 & LTE4G & \textit{67.1} & \textit{49.7} & 61.7 & \textit{76.7} & 82.1 & \textit{37.7} & OOM \\
 & TAM & 65.7 & 43.4 & \textit{63.5} & 75.2 & \textit{82.3} & 32.8 & \textit{31.2} \\
 & \cellcolor{Gray} \alg{} (Ours) & \cellcolor{Gray} \textbf{68.7} & \cellcolor{Gray} \textbf{54.9} & \cellcolor{Gray} \textbf{67.2} & \cellcolor{Gray} \textbf{78.6} & \cellcolor{Gray} \textbf{88.8} & \cellcolor{Gray} \textbf{38.6} & \cellcolor{Gray} \textbf{34.0} \\ \hline
\end{tabular}
\begin{tablenotes}
    \tiny\item[]\hspace*{-12pt}
    *OOM: Out-Of-Memory on a NVIDIA$^\text{\textregistered}$ Tesla V100 32GB GPU.
\end{tablenotes}
\end{threeparttable}
\end{table}

\subsection{Full Empirical Results with Additional GNN Backbones}\label{sec:apd-res-full}

Due to space limitation, we report the key results of our experiments in Table~\ref{tab:main} and~\ref{tab:main-varyimb}.
We now provide complete results for all settings with the standard error of 5 independent runs.
Specifically, Table~\ref{tab:main_ext_bacc} \& \ref{tab:main_ext_f1} \& \ref{tab:main_ext_disp} complement Table~\ref{tab:main}, and Table~\ref{tab:main_ext_varyimb} complements Table~\ref{tab:main-varyimb}.
We further include APPNP~\citep{gasteiger2018appnp} and GPRGNN~\citep{chien2020gprgnn} as additional GNN backbones for supported CIGL techniques.
Note that the official codebase of \cite{park2021graphens} only implemented GraphENS using modified (to enable saliency-based mixup) GCN, GAT, and GraphSAGE backbones, and extending it to other GNN backbones is difficult.
The results indicate that \alg{} can consistently boost various CIGL baselines with all GNN backbones under different types and levels of class imbalance, which aligns with our conclusions in the paper.

\input{sections/tables/tab_ext_bacc}
\input{sections/tables/tab_ext_f1}
\input{sections/tables/tab_ext_disp}
\input{sections/tables/tab_ext_varyimb}

%% file: sections/tables/tab_ext_bacc.tex
\begin{table*}[]
% \footnotesize
% \scriptsize
\caption{
    Balanced accuracy of combining \alg{} with 6 IGL baselines $\times$ 5 GNN backbones.
}
\label{tab:main_ext_bacc}
\resizebox{\textwidth}{!}{
\begin{tabular}{cl|lgg|lgg|lgg}
\hline
\multicolumn{2}{c|}{\textbf{Dataset (IR=10)}}                                    & \multicolumn{3}{c|}{\textbf{Cora}}                                                                                  & \multicolumn{3}{c|}{\textbf{CiteSeer}}                                                                              & \multicolumn{3}{c}{\textbf{PubMed}}                                                                                    \\ \hline
\multicolumn{2}{c|}{\textbf{Metric: BAcc.$\uparrow$}}                              & \multicolumn{1}{c}{Base}   & \multicolumn{1}{c}{+ \algp{}} & \multicolumn{1}{c|}{+ \algt{}} & \multicolumn{1}{c}{Base}   & \multicolumn{1}{c}{+ \algp{}} & \multicolumn{1}{c|}{+ \algt{}} & \multicolumn{1}{c}{Base}       & \multicolumn{1}{c}{+ \algp{}} & \multicolumn{1}{c}{+ \algt{}} \\ \hline
\multicolumn{1}{c|}{\multirow{8}{*}{\textbf{\rotatebox{90}{GCN}}}}    & Vanilla  & 61.56\tiny{$\pm$1.24}      & \ul{65.54}\tiny{$\pm$1.25}                & \textbf{69.80}\tiny{$\pm$1.30}             & 37.62\tiny{$\pm$1.61}      & \ul{52.65}\tiny{$\pm$1.08}                & \textbf{55.37}\tiny{$\pm$1.39}             & 64.23\tiny{$\pm$1.55}          & \textbf{68.62}\tiny{$\pm$0.77}            & \ul{67.57}\tiny{$\pm$3.22}                \\
\multicolumn{1}{c|}{}                                                 & Reweight & 67.65\tiny{$\pm$0.64}      & \ul{70.97}\tiny{$\pm$1.28}                & \textbf{72.14}\tiny{$\pm$0.72}             & 42.49\tiny{$\pm$2.66}      & \ul{57.91}\tiny{$\pm$0.98}                & \textbf{58.36}\tiny{$\pm$1.09}             & 71.20\tiny{$\pm$2.33}          & \textbf{74.19}\tiny{$\pm$1.12}            & \ul{73.37}\tiny{$\pm$0.96}                \\
\multicolumn{1}{c|}{}                                                 & ReNode   & 66.60\tiny{$\pm$1.33}      & \ul{71.37}\tiny{$\pm$0.62}                & \textbf{71.84}\tiny{$\pm$1.25}             & 42.57\tiny{$\pm$1.05}      & \ul{57.47}\tiny{$\pm$0.62}                & \textbf{59.28}\tiny{$\pm$0.59}             & 71.52\tiny{$\pm$2.16}          & \textbf{73.20}\tiny{$\pm$0.71}            & \ul{72.53}\tiny{$\pm$1.62}                \\
\multicolumn{1}{c|}{}                                                 & Resample & 59.48\tiny{$\pm$1.53}      & \ul{72.51}\tiny{$\pm$0.68}                & \textbf{74.24}\tiny{$\pm$0.91}             & 39.15\tiny{$\pm$2.05}      & \ul{57.90}\tiny{$\pm$0.33}                & \textbf{58.78}\tiny{$\pm$1.44}             & 64.97\tiny{$\pm$1.94}          & \ul{72.53}\tiny{$\pm$0.85}                & \textbf{72.87}\tiny{$\pm$1.16}            \\
\multicolumn{1}{c|}{}                                                 & SMOTE    & 58.27\tiny{$\pm$1.05}      & \ul{72.16}\tiny{$\pm$0.53}                & \textbf{73.89}\tiny{$\pm$1.06}             & 39.27\tiny{$\pm$1.90}      & \ul{60.06}\tiny{$\pm$0.81}                & \textbf{61.97}\tiny{$\pm$1.19}             & 64.41\tiny{$\pm$1.95}          & \textbf{73.17}\tiny{$\pm$0.84}            & \ul{73.13}\tiny{$\pm$0.77}                \\
\multicolumn{1}{c|}{}                                                 & GSMOTE   & 67.99\tiny{$\pm$1.37}      & \ul{68.52}\tiny{$\pm$0.81}                & \textbf{71.55}\tiny{$\pm$0.50}             & 45.05\tiny{$\pm$1.95}      & \textbf{57.68}\tiny{$\pm$1.03}            & \ul{57.65}\tiny{$\pm$1.18}                 & \ul{73.99}\tiny{$\pm$0.88}     & 73.09\tiny{$\pm$1.30}                     & \textbf{76.57}\tiny{$\pm$0.42}            \\
\multicolumn{1}{c|}{}                                                 & GENS     & 70.12\tiny{$\pm$0.43}      & \ul{72.22}\tiny{$\pm$0.57}                & \textbf{72.58}\tiny{$\pm$0.58}             & 56.01\tiny{$\pm$1.17}      & \ul{60.60}\tiny{$\pm$0.63}                & \textbf{62.67}\tiny{$\pm$0.42}             & 73.66\tiny{$\pm$1.04}          & \ul{76.11}\tiny{$\pm$0.60}                & \textbf{76.91}\tiny{$\pm$1.03}            \\ \cline{2-11} 
\multicolumn{1}{c|}{}                                                 & Best     & 70.12                      & \ul{72.51}                                & \textbf{74.24}                             & 56.01                      & \ul{60.60}                                & \textbf{62.67}                             & 73.99                          & \ul{76.11}                                & \textbf{76.91}                            \\ \hline
\multicolumn{1}{c|}{\multirow{8}{*}{\textbf{\rotatebox{90}{GAT}}}}    & Vanilla  & 61.53\tiny{$\pm$1.13}      & \ul{66.27}\tiny{$\pm$0.83}                & \textbf{70.13}\tiny{$\pm$1.07}             & 39.25\tiny{$\pm$1.84}      & \ul{55.66}\tiny{$\pm$1.23}                & \textbf{60.34}\tiny{$\pm$1.66}             & 65.46\tiny{$\pm$0.69}          & \ul{73.19}\tiny{$\pm$0.86}                & \textbf{74.75}\tiny{$\pm$1.18}            \\
\multicolumn{1}{c|}{}                                                 & Reweight & 66.94\tiny{$\pm$1.24}      & \textbf{71.80}\tiny{$\pm$0.48}            & \ul{71.61}\tiny{$\pm$0.85}                 & 41.29\tiny{$\pm$3.39}      & \ul{59.33}\tiny{$\pm$0.51}                & \textbf{61.23}\tiny{$\pm$0.99}             & 68.37\tiny{$\pm$1.41}          & \textbf{75.30}\tiny{$\pm$1.07}            & \ul{74.52}\tiny{$\pm$1.14}                \\
\multicolumn{1}{c|}{}                                                 & ReNode   & 66.81\tiny{$\pm$0.98}      & \textbf{72.14}\tiny{$\pm$1.24}            & \ul{70.31}\tiny{$\pm$1.38}                 & 43.25\tiny{$\pm$1.78}      & \ul{58.26}\tiny{$\pm$1.98}                & \textbf{59.05}\tiny{$\pm$0.88}             & 71.18\tiny{$\pm$2.13}          & \textbf{75.55}\tiny{$\pm$1.01}            & \ul{75.22}\tiny{$\pm$0.84}                \\
\multicolumn{1}{c|}{}                                                 & Resample & 57.76\tiny{$\pm$1.73}      & \ul{71.90}\tiny{$\pm$0.88}                & \textbf{73.29}\tiny{$\pm$1.08}             & 35.97\tiny{$\pm$1.42}      & \ul{60.10}\tiny{$\pm$1.26}                & \textbf{60.33}\tiny{$\pm$0.75}             & 65.14\tiny{$\pm$0.86}          & \ul{73.27}\tiny{$\pm$0.61}                & \textbf{73.89}\tiny{$\pm$0.40}            \\
\multicolumn{1}{c|}{}                                                 & SMOTE    & 58.81\tiny{$\pm$0.64}      & \ul{70.50}\tiny{$\pm$0.44}                & \textbf{72.19}\tiny{$\pm$0.75}             & 36.95\tiny{$\pm$1.86}      & \ul{60.59}\tiny{$\pm$1.19}                & \textbf{62.36}\tiny{$\pm$1.18}             & 64.81\tiny{$\pm$1.47}          & \ul{73.90}\tiny{$\pm$0.68}                & \textbf{74.08}\tiny{$\pm$0.51}            \\
\multicolumn{1}{c|}{}                                                 & GSMOTE   & 64.68\tiny{$\pm$1.02}      & \ul{69.29}\tiny{$\pm$1.82}                & \textbf{71.14}\tiny{$\pm$0.96}             & 41.82\tiny{$\pm$1.14}      & \ul{56.11}\tiny{$\pm$1.23}                & \textbf{57.71}\tiny{$\pm$2.58}             & 68.72\tiny{$\pm$1.69}          & \textbf{74.65}\tiny{$\pm$0.65}            & \ul{74.41}\tiny{$\pm$1.57}                \\
\multicolumn{1}{c|}{}                                                 & GENS     & 69.76\tiny{$\pm$0.45}      & \ul{70.63}\tiny{$\pm$0.40}                & \textbf{71.02}\tiny{$\pm$1.22}             & 51.50\tiny{$\pm$2.21}      & \ul{60.95}\tiny{$\pm$1.51}                & \textbf{63.49}\tiny{$\pm$0.75}             & 73.13\tiny{$\pm$1.18}          & \ul{74.34}\tiny{$\pm$0.35}                & \textbf{75.65}\tiny{$\pm$0.82}            \\ \cline{2-11} 
\multicolumn{1}{c|}{}                                                 & Best     & 69.76                      & \ul{72.14}                                & \textbf{73.29}                             & 51.50                      & \ul{60.95}                                & \textbf{63.49}                             & 73.13                          & \ul{75.55}                                & \textbf{75.65}                            \\ \hline
\multicolumn{1}{c|}{\multirow{8}{*}{\textbf{\rotatebox{90}{SAGE}}}}   & Vanilla  & 59.17\tiny{$\pm$1.23}      & \ul{66.24}\tiny{$\pm$0.92}                & \textbf{66.53}\tiny{$\pm$0.80}             & 42.96\tiny{$\pm$0.28}      & \textbf{54.99}\tiny{$\pm$2.51}            & \ul{53.18}\tiny{$\pm$2.90}                 & 67.56\tiny{$\pm$0.84}          & \ul{75.31}\tiny{$\pm$0.93}                & \textbf{77.38}\tiny{$\pm$0.68}            \\
\multicolumn{1}{c|}{}                                                 & Reweight & 63.76\tiny{$\pm$0.89}      & \ul{70.15}\tiny{$\pm$1.15}                & \textbf{71.14}\tiny{$\pm$0.84}             & 45.91\tiny{$\pm$2.05}      & \textbf{57.95}\tiny{$\pm$0.73}            & \ul{55.90}\tiny{$\pm$0.93}                 & 68.03\tiny{$\pm$1.69}          & \ul{74.56}\tiny{$\pm$0.41}                & \textbf{75.39}\tiny{$\pm$0.38}            \\
\multicolumn{1}{c|}{}                                                 & ReNode   & 65.32\tiny{$\pm$1.07}      & \ul{71.31}\tiny{$\pm$1.29}                & \textbf{71.54}\tiny{$\pm$0.85}             & 48.55\tiny{$\pm$2.31}      & \ul{56.32}\tiny{$\pm$0.40}                & \textbf{56.49}\tiny{$\pm$1.73}             & 69.08\tiny{$\pm$2.04}          & \ul{74.24}\tiny{$\pm$0.20}                & \textbf{75.28}\tiny{$\pm$0.69}            \\
\multicolumn{1}{c|}{}                                                 & Resample & 57.77\tiny{$\pm$1.35}      & \ul{71.24}\tiny{$\pm$1.08}                & \textbf{73.01}\tiny{$\pm$1.02}             & 39.37\tiny{$\pm$1.40}      & \ul{61.41}\tiny{$\pm$1.11}                & \textbf{61.93}\tiny{$\pm$1.40}             & 69.22\tiny{$\pm$1.28}          & \ul{74.91}\tiny{$\pm$1.09}                & \textbf{75.80}\tiny{$\pm$0.39}            \\
\multicolumn{1}{c|}{}                                                 & SMOTE    & 58.81\tiny{$\pm$1.97}      & \ul{70.31}\tiny{$\pm$1.35}                & \textbf{73.02}\tiny{$\pm$2.29}             & 38.42\tiny{$\pm$1.69}      & \ul{64.14}\tiny{$\pm$0.75}                & \textbf{66.35}\tiny{$\pm$0.70}             & 64.96\tiny{$\pm$1.56}          & \ul{74.59}\tiny{$\pm$0.96}                & \textbf{77.31}\tiny{$\pm$0.45}            \\
\multicolumn{1}{c|}{}                                                 & GSMOTE   & 61.57\tiny{$\pm$1.78}      & \ul{69.88}\tiny{$\pm$0.96}                & \textbf{72.28}\tiny{$\pm$1.48}             & 42.21\tiny{$\pm$2.12}      & \ul{60.91}\tiny{$\pm$1.33}                & \textbf{62.32}\tiny{$\pm$1.06}             & 71.55\tiny{$\pm$0.64}          & \ul{74.74}\tiny{$\pm$0.81}                & \textbf{76.14}\tiny{$\pm$0.21}            \\
\multicolumn{1}{c|}{}                                                 & GENS     & 68.84\tiny{$\pm$0.41}      & \ul{69.78}\tiny{$\pm$1.18}                & \textbf{71.92}\tiny{$\pm$0.71}             & 52.57\tiny{$\pm$1.78}      & \textbf{64.36}\tiny{$\pm$0.68}            & \ul{63.84}\tiny{$\pm$0.68}                 & 71.38\tiny{$\pm$0.99}          & \ul{75.89}\tiny{$\pm$1.17}                & \textbf{76.46}\tiny{$\pm$1.29}            \\ \cline{2-11} 
\multicolumn{1}{c|}{}                                                 & Best     & 68.84                      & \ul{71.31}                                & \textbf{73.02}                             & 52.57                      & \ul{64.36}                                & \textbf{66.35}                             & 71.55                          & \ul{75.89}                                & \textbf{77.38}                            \\ \hline
\multicolumn{1}{l|}{\multirow{7}{*}{\textbf{\rotatebox{90}{APPNP}}}}   & Vanilla  & 55.37\tiny{$\pm$1.65}      & \ul{58.13}\tiny{$\pm$1.69}                & \textbf{61.71}\tiny{$\pm$1.66}             & \ul{35.69}\tiny{$\pm$0.14} & 35.68\tiny{$\pm$0.15}                     & \textbf{36.02}\tiny{$\pm$0.25}             & \textbf{59.30}\tiny{$\pm$0.50} & 55.62\tiny{$\pm$0.31}                     & \ul{57.82}\tiny{$\pm$0.29}                \\
\multicolumn{1}{l|}{}                                                 & Reweight & \ul{72.62}\tiny{$\pm$0.47} & \textbf{73.62}\tiny{$\pm$0.89}            & 72.51\tiny{$\pm$0.87}                      & 50.88\tiny{$\pm$3.64}      & \ul{63.54}\tiny{$\pm$1.02}                & \textbf{65.57}\tiny{$\pm$1.11}             & \ul{72.00}\tiny{$\pm$0.81}     & \textbf{72.15}\tiny{$\pm$0.60}            & 71.22\tiny{$\pm$1.10}                     \\
\multicolumn{1}{l|}{}                                                 & ReNode   & \ul{73.74}\tiny{$\pm$1.12} & \textbf{75.02}\tiny{$\pm$1.54}            & 72.15\tiny{$\pm$0.76}                      & 50.50\tiny{$\pm$3.51}      & \ul{63.73}\tiny{$\pm$0.54}                & \textbf{65.13}\tiny{$\pm$0.40}             & \textbf{72.76}\tiny{$\pm$1.37} & 71.54\tiny{$\pm$0.96}                     & \ul{71.88}\tiny{$\pm$0.70}                \\
\multicolumn{1}{l|}{}                                                 & Resample & 65.78\tiny{$\pm$1.72}      & \ul{73.14}\tiny{$\pm$0.94}                & \textbf{73.57}\tiny{$\pm$0.92}             & 40.79\tiny{$\pm$1.87}      & \textbf{66.54}\tiny{$\pm$0.49}            & \ul{59.51}\tiny{$\pm$4.16}                 & 67.74\tiny{$\pm$1.94}          & \ul{72.25}\tiny{$\pm$0.81}                & \textbf{74.41}\tiny{$\pm$0.95}            \\
\multicolumn{1}{l|}{}                                                 & SMOTE    & 65.34\tiny{$\pm$1.68}      & \textbf{73.18}\tiny{$\pm$1.02}            & \ul{72.88}\tiny{$\pm$0.90}                 & 40.79\tiny{$\pm$2.05}      & \textbf{66.62}\tiny{$\pm$0.33}            & \ul{58.82}\tiny{$\pm$4.59}                 & 67.24\tiny{$\pm$2.10}          & \ul{72.67}\tiny{$\pm$1.65}                & \textbf{73.33}\tiny{$\pm$1.37}            \\
\multicolumn{1}{l|}{}                                                 & GSMOTE   & 71.13\tiny{$\pm$0.72}      & \ul{73.37}\tiny{$\pm$0.82}                & \textbf{73.78}\tiny{$\pm$0.71}             & 45.37\tiny{$\pm$2.75}      & \textbf{64.95}\tiny{$\pm$0.11}            & \ul{62.95}\tiny{$\pm$2.58}                 & 69.57\tiny{$\pm$2.20}          & \ul{73.37}\tiny{$\pm$0.95}                & \textbf{74.90}\tiny{$\pm$1.27}            \\ \cline{2-11} 
\multicolumn{1}{l|}{}                                                 & Best     & 73.74                      & \textbf{75.02}                            & \ul{73.78}                                 & 50.88                      & \textbf{66.62}                            & \ul{65.57}                                 & 72.76                          & \ul{73.37}                                & \textbf{74.90}                            \\ \hline
\multicolumn{1}{l|}{\multirow{7}{*}{\textbf{\rotatebox{90}{GPRGNN}}}} & Vanilla  & 67.97\tiny{$\pm$0.51}      & \ul{71.99}\tiny{$\pm$1.14}                & \textbf{73.38}\tiny{$\pm$1.18}             & 42.31\tiny{$\pm$2.16}      & \ul{55.85}\tiny{$\pm$0.89}                & \textbf{58.82}\tiny{$\pm$1.91}             & \ul{67.04}\tiny{$\pm$1.82}     & 57.92\tiny{$\pm$0.45}                     & \textbf{77.49}\tiny{$\pm$1.15}            \\
\multicolumn{1}{l|}{}                                                 & Reweight & 72.15\tiny{$\pm$0.57}      & \ul{72.90}\tiny{$\pm$1.33}                & \textbf{73.22}\tiny{$\pm$0.55}             & 53.22\tiny{$\pm$2.89}      & \ul{59.78}\tiny{$\pm$0.76}                & \textbf{61.00}\tiny{$\pm$1.82}             & 73.35\tiny{$\pm$1.07}          & \ul{75.22}\tiny{$\pm$1.02}                & \textbf{76.86}\tiny{$\pm$0.76}            \\
\multicolumn{1}{l|}{}                                                 & ReNode   & 73.38\tiny{$\pm$0.67}      & \ul{73.71}\tiny{$\pm$0.57}                & \textbf{73.93}\tiny{$\pm$1.60}             & 54.66\tiny{$\pm$2.82}      & \ul{59.69}\tiny{$\pm$0.73}                & \textbf{60.34}\tiny{$\pm$1.31}             & 73.56\tiny{$\pm$0.98}          & \ul{75.69}\tiny{$\pm$1.10}                & \textbf{76.25}\tiny{$\pm$0.67}            \\
\multicolumn{1}{l|}{}                                                 & Resample & 67.00\tiny{$\pm$1.33}      & \ul{72.94}\tiny{$\pm$1.02}                & \textbf{74.89}\tiny{$\pm$0.86}             & 42.27\tiny{$\pm$2.15}      & \textbf{64.16}\tiny{$\pm$0.62}            & \ul{63.89}\tiny{$\pm$0.98}                 & 70.42\tiny{$\pm$1.51}          & \ul{73.79}\tiny{$\pm$0.83}                & \textbf{75.31}\tiny{$\pm$0.54}            \\
\multicolumn{1}{l|}{}                                                 & SMOTE    & 66.99\tiny{$\pm$1.33}      & \ul{74.01}\tiny{$\pm$1.51}                & \textbf{74.41}\tiny{$\pm$1.05}             & 40.97\tiny{$\pm$2.02}      & \textbf{63.88}\tiny{$\pm$0.55}            & \ul{62.60}\tiny{$\pm$1.72}                 & 70.29\tiny{$\pm$1.47}          & \ul{73.89}\tiny{$\pm$0.69}                & \textbf{75.48}\tiny{$\pm$1.02}            \\
\multicolumn{1}{l|}{}                                                 & GSMOTE   & 70.94\tiny{$\pm$0.57}      & \ul{73.63}\tiny{$\pm$1.25}                & \textbf{74.02}\tiny{$\pm$0.90}             & 48.01\tiny{$\pm$3.28}      & \textbf{63.03}\tiny{$\pm$0.92}            & \ul{61.68}\tiny{$\pm$0.86}                 & 71.51\tiny{$\pm$1.91}          & \ul{72.16}\tiny{$\pm$0.58}                & \textbf{74.77}\tiny{$\pm$0.83}            \\ \cline{2-11} 
\multicolumn{1}{l|}{}                                                 & Best     & 73.38                      & \ul{74.01}                                & \textbf{74.89}                             & 54.66                      & \textbf{64.16}                            & \ul{63.89}                                 & 73.56                          & \ul{75.69}                                & \textbf{77.49}                            \\ \hline
\end{tabular}
}
\end{table*}

%% file: sections/tables/tab_ext_f1.tex
\begin{table*}[]
\caption{
    Macro-F1 score of combining \alg{} with 6 IGL baselines $\times$ 5 GNN backbones.
}
\label{tab:main_ext_f1}
\resizebox{\textwidth}{!}
{
\begin{tabular}{cl|lgg|lgg|lgg}
\hline
\multicolumn{2}{c|}{\textbf{Dataset (IR=10)}}                                    & \multicolumn{3}{c|}{\textbf{Cora}}                                                                                      & \multicolumn{3}{c|}{\textbf{CiteSeer}}                                                                            & \multicolumn{3}{c}{\textbf{PubMed}}                                                                                \\ \hline
\multicolumn{2}{c|}{\textbf{Metric: Macro-F1$\uparrow$}}                           & \multicolumn{1}{c}{Base}       & \multicolumn{1}{c}{+ \algp{}} & \multicolumn{1}{c|}{+ \algt{}} & \multicolumn{1}{c}{Base} & \multicolumn{1}{c}{+ \algp{}} & \multicolumn{1}{c|}{+ \algt{}} & \multicolumn{1}{c}{Base}   & \multicolumn{1}{c}{+ \algp{}} & \multicolumn{1}{c}{+ \algt{}} \\ \hline
\multicolumn{1}{c|}{\multirow{8}{*}{\textbf{\rotatebox{90}{GCN}}}}    & Vanilla  & 60.10\tiny{$\pm$1.53}          & \ul{63.28}\tiny{$\pm$1.07}                & \textbf{68.68}\tiny{$\pm$1.49}             & 28.05\tiny{$\pm$2.53}    & \ul{51.55}\tiny{$\pm$1.28}                & \textbf{54.94}\tiny{$\pm$1.44}             & 55.09\tiny{$\pm$2.48}      & \textbf{67.16}\tiny{$\pm$1.53}            & \ul{64.40}\tiny{$\pm$3.68}                \\
\multicolumn{1}{c|}{}                                                 & Reweight & 67.85\tiny{$\pm$0.62}          & \ul{69.41}\tiny{$\pm$1.01}                & \textbf{70.31}\tiny{$\pm$0.82}             & 36.59\tiny{$\pm$3.66}    & \ul{56.84}\tiny{$\pm$1.06}                & \textbf{57.54}\tiny{$\pm$1.08}             & 67.07\tiny{$\pm$3.42}      & \ul{72.94}\tiny{$\pm$0.81}                & \textbf{73.24}\tiny{$\pm$0.90}            \\
\multicolumn{1}{c|}{}                                                 & ReNode   & 66.66\tiny{$\pm$1.59}          & \ul{69.79}\tiny{$\pm$0.79}                & \textbf{70.59}\tiny{$\pm$1.25}             & 34.64\tiny{$\pm$1.54}    & \ul{56.69}\tiny{$\pm$0.64}                & \textbf{58.07}\tiny{$\pm$0.77}             & 67.86\tiny{$\pm$3.99}      & \textbf{72.61}\tiny{$\pm$0.41}            & \ul{72.25}\tiny{$\pm$0.89}                \\
\multicolumn{1}{c|}{}                                                 & Resample & 57.34\tiny{$\pm$2.27}          & \ul{71.36}\tiny{$\pm$0.39}                & \textbf{72.82}\tiny{$\pm$1.13}             & 29.73\tiny{$\pm$2.77}    & \ul{57.17}\tiny{$\pm$0.48}                & \textbf{58.03}\tiny{$\pm$1.42}             & 56.74\tiny{$\pm$3.54}      & \ul{71.19}\tiny{$\pm$0.83}                & \textbf{73.13}\tiny{$\pm$1.33}            \\
\multicolumn{1}{c|}{}                                                 & SMOTE    & 55.65\tiny{$\pm$1.62}          & \ul{71.04}\tiny{$\pm$0.16}                & \textbf{72.82}\tiny{$\pm$0.86}             & 29.39\tiny{$\pm$2.81}    & \ul{59.53}\tiny{$\pm$0.88}                & \textbf{61.53}\tiny{$\pm$1.24}             & 56.14\tiny{$\pm$3.74}      & \ul{71.72}\tiny{$\pm$0.60}                & \textbf{72.83}\tiny{$\pm$1.20}            \\
\multicolumn{1}{c|}{}                                                 & GSMOTE   & 67.60\tiny{$\pm$1.67}     & \ul{68.01}\tiny{$\pm$1.00}                     & \textbf{70.28}\tiny{$\pm$0.48}             & 40.07\tiny{$\pm$3.02}    & \textbf{56.64}\tiny{$\pm$1.09}            & \ul{56.25}\tiny{$\pm$1.50}                 & 70.60\tiny{$\pm$1.17}      & \ul{72.95}\tiny{$\pm$1.39}                & \textbf{75.70}\tiny{$\pm$0.35}            \\
\multicolumn{1}{c|}{}                                                 & GENS     & 69.96\tiny{$\pm$0.29}          & \ul{71.62}\tiny{$\pm$0.64}                & \textbf{72.28}\tiny{$\pm$0.65}             & 54.45\tiny{$\pm$1.69}    & \ul{59.89}\tiny{$\pm$0.68}                & \textbf{62.46}\tiny{$\pm$0.43}             & 71.28\tiny{$\pm$1.84}      & \ul{75.77}\tiny{$\pm$0.55}                & \textbf{76.86}\tiny{$\pm$0.93}            \\ \cline{2-11} 
\multicolumn{1}{c|}{}                                                 & Best     & 69.96                          & \ul{71.62}                                & \textbf{72.82}                             & 54.45                    & \ul{59.89}                                & \textbf{62.46}                             & 71.28                      & \ul{75.77}                                & \textbf{76.86}                            \\ \hline
\multicolumn{1}{c|}{\multirow{8}{*}{\textbf{\rotatebox{90}{GAT}}}}    & Vanilla  & 60.71\tiny{$\pm$1.61}          & \ul{64.27}\tiny{$\pm$0.95}                & \textbf{68.93}\tiny{$\pm$0.79}             & 31.12\tiny{$\pm$3.15}    & \ul{54.71}\tiny{$\pm$1.18}                & \textbf{59.42}\tiny{$\pm$1.55}             & 57.32\tiny{$\pm$1.55}      & \ul{71.27}\tiny{$\pm$1.11}                & \textbf{74.03}\tiny{$\pm$1.08}            \\
\multicolumn{1}{c|}{}                                                 & Reweight & 66.49\tiny{$\pm$1.34}          & \textbf{69.84}\tiny{$\pm$0.91}            & \ul{69.79}\tiny{$\pm$0.77}                 & 34.94\tiny{$\pm$4.09}    & \ul{58.53}\tiny{$\pm$0.68}                & \textbf{60.28}\tiny{$\pm$1.12}             & 63.82\tiny{$\pm$1.60}      & \textbf{75.13}\tiny{$\pm$1.13}            & \ul{73.88}\tiny{$\pm$1.38}                \\
\multicolumn{1}{c|}{}                                                 & ReNode   & 67.27\tiny{$\pm$1.23}          & \textbf{70.61}\tiny{$\pm$0.83}            & \ul{68.24}\tiny{$\pm$1.48}                 & 37.72\tiny{$\pm$2.61}    & \ul{57.64}\tiny{$\pm$2.11}                & \textbf{58.57}\tiny{$\pm$0.75}             & 67.38\tiny{$\pm$3.22}      & \ul{74.88}\tiny{$\pm$0.99}                & \textbf{74.96}\tiny{$\pm$1.18}            \\
\multicolumn{1}{c|}{}                                                 & Resample & 55.36\tiny{$\pm$2.47}          & \ul{70.87}\tiny{$\pm$0.94}                & \textbf{72.31}\tiny{$\pm$1.07}             & 25.71\tiny{$\pm$1.97}    & \textbf{59.77}\tiny{$\pm$1.31}            & \ul{59.66}\tiny{$\pm$0.95}                 & 57.24\tiny{$\pm$1.54}      & \ul{72.53}\tiny{$\pm$0.66}                & \textbf{73.09}\tiny{$\pm$0.83}            \\
\multicolumn{1}{c|}{}                                                 & SMOTE    & 57.49\tiny{$\pm$0.60}          & \ul{69.68}\tiny{$\pm$0.66}                & \textbf{71.74}\tiny{$\pm$1.03}             & 26.05\tiny{$\pm$2.30}    & \ul{59.83}\tiny{$\pm$1.33}                & \textbf{61.75}\tiny{$\pm$1.30}             & 55.66\tiny{$\pm$2.76}      & \textbf{73.33}\tiny{$\pm$1.00}            & \ul{73.30}\tiny{$\pm$0.16}                \\
\multicolumn{1}{c|}{}                                                 & GSMOTE   & 64.34\tiny{$\pm$1.69}          & \ul{68.23}\tiny{$\pm$1.80}                & \textbf{69.77}\tiny{$\pm$1.08}             & 35.07\tiny{$\pm$1.77}    & \ul{55.86}\tiny{$\pm$1.10}                & \textbf{57.13}\tiny{$\pm$2.69}             & 63.35\tiny{$\pm$2.92}      & \textbf{74.23}\tiny{$\pm$0.84}            & \ul{73.34}\tiny{$\pm$2.06}                \\
\multicolumn{1}{c|}{}                                                 & GENS     & \ul{69.96}\tiny{$\pm$0.62}     & 69.83\tiny{$\pm$0.41}                     & \textbf{70.71}\tiny{$\pm$1.16}             & 48.34\tiny{$\pm$2.19}    & \ul{60.04}\tiny{$\pm$1.85}                & \textbf{62.55}\tiny{$\pm$0.86}             & 71.78\tiny{$\pm$1.19}      & \ul{72.69}\tiny{$\pm$0.84}                & \textbf{74.42}\tiny{$\pm$1.12}            \\ \cline{2-11} 
\multicolumn{1}{c|}{}                                                 & Best     & 69.96                          & \ul{70.87}                                & \textbf{72.31}                             & 48.34                    & \ul{60.04}                                & \textbf{62.55}                             & 71.78                      & \textbf{75.13}                            & \ul{74.96}                                \\ \hline
\multicolumn{1}{c|}{\multirow{8}{*}{\textbf{\rotatebox{90}{SAGE}}}}   & Vanilla  & 57.36\tiny{$\pm$1.77}          & \ul{64.90}\tiny{$\pm$0.87}                & \textbf{65.61}\tiny{$\pm$0.97}             & 36.07\tiny{$\pm$1.06}    & \textbf{54.76}\tiny{$\pm$2.47}            & \ul{51.86}\tiny{$\pm$3.25}                 & 63.75\tiny{$\pm$1.24}      & \ul{74.35}\tiny{$\pm$0.74}                & \textbf{76.92}\tiny{$\pm$0.63}            \\
\multicolumn{1}{c|}{}                                                 & Reweight & 63.72\tiny{$\pm$1.10}          & \ul{69.06}\tiny{$\pm$0.90}                & \textbf{69.59}\tiny{$\pm$0.53}             & 39.64\tiny{$\pm$2.57}    & \textbf{57.17}\tiny{$\pm$0.76}            & \ul{54.83}\tiny{$\pm$0.74}                 & 62.83\tiny{$\pm$2.57}      & \ul{73.88}\tiny{$\pm$0.40}                & \textbf{75.42}\tiny{$\pm$0.48}            \\
\multicolumn{1}{c|}{}                                                 & ReNode   & 65.59\tiny{$\pm$1.44}          & \textbf{69.99}\tiny{$\pm$1.35}            & \ul{69.86}\tiny{$\pm$1.27}                 & 44.20\tiny{$\pm$3.68}    & \ul{55.41}\tiny{$\pm$0.48}                & \textbf{55.78}\tiny{$\pm$1.63}             & 64.97\tiny{$\pm$3.00}      & \ul{74.33}\tiny{$\pm$0.20}                & \textbf{74.88}\tiny{$\pm$0.53}            \\
\multicolumn{1}{c|}{}                                                 & Resample & 55.29\tiny{$\pm$2.12}          & \ul{70.40}\tiny{$\pm$1.11}                & \textbf{71.49}\tiny{$\pm$0.79}             & 30.14\tiny{$\pm$2.20}    & \ul{60.71}\tiny{$\pm$1.25}                & \textbf{61.29}\tiny{$\pm$1.48}             & 65.23\tiny{$\pm$2.26}      & \ul{74.28}\tiny{$\pm$0.96}                & \textbf{75.48}\tiny{$\pm$0.44}            \\
\multicolumn{1}{c|}{}                                                 & SMOTE    & 56.72\tiny{$\pm$2.69}          & \ul{69.42}\tiny{$\pm$1.29}                & \textbf{71.71}\tiny{$\pm$1.94}             & 29.22\tiny{$\pm$2.33}    & \ul{63.61}\tiny{$\pm$0.87}                & \textbf{65.91}\tiny{$\pm$0.68}             & 57.60\tiny{$\pm$3.22}      & \ul{72.98}\tiny{$\pm$0.69}                & \textbf{76.45}\tiny{$\pm$0.77}            \\
\multicolumn{1}{c|}{}                                                 & GSMOTE   & 59.44\tiny{$\pm$2.25}          & \ul{69.10}\tiny{$\pm$0.95}                & \textbf{71.30}\tiny{$\pm$1.47}             & 34.86\tiny{$\pm$3.46}    & \ul{60.53}\tiny{$\pm$1.27}                & \textbf{61.96}\tiny{$\pm$1.12}             & 67.23\tiny{$\pm$0.61}      & \ul{74.36}\tiny{$\pm$1.02}                & \textbf{75.68}\tiny{$\pm$0.31}            \\
\multicolumn{1}{c|}{}                                                 & GENS     & 68.23\tiny{$\pm$0.72}          & \ul{69.76}\tiny{$\pm$0.95}                & \textbf{71.11}\tiny{$\pm$0.81}             & 51.05\tiny{$\pm$2.03}    & \textbf{63.87}\tiny{$\pm$0.82}            & \ul{63.41}\tiny{$\pm$0.57}                 & 70.06\tiny{$\pm$0.86}      & \ul{75.33}\tiny{$\pm$1.46}                & \textbf{76.01}\tiny{$\pm$1.14}            \\ \cline{2-11} 
\multicolumn{1}{c|}{}                                                 & Best     & 68.23                          & \ul{70.40}                                & \textbf{71.71}                             & 51.05                    & \ul{63.87}                                & \textbf{65.91}                             & 70.06                      & \ul{75.33}                                & \textbf{76.92}                            \\ \hline
\multicolumn{1}{l|}{\multirow{7}{*}{\textbf{\rotatebox{90}{APPNP}}}}   & Vanilla  & 50.39\tiny{$\pm$2.81}          & \ul{54.19}\tiny{$\pm$2.58}                & \textbf{59.99}\tiny{$\pm$2.49}             & 22.21\tiny{$\pm$0.13}    & \ul{22.54}\tiny{$\pm$0.25}                & \textbf{22.89}\tiny{$\pm$0.22}             & 44.50\tiny{$\pm$0.21}      & \textbf{44.67}\tiny{$\pm$0.07}            & \ul{44.59}\tiny{$\pm$0.06}                \\
\multicolumn{1}{l|}{}                                                 & Reweight & \ul{72.63}\tiny{$\pm$0.53}     & \textbf{72.71}\tiny{$\pm$0.60}            & 70.61\tiny{$\pm$0.65}                      & 45.25\tiny{$\pm$4.85}    & \ul{63.08}\tiny{$\pm$1.03}                & \textbf{65.20}\tiny{$\pm$1.20}             & 69.53\tiny{$\pm$1.14}      & \ul{72.24}\tiny{$\pm$0.58}                & \textbf{72.26}\tiny{$\pm$0.80}            \\
\multicolumn{1}{l|}{}                                                 & ReNode   & \ul{73.67}\tiny{$\pm$0.98}     & \textbf{73.67}\tiny{$\pm$1.18}            & 69.79\tiny{$\pm$0.72}                      & 44.91\tiny{$\pm$4.99}    & \ul{62.97}\tiny{$\pm$0.78}                & \textbf{64.47}\tiny{$\pm$0.40}             & 70.65\tiny{$\pm$1.66}      & \textbf{72.33}\tiny{$\pm$0.90}            & \ul{72.18}\tiny{$\pm$0.55}                \\
\multicolumn{1}{l|}{}                                                 & Resample & 65.20\tiny{$\pm$2.08}          & \ul{72.25}\tiny{$\pm$0.82}                & \textbf{72.72}\tiny{$\pm$0.97}             & 31.04\tiny{$\pm$2.76}    & \textbf{66.06}\tiny{$\pm$0.54}            & \ul{54.57}\tiny{$\pm$6.08}                 & 62.42\tiny{$\pm$3.62}      & \ul{72.32}\tiny{$\pm$0.93}                & \textbf{74.27}\tiny{$\pm$1.08}            \\
\multicolumn{1}{l|}{}                                                 & SMOTE    & 64.70\tiny{$\pm$2.06}          & \textbf{72.90}\tiny{$\pm$0.83}            & \ul{72.31}\tiny{$\pm$0.94}                 & 30.90\tiny{$\pm$2.86}    & \textbf{66.18}\tiny{$\pm$0.37}            & \ul{53.90}\tiny{$\pm$6.26}                 & 61.83\tiny{$\pm$3.65}      & \ul{72.55}\tiny{$\pm$1.61}                & \textbf{73.87}\tiny{$\pm$1.37}            \\
\multicolumn{1}{l|}{}                                                 & GSMOTE   & 71.20\tiny{$\pm$0.67}          & \ul{73.02}\tiny{$\pm$0.74}                & \textbf{73.22}\tiny{$\pm$0.92}             & 37.90\tiny{$\pm$4.29}    & \textbf{64.56}\tiny{$\pm$0.18}            & \ul{60.41}\tiny{$\pm$3.84}                 & 65.65\tiny{$\pm$3.06}      & \ul{72.54}\tiny{$\pm$0.85}                & \textbf{74.61}\tiny{$\pm$1.36}            \\ \cline{2-11} 
\multicolumn{1}{l|}{}                                                 & Best     & \ul{73.67}                     & \textbf{73.67}                            & 73.22                                      & 45.25                    & \textbf{66.18}                            & \ul{65.20}                                 & 70.65                      & \ul{72.55}                                & \textbf{74.61}                            \\ \hline
\multicolumn{1}{l|}{\multirow{7}{*}{\textbf{\rotatebox{90}{GPRGNN}}}} & Vanilla  & 67.86\tiny{$\pm$0.79}          & \ul{70.80}\tiny{$\pm$1.16}                & \textbf{72.32}\tiny{$\pm$1.18}             & 35.00\tiny{$\pm$2.96}    & \ul{55.06}\tiny{$\pm$0.89}                & \textbf{56.31}\tiny{$\pm$2.87}             & \ul{59.01}\tiny{$\pm$3.62} & 50.12\tiny{$\pm$1.46}                     & \textbf{77.62}\tiny{$\pm$1.04}            \\
\multicolumn{1}{l|}{}                                                 & Reweight & \textbf{71.66}\tiny{$\pm$0.85} & 70.46\tiny{$\pm$0.98}                     & \ul{71.24}\tiny{$\pm$0.49}                 & 49.19\tiny{$\pm$3.61}    & \ul{59.11}\tiny{$\pm$0.73}                & \textbf{60.30}\tiny{$\pm$2.04}             & 71.18\tiny{$\pm$0.95}      & \ul{75.47}\tiny{$\pm$0.90}                & \textbf{77.01}\tiny{$\pm$0.52}            \\
\multicolumn{1}{l|}{}                                                 & ReNode   & \textbf{73.08}\tiny{$\pm$0.66} & 71.52\tiny{$\pm$0.50}                     & \ul{71.72}\tiny{$\pm$1.51}                 & 50.34\tiny{$\pm$3.18}    & \textbf{59.10}\tiny{$\pm$0.75}            & \ul{58.94}\tiny{$\pm$1.36}                 & 71.45\tiny{$\pm$1.19}      & \ul{75.08}\tiny{$\pm$1.06}                & \textbf{75.76}\tiny{$\pm$0.84}            \\
\multicolumn{1}{l|}{}                                                 & Resample & 66.42\tiny{$\pm$1.65}          & \ul{71.70}\tiny{$\pm$0.86}                & \textbf{73.54}\tiny{$\pm$0.83}             & 32.60\tiny{$\pm$2.71}    & \textbf{63.59}\tiny{$\pm$0.65}            & \ul{63.12}\tiny{$\pm$1.06}                 & 66.58\tiny{$\pm$2.08}      & \ul{73.66}\tiny{$\pm$0.86}                & \textbf{75.42}\tiny{$\pm$0.35}            \\
\multicolumn{1}{l|}{}                                                 & SMOTE    & 66.43\tiny{$\pm$1.74}          & \ul{72.89}\tiny{$\pm$1.23}                & \textbf{73.47}\tiny{$\pm$1.12}             & 31.38\tiny{$\pm$2.70}    & \textbf{63.41}\tiny{$\pm$0.55}            & \ul{61.23}\tiny{$\pm$2.59}                 & 66.78\tiny{$\pm$1.97}      & \ul{73.98}\tiny{$\pm$0.70}                & \textbf{75.63}\tiny{$\pm$0.91}            \\
\multicolumn{1}{l|}{}                                                 & GSMOTE   & 70.87\tiny{$\pm$0.53}          & \ul{72.53}\tiny{$\pm$0.85}                & \textbf{73.12}\tiny{$\pm$0.95}             & 42.82\tiny{$\pm$4.52}    & \textbf{62.09}\tiny{$\pm$1.04}            & \ul{60.82}\tiny{$\pm$0.88}                 & 67.93\tiny{$\pm$3.01}      & \ul{72.72}\tiny{$\pm$0.72}                & \textbf{74.66}\tiny{$\pm$0.74}            \\ \cline{2-11} 
\multicolumn{1}{l|}{}                                                 & Best     & \ul{73.08}                     & 72.89                                     & \textbf{73.54}                             & 50.34                    & \textbf{63.59}                            & \ul{63.12}                                 & 71.45                      & \ul{75.47}                                & \textbf{77.62}                            \\ \hline
\end{tabular}
}
\end{table*}

%% file: sections/tables/tab_ext_disp.tex
\begin{table*}[]
\caption{
    Performance deviation of combining \alg{} with 6 IGL baselines $\times$ 5 GNN backbones.
}
\label{tab:main_ext_disp}
\resizebox{\textwidth}{!}
{
\begin{tabular}{cl|lgg|lgg|lgg}
\hline
\multicolumn{2}{c|}{\textbf{Dataset (IR=10)}}                                    & \multicolumn{3}{c|}{\textbf{Cora}}                                                                                  & \multicolumn{3}{c|}{\textbf{CiteSeer}}                                                                              & \multicolumn{3}{c}{\textbf{PubMed}}                                                                                \\ \hline
\multicolumn{2}{c|}{\textbf{Metric: PerfStd$\downarrow$}}                        & \multicolumn{1}{c}{Base}   & \multicolumn{1}{c}{+ \algp{}} & \multicolumn{1}{c|}{+ \algt{}} & \multicolumn{1}{c}{Base}   & \multicolumn{1}{c}{+ \algp{}} & \multicolumn{1}{c|}{+ \algt{}} & \multicolumn{1}{c}{Base}   & \multicolumn{1}{c}{+ \algp{}} & \multicolumn{1}{c}{+ \algt{}} \\ \hline
\multicolumn{1}{c|}{\multirow{8}{*}{\textbf{\rotatebox{90}{GCN}}}}    & Vanilla  & 27.88\tiny{$\pm$1.79}      & \ul{21.27}\tiny{$\pm$1.76}                & \textbf{18.49}\tiny{$\pm$2.68}             & 29.93\tiny{$\pm$1.38}      & \textbf{13.82}\tiny{$\pm$2.06}            & \ul{13.93}\tiny{$\pm$0.80}                 & 34.73\tiny{$\pm$2.14}      & \textbf{9.23}\tiny{$\pm$2.78}             & \ul{21.81}\tiny{$\pm$4.51}                \\
\multicolumn{1}{c|}{}                                                 & Reweight & 22.29\tiny{$\pm$1.41}      & \textbf{14.43}\tiny{$\pm$2.51}            & \ul{18.32}\tiny{$\pm$2.20}                 & 25.47\tiny{$\pm$1.78}      & \textbf{19.10}\tiny{$\pm$1.48}            & \ul{22.64}\tiny{$\pm$0.79}                 & 19.33\tiny{$\pm$5.26}      & \ul{10.21}\tiny{$\pm$1.58}                & \textbf{5.88}\tiny{$\pm$0.83}             \\
\multicolumn{1}{c|}{}                                                 & ReNode   & 22.88\tiny{$\pm$1.64}      & \textbf{14.65}\tiny{$\pm$2.07}            & \ul{17.00}\tiny{$\pm$2.13}                 & 30.31\tiny{$\pm$1.51}      & \textbf{20.22}\tiny{$\pm$0.88}            & \ul{22.99}\tiny{$\pm$1.09}                 & 18.14\tiny{$\pm$5.79}      & \ul{12.99}\tiny{$\pm$1.57}                & \textbf{10.96}\tiny{$\pm$1.92}            \\
\multicolumn{1}{c|}{}                                                 & Resample & 31.57\tiny{$\pm$1.85}      & \textbf{15.13}\tiny{$\pm$2.14}            & \ul{15.25}\tiny{$\pm$2.79}                 & 31.00\tiny{$\pm$1.32}      & \textbf{16.30}\tiny{$\pm$1.89}            & \ul{20.79}\tiny{$\pm$0.43}                 & 30.90\tiny{$\pm$5.67}      & \ul{11.63}\tiny{$\pm$3.20}                & \textbf{7.82}\tiny{$\pm$0.80}             \\
\multicolumn{1}{c|}{}                                                 & SMOTE    & 33.32\tiny{$\pm$1.38}      & \textbf{16.33}\tiny{$\pm$1.12}            & \ul{17.95}\tiny{$\pm$2.50}                 & 32.61\tiny{$\pm$1.45}      & \textbf{17.27}\tiny{$\pm$0.86}            & \ul{18.25}\tiny{$\pm$0.89}                 & 31.79\tiny{$\pm$5.21}      & \textbf{10.56}\tiny{$\pm$1.82}            & \ul{11.66}\tiny{$\pm$2.58}                \\
\multicolumn{1}{c|}{}                                                 & GSMOTE   & 21.78\tiny{$\pm$1.79}      & \textbf{17.90}\tiny{$\pm$2.75}            & \ul{18.44}\tiny{$\pm$2.20}                 & 22.64\tiny{$\pm$2.69}      & \ul{21.37}\tiny{$\pm$1.25}                & \textbf{21.01}\tiny{$\pm$1.45}             & 15.87\tiny{$\pm$2.34}      & \textbf{3.35}\tiny{$\pm$1.08}             & \ul{5.83}\tiny{$\pm$1.27}                 \\
\multicolumn{1}{c|}{}                                                 & GENS     & 20.04\tiny{$\pm$1.12}      & \textbf{16.98}\tiny{$\pm$3.02}            & \ul{18.02}\tiny{$\pm$2.23}                 & 16.95\tiny{$\pm$2.64}      & \textbf{14.94}\tiny{$\pm$0.75}            & \ul{15.54}\tiny{$\pm$0.60}                 & 11.93\tiny{$\pm$3.46}      & \ul{5.95}\tiny{$\pm$1.85}                 & \textbf{5.15}\tiny{$\pm$0.80}             \\ \cline{2-11} 
\multicolumn{1}{c|}{}                                                 & Best     & 20.04                      & \textbf{14.43}                            & \ul{15.25}                                 & 16.95                      & \textbf{13.82}                            & \ul{13.93}                                 & 11.93                      & \textbf{3.35}                             & \ul{5.15}                                 \\ \hline
\multicolumn{1}{c|}{\multirow{8}{*}{\textbf{\rotatebox{90}{GAT}}}}    & Vanilla  & 27.38\tiny{$\pm$1.71}      & \ul{19.23}\tiny{$\pm$0.80}                & \textbf{17.97}\tiny{$\pm$2.65}             & 28.32\tiny{$\pm$2.07}      & \textbf{15.62}\tiny{$\pm$0.77}            & \ul{15.90}\tiny{$\pm$0.95}                 & 30.94\tiny{$\pm$1.27}      & \ul{10.77}\tiny{$\pm$2.04}                & \textbf{8.51}\tiny{$\pm$2.48}             \\
\multicolumn{1}{c|}{}                                                 & Reweight & 22.90\tiny{$\pm$1.67}      & \textbf{16.44}\tiny{$\pm$2.71}            & \ul{17.32}\tiny{$\pm$2.83}                 & 30.27\tiny{$\pm$1.26}      & \textbf{18.64}\tiny{$\pm$1.30}            & \ul{20.83}\tiny{$\pm$1.06}                 & 24.92\tiny{$\pm$1.88}      & \textbf{3.01}\tiny{$\pm$0.96}             & \ul{5.44}\tiny{$\pm$1.33}                 \\
\multicolumn{1}{c|}{}                                                 & ReNode   & 23.13\tiny{$\pm$1.54}      & \textbf{15.05}\tiny{$\pm$1.59}            & \ul{18.96}\tiny{$\pm$1.65}                 & 25.21\tiny{$\pm$1.85}      & \textbf{20.48}\tiny{$\pm$0.82}            & \ul{20.51}\tiny{$\pm$0.49}                 & 18.15\tiny{$\pm$4.37}      & \textbf{4.77}\tiny{$\pm$1.22}             & \ul{6.17}\tiny{$\pm$0.42}                 \\
\multicolumn{1}{c|}{}                                                 & Resample & 32.73\tiny{$\pm$2.12}      & \ul{17.87}\tiny{$\pm$2.04}                & \textbf{17.64}\tiny{$\pm$2.57}             & 32.59\tiny{$\pm$0.89}      & \textbf{17.76}\tiny{$\pm$1.79}            & \ul{18.73}\tiny{$\pm$1.02}                 & 31.67\tiny{$\pm$0.98}      & \ul{6.18}\tiny{$\pm$1.35}                 & \textbf{4.58}\tiny{$\pm$1.14}             \\
\multicolumn{1}{c|}{}                                                 & SMOTE    & 31.17\tiny{$\pm$0.69}      & \ul{18.40}\tiny{$\pm$1.03}                & \textbf{18.26}\tiny{$\pm$1.87}             & 33.32\tiny{$\pm$0.88}      & \textbf{10.68}\tiny{$\pm$0.68}            & \ul{13.24}\tiny{$\pm$1.12}                 & 32.79\tiny{$\pm$2.13}      & \ul{8.14}\tiny{$\pm$2.00}                 & \textbf{7.56}\tiny{$\pm$1.05}             \\
\multicolumn{1}{c|}{}                                                 & GSMOTE   & 24.84\tiny{$\pm$1.60}      & \textbf{15.48}\tiny{$\pm$2.08}            & \ul{18.23}\tiny{$\pm$2.00}                 & 26.74\tiny{$\pm$1.53}      & \textbf{18.34}\tiny{$\pm$1.64}            & \ul{19.76}\tiny{$\pm$0.42}                 & 24.50\tiny{$\pm$2.78}      & \textbf{5.12}\tiny{$\pm$1.38}             & \ul{8.54}\tiny{$\pm$2.03}                 \\
\multicolumn{1}{c|}{}                                                 & GENS     & 20.08\tiny{$\pm$1.56}      & \textbf{17.75}\tiny{$\pm$2.40}            & \ul{17.88}\tiny{$\pm$2.50}                 & 26.49\tiny{$\pm$1.18}      & \textbf{12.89}\tiny{$\pm$0.77}            & \ul{15.09}\tiny{$\pm$0.95}                 & 10.29\tiny{$\pm$2.75}      & \ul{7.83}\tiny{$\pm$2.27}                 & \textbf{7.55}\tiny{$\pm$2.38}             \\ \cline{2-11} 
\multicolumn{1}{c|}{}                                                 & Best     & 20.08                      & \textbf{15.05}                            & \ul{17.32}                                 & 25.21                      & \textbf{10.68}                            & \ul{13.24}                                 & 10.29                      & \textbf{3.01}                             & \ul{4.58}                                 \\ \hline
\multicolumn{1}{c|}{\multirow{8}{*}{\textbf{\rotatebox{90}{SAGE}}}}   & Vanilla  & 29.94\tiny{$\pm$1.75}      & \textbf{18.62}\tiny{$\pm$2.13}            & \ul{19.49}\tiny{$\pm$1.67}                 & 26.75\tiny{$\pm$1.58}      & \textbf{14.56}\tiny{$\pm$1.16}            & \ul{18.13}\tiny{$\pm$1.32}                 & 21.09\tiny{$\pm$3.43}      & \ul{10.96}\tiny{$\pm$1.99}                & \textbf{4.09}\tiny{$\pm$1.17}             \\
\multicolumn{1}{c|}{}                                                 & Reweight & 25.61\tiny{$\pm$1.60}      & \textbf{15.24}\tiny{$\pm$2.66}            & \ul{17.54}\tiny{$\pm$2.45}                 & 29.95\tiny{$\pm$1.83}      & \textbf{19.05}\tiny{$\pm$1.60}            & \ul{22.94}\tiny{$\pm$0.49}                 & 25.47\tiny{$\pm$3.49}      & \textbf{3.35}\tiny{$\pm$0.72}             & \ul{8.09}\tiny{$\pm$0.19}                 \\
\multicolumn{1}{c|}{}                                                 & ReNode   & 24.12\tiny{$\pm$1.73}      & \textbf{13.32}\tiny{$\pm$3.03}            & \ul{15.45}\tiny{$\pm$2.41}                 & \ul{22.41}\tiny{$\pm$4.31} & \textbf{22.20}\tiny{$\pm$0.97}            & 22.75\tiny{$\pm$0.87}                      & 22.92\tiny{$\pm$4.36}      & \ul{7.63}\tiny{$\pm$1.23}                 & \textbf{5.77}\tiny{$\pm$1.55}             \\
\multicolumn{1}{c|}{}                                                 & Resample & 31.66\tiny{$\pm$1.47}      & \ul{15.77}\tiny{$\pm$2.75}                & \textbf{15.08}\tiny{$\pm$2.74}             & 30.29\tiny{$\pm$1.16}      & \ul{18.72}\tiny{$\pm$0.90}                & \textbf{18.48}\tiny{$\pm$2.00}             & 21.41\tiny{$\pm$2.88}      & \textbf{4.68}\tiny{$\pm$1.42}             & \ul{4.76}\tiny{$\pm$1.09}                 \\
\multicolumn{1}{c|}{}                                                 & SMOTE    & 30.86\tiny{$\pm$2.64}      & \ul{17.30}\tiny{$\pm$2.09}                & \textbf{14.87}\tiny{$\pm$3.23}             & 32.07\tiny{$\pm$1.00}      & \ul{13.17}\tiny{$\pm$1.33}                & \textbf{12.78}\tiny{$\pm$0.43}             & 31.62\tiny{$\pm$2.86}      & \ul{13.88}\tiny{$\pm$1.44}                & \textbf{11.63}\tiny{$\pm$2.28}            \\
\multicolumn{1}{c|}{}                                                 & GSMOTE   & 27.71\tiny{$\pm$1.86}      & \ul{17.28}\tiny{$\pm$2.25}                & \textbf{16.10}\tiny{$\pm$2.94}             & 28.77\tiny{$\pm$2.61}      & \ul{18.69}\tiny{$\pm$0.76}                & \textbf{18.05}\tiny{$\pm$1.21}             & 20.10\tiny{$\pm$0.90}      & \ul{5.37}\tiny{$\pm$1.21}                 & \textbf{4.64}\tiny{$\pm$1.61}             \\
\multicolumn{1}{c|}{}                                                 & GENS     & 19.81\tiny{$\pm$1.65}      & \textbf{17.50}\tiny{$\pm$2.05}            & \ul{17.63}\tiny{$\pm$2.11}                 & 19.76\tiny{$\pm$2.07}      & \textbf{15.99}\tiny{$\pm$0.81}            & \ul{16.99}\tiny{$\pm$0.85}                 & 11.76\tiny{$\pm$2.91}      & \textbf{7.63}\tiny{$\pm$1.51}             & \ul{8.31}\tiny{$\pm$1.64}                 \\ \cline{2-11} 
\multicolumn{1}{c|}{}                                                 & Best     & 19.81                      & \textbf{13.32}                            & \ul{14.87}                                 & 19.76                      & \ul{13.17}                                & \textbf{12.78}                             & 11.76                      & \textbf{3.35}                             & \ul{4.09}                                 \\ \hline
\multicolumn{1}{l|}{\multirow{7}{*}{\textbf{\rotatebox{90}{APPNP}}}}   & Vanilla  & 38.32\tiny{$\pm$1.94}      & \ul{35.50}\tiny{$\pm$2.10}                & \textbf{32.18}\tiny{$\pm$1.68}             & \ul{36.82}\tiny{$\pm$0.10} & \textbf{36.67}\tiny{$\pm$0.25}            & 36.83\tiny{$\pm$0.36}                      & 42.13\tiny{$\pm$0.27}      & \textbf{40.45}\tiny{$\pm$0.11}            & \ul{41.34}\tiny{$\pm$0.16}                \\
\multicolumn{1}{l|}{}                                                 & Reweight & 19.83\tiny{$\pm$1.46}      & \textbf{17.33}\tiny{$\pm$2.88}            & \ul{18.46}\tiny{$\pm$2.42}                 & 26.19\tiny{$\pm$2.93}      & \ul{20.96}\tiny{$\pm$0.58}                & \textbf{19.38}\tiny{$\pm$0.72}             & 16.96\tiny{$\pm$2.84}      & \textbf{8.04}\tiny{$\pm$1.94}             & \ul{9.13}\tiny{$\pm$1.05}                 \\
\multicolumn{1}{l|}{}                                                 & ReNode   & \ul{18.09}\tiny{$\pm$2.52} & \textbf{16.87}\tiny{$\pm$2.95}            & 19.47\tiny{$\pm$2.00}                      & 25.95\tiny{$\pm$3.66}      & \ul{22.09}\tiny{$\pm$1.43}                & \textbf{20.42}\tiny{$\pm$1.57}             & 14.49\tiny{$\pm$3.81}      & \ul{10.25}\tiny{$\pm$1.93}                & \textbf{3.95}\tiny{$\pm$1.02}             \\
\multicolumn{1}{l|}{}                                                 & Resample & 27.28\tiny{$\pm$2.13}      & \textbf{18.37}\tiny{$\pm$2.34}            & \ul{18.72}\tiny{$\pm$2.32}                 & 32.71\tiny{$\pm$1.23}      & \textbf{15.87}\tiny{$\pm$1.02}            & \ul{23.72}\tiny{$\pm$4.14}                 & 25.86\tiny{$\pm$4.38}      & \ul{13.60}\tiny{$\pm$1.68}                & \textbf{9.49}\tiny{$\pm$1.65}             \\
\multicolumn{1}{l|}{}                                                 & SMOTE    & 27.86\tiny{$\pm$1.78}      & \textbf{18.61}\tiny{$\pm$2.35}            & \ul{19.42}\tiny{$\pm$1.81}                 & 33.26\tiny{$\pm$1.10}      & \textbf{14.91}\tiny{$\pm$0.93}            & \ul{22.90}\tiny{$\pm$4.49}                 & 26.37\tiny{$\pm$4.47}      & \ul{13.37}\tiny{$\pm$1.97}                & \textbf{8.70}\tiny{$\pm$2.29}             \\
\multicolumn{1}{l|}{}                                                 & GSMOTE   & 20.98\tiny{$\pm$1.45}      & \textbf{18.19}\tiny{$\pm$2.59}            & \ul{18.55}\tiny{$\pm$2.28}                 & 29.39\tiny{$\pm$2.20}      & \textbf{16.49}\tiny{$\pm$1.12}            & \ul{19.19}\tiny{$\pm$3.44}                 & 22.32\tiny{$\pm$4.21}      & \ul{11.53}\tiny{$\pm$3.00}                & \textbf{10.69}\tiny{$\pm$2.27}            \\ \cline{2-11} 
\multicolumn{1}{l|}{}                                                 & Best     & \ul{18.09}                 & \textbf{16.87}                            & 18.46                                      & 25.95                      & \textbf{14.91}                            & \ul{19.19}                                 & 14.49                      & \ul{8.04}                                 & \textbf{3.95}                             \\ \hline
\multicolumn{1}{l|}{\multirow{7}{*}{\textbf{\rotatebox{90}{GPRGNN}}}} & Vanilla  & 22.96\tiny{$\pm$1.20}      & \ul{18.12}\tiny{$\pm$2.29}                & \textbf{17.00}\tiny{$\pm$2.98}             & 27.57\tiny{$\pm$1.32}      & \textbf{17.10}\tiny{$\pm$1.17}            & \ul{20.94}\tiny{$\pm$2.58}                 & \ul{29.94}\tiny{$\pm$3.68} & 36.57\tiny{$\pm$1.46}                     & \textbf{5.30}\tiny{$\pm$0.91}             \\
\multicolumn{1}{l|}{}                                                 & Reweight & 20.94\tiny{$\pm$1.21}      & \textbf{17.83}\tiny{$\pm$2.82}            & \ul{19.67}\tiny{$\pm$1.81}                 & 22.43\tiny{$\pm$2.39}      & \ul{21.52}\tiny{$\pm$1.06}                & \textbf{20.03}\tiny{$\pm$1.81}             & 16.12\tiny{$\pm$1.84}      & \ul{7.54}\tiny{$\pm$0.49}                 & \textbf{5.48}\tiny{$\pm$1.49}             \\
\multicolumn{1}{l|}{}                                                 & ReNode   & 18.84\tiny{$\pm$2.19}      & \textbf{16.78}\tiny{$\pm$2.53}            & \ul{17.89}\tiny{$\pm$2.96}                 & 24.14\tiny{$\pm$1.47}      & \textbf{19.84}\tiny{$\pm$1.79}            & \ul{22.83}\tiny{$\pm$1.40}                 & 14.40\tiny{$\pm$3.18}      & \ul{9.75}\tiny{$\pm$2.20}                 & \textbf{6.61}\tiny{$\pm$1.47}             \\
\multicolumn{1}{l|}{}                                                 & Resample & 25.62\tiny{$\pm$1.80}      & \ul{19.23}\tiny{$\pm$2.26}                & \textbf{17.61}\tiny{$\pm$2.77}             & 33.08\tiny{$\pm$0.66}      & \ul{17.04}\tiny{$\pm$0.78}                & \textbf{15.98}\tiny{$\pm$0.93}             & 22.59\tiny{$\pm$2.75}      & \textbf{7.62}\tiny{$\pm$2.50}             & \ul{7.76}\tiny{$\pm$0.85}                 \\
\multicolumn{1}{l|}{}                                                 & SMOTE    & 25.44\tiny{$\pm$1.88}      & \textbf{16.97}\tiny{$\pm$3.19}            & \ul{17.38}\tiny{$\pm$2.78}                 & 32.85\tiny{$\pm$0.95}      & \textbf{15.09}\tiny{$\pm$1.23}            & \ul{16.85}\tiny{$\pm$2.51}                 & 21.35\tiny{$\pm$2.76}      & \ul{9.41}\tiny{$\pm$2.67}                 & \textbf{6.09}\tiny{$\pm$0.62}             \\
\multicolumn{1}{l|}{}                                                 & GSMOTE   & 21.23\tiny{$\pm$1.48}      & \textbf{18.02}\tiny{$\pm$2.62}            & \ul{19.06}\tiny{$\pm$2.28}                 & 24.21\tiny{$\pm$3.06}      & \textbf{14.83}\tiny{$\pm$0.95}            & \ul{19.11}\tiny{$\pm$1.73}                 & 20.08\tiny{$\pm$3.77}      & \textbf{5.99}\tiny{$\pm$1.49}             & \ul{8.27}\tiny{$\pm$0.75}                 \\ \cline{2-11} 
\multicolumn{1}{l|}{}                                                 & Best     & 18.84                      & \textbf{16.78}                            & \ul{17.00}                                 & 22.43                      & \textbf{14.83}                            & \ul{15.98}                                 & 14.40                      & \ul{5.99}                                 & \textbf{5.30}                             \\ \hline
\end{tabular}
}
\end{table*}

%% file: sections/tables/tab_ext_varyimb.tex
\begin{table*}[]
\caption{
    Performance of \alg{} under varying types and levels of class imbalance.
    For each setting, we report the relative gain over base and mark the best/second-best score in \textbf{bold}/\ul{underlined}.
}
\label{tab:main_ext_varyimb}
\begin{threeparttable}
\resizebox{\textwidth}{!}
{
\begin{tabular}{ll|ll|ll|ll|ll|ll}
\toprule
\multicolumn{2}{c|}{\textbf{Dataset}} & \multicolumn{2}{c}{\textbf{Cora}} & \multicolumn{2}{c}{\textbf{CiteSeer}} & \multicolumn{2}{c}{\textbf{PubMed}} & \multicolumn{2}{c}{\textbf{CS}} & \multicolumn{2}{c}{\textbf{Physics}} \\ \midrule
\multicolumn{2}{c|}{\textbf{Step IR}} & \multicolumn{1}{c}{10} & \multicolumn{1}{c}{20} & \multicolumn{1}{c}{10} & \multicolumn{1}{c}{20} & \multicolumn{1}{c}{10} & \multicolumn{1}{c}{20} & \multicolumn{1}{c}{10} & \multicolumn{1}{c}{20} & \multicolumn{1}{c}{10} & \multicolumn{1}{c}{20} \\ \midrule
\multicolumn{1}{l|}{\multirow{4}{*}{\textbf{\rotatebox{90}{\footnotesize BAcc.$\uparrow$}}}} & \hspace{8pt}Base & 61.6 & 52.7 & 37.6 & 34.2 & 64.2 & 60.8 & 75.4 & 65.3 & 80.1 & 67.7 \\
\multicolumn{1}{l|}{} & \cellcolor{Gray}+ \alg{} & \cellcolor{Gray}69.8\tiny{+13.4\%} & \cellcolor{Gray}\ul{71.3}\tiny{+35.2\%} & \cellcolor{Gray}55.4\tiny{+47.2\%} & \cellcolor{Gray}\ul{51.3}\tiny{+49.9\%} & \cellcolor{Gray}68.6\tiny{+6.8\%} & \cellcolor{Gray}63.3\tiny{+4.1\%} & \cellcolor{Gray}82.6\tiny{+9.6\%} & \cellcolor{Gray}79.9\tiny{+22.2\%} & \cellcolor{Gray}87.6\tiny{+9.4\%} & \cellcolor{Gray}\ul{88.0}\tiny{+29.9\%} \\
\multicolumn{1}{l|}{} & \hspace{8pt}BestIGL & \ul{70.1}\tiny{+13.9\%} & 66.5\tiny{+26.2\%} & \ul{56.0}\tiny{+48.9\%} & 47.2\tiny{+38.0\%} & \ul{74.0}\tiny{+15.2\%} & \ul{71.1}\tiny{+17.0\%} & \ul{84.1}\tiny{+11.6\%} & \ul{81.3}\tiny{+24.4\%} & \ul{89.4}\tiny{+11.6\%} & 85.7\tiny{+26.6\%} \\
\multicolumn{1}{l|}{} & \cellcolor{Gray}+ \alg{} & \cellcolor{Gray}\textbf{74.2}\tiny{+20.6\%} & \cellcolor{Gray}\textbf{71.6}\tiny{+35.9\%} & \cellcolor{Gray}\textbf{62.7}\tiny{+66.6\%} & \cellcolor{Gray}\textbf{62.5}\tiny{+82.6\%} & \cellcolor{Gray}\textbf{76.9}\tiny{+19.7\%} & \cellcolor{Gray}\textbf{75.7}\tiny{+24.5\%} & \cellcolor{Gray}\textbf{86.3}\tiny{+14.5\%} & \cellcolor{Gray}\textbf{85.6}\tiny{+31.0\%} & \cellcolor{Gray}\textbf{91.2}\tiny{+13.9\%} & \cellcolor{Gray}\textbf{90.9}\tiny{+34.2\%} \\ \midrule
\multicolumn{1}{l|}{\multirow{4}{*}{\textbf{\rotatebox{90}{\footnotesize Macro-F1$\uparrow$}}}} & \hspace{8pt}Base & 60.1 & 47.0 & 28.1 & 21.9 & 55.1 & 46.4 & 72.7 & 59.2 & 80.7 & 64.7 \\
\multicolumn{1}{l|}{} & \cellcolor{Gray}+ \alg{} & \cellcolor{Gray}68.7\tiny{+14.3\%} & \cellcolor{Gray}\ul{69.6}\tiny{+48.1\%} & \cellcolor{Gray}\ul{54.9}\tiny{+95.8\%} & \cellcolor{Gray}\ul{48.9}\tiny{+123.5\%} & \cellcolor{Gray}67.2\tiny{+21.9\%} & \cellcolor{Gray}60.7\tiny{+30.8\%} & \cellcolor{Gray}78.6\tiny{+8.1\%} & \cellcolor{Gray}74.7\tiny{+26.1\%} & \cellcolor{Gray}88.8\tiny{+10.0\%} & \cellcolor{Gray}\ul{87.8}\tiny{+35.8\%} \\
\multicolumn{1}{l|}{} & \hspace{8pt}BestIGL & \ul{70.0}\tiny{+16.4\%} & 66.2\tiny{+40.9\%} & 54.5\tiny{+94.1\%} & 45.0\tiny{+105.6\%} & \ul{71.3}\tiny{+29.4\%} & \ul{68.9}\tiny{+48.3\%} & \ul{83.9}\tiny{+15.3\%} & \ul{80.9}\tiny{+36.7\%} & \ul{89.5}\tiny{+10.9\%} & 86.2\tiny{+33.2\%} \\
\multicolumn{1}{l|}{} & \cellcolor{Gray}+ \alg{} & \cellcolor{Gray}\textbf{72.8}\tiny{+21.2\%} & \cellcolor{Gray}\textbf{70.2}\tiny{+49.4\%} & \cellcolor{Gray}\textbf{62.5}\tiny{+122.7\%} & \cellcolor{Gray}\textbf{62.1}\tiny{+183.6\%} & \cellcolor{Gray}\textbf{76.9}\tiny{+39.5\%} & \cellcolor{Gray}\textbf{74.9}\tiny{+61.2\%} & \cellcolor{Gray}\textbf{85.4}\tiny{+17.5\%} & \cellcolor{Gray}\textbf{84.6}\tiny{+43.0\%} & \cellcolor{Gray}\textbf{90.7}\tiny{+12.4\%} & \cellcolor{Gray}\textbf{90.0}\tiny{+39.2\%} \\ \midrule
\multicolumn{1}{l|}{\multirow{4}{*}{\textbf{\rotatebox{90}{\footnotesize PerfStd$\downarrow$}}}} & \hspace{8pt}Base & 27.9 & 39.0 & 29.9 & 35.1 & 34.7 & 41.5 & 21.2 & 32.1 & 22.2 & 36.0 \\
\multicolumn{1}{l|}{} & \cellcolor{Gray}+ \alg{} & \cellcolor{Gray}21.3\tiny{-23.7\%} & \cellcolor{Gray}24.4\tiny{-37.5\%} & \cellcolor{Gray}\textbf{13.9}\tiny{-53.5\%} & \cellcolor{Gray}\textbf{16.7}\tiny{-52.5\%} & \cellcolor{Gray}21.8\tiny{-37.2\%} & \cellcolor{Gray}29.1\tiny{-29.9\%} & \cellcolor{Gray}17.4\tiny{-18.2\%} & \cellcolor{Gray}22.9\tiny{-28.8\%} & \cellcolor{Gray}11.5\tiny{-48.3\%} & \cellcolor{Gray}25.6\tiny{-29.0\%} \\
\multicolumn{1}{l|}{} & \hspace{8pt}BestIGL & \ul{20.0}\tiny{-28.1\%} & \ul{21.9}\tiny{-43.8\%} & 16.9\tiny{-43.4\%} & 18.0\tiny{-48.6\%} & \ul{11.9}\tiny{-65.6\%} & \ul{14.2}\tiny{-65.7\%} & \ul{8.9}\tiny{-58.3\%} & \ul{12.3}\tiny{-61.8\%} & \textbf{6.3}\tiny{-71.7\%} & \ul{12.4}\tiny{-65.5\%} \\
\multicolumn{1}{l|}{} & \cellcolor{Gray}+ \alg{} & \cellcolor{Gray}\textbf{15.2}\tiny{-45.3\%} & \cellcolor{Gray}\textbf{17.5}\tiny{-55.2\%} & \cellcolor{Gray}\ul{13.9}\tiny{-53.5\%} & \cellcolor{Gray}\ul{16.7}\tiny{-52.5\%} & \cellcolor{Gray}\textbf{5.1}\tiny{-85.2\%} & \cellcolor{Gray}\textbf{4.6}\tiny{-89.0\%} & \cellcolor{Gray}\textbf{7.9}\tiny{-62.7\%} & \cellcolor{Gray}\textbf{10.1}\tiny{-68.5\%} & \cellcolor{Gray}\ul{6.6}\tiny{-70.2\%} & \cellcolor{Gray}\textbf{6.9}\tiny{-80.8\%} \\ \midrule
\multicolumn{2}{c|}{\textbf{Natural IR}} & \multicolumn{1}{c}{50} & \multicolumn{1}{c}{100} & \multicolumn{1}{c}{50} & \multicolumn{1}{c}{100} & \multicolumn{1}{c}{50} & \multicolumn{1}{c}{100} & \multicolumn{1}{c}{50} & \multicolumn{1}{c}{100} & \multicolumn{1}{c}{50} & \multicolumn{1}{c}{100} \\ \midrule
\multicolumn{1}{l|}{\multirow{4}{*}{\textbf{\rotatebox{90}{\footnotesize BAcc.$\uparrow$}}}} & \hspace{8pt}Base & 58.1 & 61.8 & 44.9 & 44.7 & 52.0 & 51.1 & 73.8 & 71.4 & 76.0 & 77.7 \\
\multicolumn{1}{l|}{} & \cellcolor{Gray}+ \alg{} & \cellcolor{Gray}69.1\tiny{+18.9\%} & \cellcolor{Gray}68.3\tiny{+10.6\%} & \cellcolor{Gray}\ul{58.4}\tiny{+29.9\%} & \cellcolor{Gray}\ul{57.4}\tiny{+28.5\%} & \cellcolor{Gray}55.6\tiny{+7.0\%} & \cellcolor{Gray}56.5\tiny{+10.4\%} & \cellcolor{Gray}\ul{82.1}\tiny{+11.3\%} & \cellcolor{Gray}\ul{81.9}\tiny{+14.8\%} & \cellcolor{Gray}\ul{86.9}\tiny{+14.3\%} & \cellcolor{Gray}84.1\tiny{+8.3\%} \\
\multicolumn{1}{l|}{} & \hspace{8pt}BestIGL & \ul{71.0}\tiny{+22.3\%} & \ul{73.8}\tiny{+19.5\%} & 56.3\tiny{+25.3\%} & 56.3\tiny{+26.0\%} & \ul{72.7}\tiny{+39.8\%} & \ul{72.8}\tiny{+42.5\%} & 81.2\tiny{+10.0\%} & 81.4\tiny{+14.0\%} & 85.8\tiny{+12.9\%} & \ul{87.2}\tiny{+12.2\%} \\
\multicolumn{1}{l|}{} & \cellcolor{Gray}+ \alg{} & \cellcolor{Gray}\textbf{73.1}\tiny{+25.8\%} & \cellcolor{Gray}\textbf{76.9}\tiny{+24.5\%} & \cellcolor{Gray}\textbf{62.1}\tiny{+38.2\%} & \cellcolor{Gray}\textbf{61.3}\tiny{+37.3\%} & \cellcolor{Gray}\textbf{75.8}\tiny{+45.7\%} & \cellcolor{Gray}\textbf{75.9}\tiny{+48.5\%} & \cellcolor{Gray}\textbf{85.0}\tiny{+15.1\%} & \cellcolor{Gray}\textbf{84.5}\tiny{+18.5\%} & \cellcolor{Gray}\textbf{88.6}\tiny{+16.5\%} & \cellcolor{Gray}\textbf{89.7}\tiny{+15.4\%} \\ \midrule
\multicolumn{1}{l|}{\multirow{4}{*}{\textbf{\rotatebox{90}{\footnotesize Macro-F1$\uparrow$}}}} & \hspace{8pt}Base & 58.7 & 61.4 & 37.5 & 36.2 & 47.3 & 45.1 & 75.3 & 73.2 & 78.0 & 79.8 \\
\multicolumn{1}{l|}{} & \cellcolor{Gray}+ \alg{} & \cellcolor{Gray}68.7\tiny{+17.1\%} & \cellcolor{Gray}67.5\tiny{+10.0\%} & \cellcolor{Gray}\ul{57.1}\tiny{+52.6\%} & \cellcolor{Gray}\ul{55.8}\tiny{+54.3\%} & \cellcolor{Gray}52.8\tiny{+11.6\%} & \cellcolor{Gray}52.0\tiny{+15.4\%} & \cellcolor{Gray}\ul{82.6}\tiny{+9.7\%} & \cellcolor{Gray}\ul{82.6}\tiny{+12.8\%} & \cellcolor{Gray}87.6\tiny{+12.3\%} & \cellcolor{Gray}85.2\tiny{+6.8\%} \\
\multicolumn{1}{l|}{} & \hspace{8pt}BestIGL & \ul{71.1}\tiny{+21.2\%} & \ul{73.4}\tiny{+19.5\%} & 54.3\tiny{+44.8\%} & 53.8\tiny{+48.8\%} & \ul{72.9}\tiny{+53.9\%} & \ul{73.7}\tiny{+63.6\%} & 82.5\tiny{+9.5\%} & 82.4\tiny{+12.6\%} & \ul{87.7}\tiny{+12.4\%} & \ul{88.3}\tiny{+10.6\%} \\
\multicolumn{1}{l|}{} & \cellcolor{Gray}+ \alg{} & \cellcolor{Gray}\textbf{72.7}\tiny{+23.9\%} & \cellcolor{Gray}\textbf{76.0}\tiny{+23.9\%} & \cellcolor{Gray}\textbf{60.2}\tiny{+60.8\%} & \cellcolor{Gray}\textbf{59.4}\tiny{+64.3\%} & \cellcolor{Gray}\textbf{75.3}\tiny{+59.2\%} & \cellcolor{Gray}\textbf{76.1}\tiny{+68.8\%} & \cellcolor{Gray}\textbf{85.7}\tiny{+13.7\%} & \cellcolor{Gray}\textbf{85.1}\tiny{+16.2\%} & \cellcolor{Gray}\textbf{88.8}\tiny{+13.8\%} & \cellcolor{Gray}\textbf{89.4}\tiny{+12.0\%} \\ \midrule
\multicolumn{1}{l|}{\multirow{4}{*}{\textbf{\rotatebox{90}{\footnotesize PerfStd$\downarrow$}}}} & \hspace{8pt}Base & 28.8 & 31.0 & 38.7 & 39.8 & 36.2 & 38.2 & 26.3 & 28.2 & 23.8 & 21.0 \\
\multicolumn{1}{l|}{} & \cellcolor{Gray}+ \alg{} & \cellcolor{Gray}\ul{18.3}\tiny{-36.4\%} & \cellcolor{Gray}25.4\tiny{-18.1\%} & \cellcolor{Gray}\ul{24.9}\tiny{-35.6\%} & \cellcolor{Gray}33.1\tiny{-17.0\%} & \cellcolor{Gray}33.3\tiny{-8.1\%} & \cellcolor{Gray}35.9\tiny{-6.2\%} & \cellcolor{Gray}19.0\tiny{-27.9\%} & \cellcolor{Gray}19.5\tiny{-30.9\%} & \cellcolor{Gray}17.0\tiny{-28.7\%} & \cellcolor{Gray}19.6\tiny{-6.7\%} \\
\multicolumn{1}{l|}{} & \hspace{8pt}BestIGL & 18.9\tiny{-34.4\%} & \ul{17.3}\tiny{-44.4\%} & 28.7\tiny{-25.9\%} & \ul{29.7}\tiny{-25.3\%} & \ul{6.0}\tiny{-83.4\%} & \ul{9.6}\tiny{-75.0\%} & \ul{14.4}\tiny{-45.4\%} & \ul{15.4}\tiny{-45.5\%} & \ul{11.2}\tiny{-53.1\%} & \ul{9.7}\tiny{-53.8\%} \\
\multicolumn{1}{l|}{} & \cellcolor{Gray}+ \alg{} & \cellcolor{Gray}\textbf{15.9}\tiny{-44.8\%} & \cellcolor{Gray}\textbf{14.7}\tiny{-52.8\%} & \cellcolor{Gray}\textbf{21.9}\tiny{-43.4\%} & \cellcolor{Gray}\textbf{19.8}\tiny{-50.3\%} & \cellcolor{Gray}\textbf{4.2}\tiny{-88.3\%} & \cellcolor{Gray}\textbf{5.6}\tiny{-85.3\%} & \cellcolor{Gray}\textbf{12.2}\tiny{-53.5\%} & \cellcolor{Gray}\textbf{12.8}\tiny{-54.7\%} & \cellcolor{Gray}\textbf{7.4}\tiny{-68.9\%} & \cellcolor{Gray}\textbf{7.2}\tiny{-65.7\%} \\ 
\bottomrule
\end{tabular}
}
\end{threeparttable}
\end{table*}